\documentclass{article}

\usepackage[utf8]{inputenc}
\usepackage[T1]{fontenc}
\usepackage{lmodern}
\usepackage[margin=1in]{geometry}
\usepackage[sort,numbers]{natbib}
\usepackage{hyperref}
\usepackage{url}
\providecommand{\doi}[1]{\url{https://doi.org/#1}}
\usepackage{booktabs}
\usepackage{amsfonts}
\usepackage{amsmath}
\usepackage{amssymb}
\usepackage{nicefrac}
\usepackage{microtype}
\usepackage{xcolor}
\usepackage{graphicx}
\usepackage{multirow}
\usepackage{enumitem}
\usepackage{rotating}
\usepackage{tabularx}
\usepackage{xfp}
\usepackage{tikz}
\usetikzlibrary{arrows.meta,positioning,fit,shapes.geometric,calc}
\usepackage{fancyvrb}  %
\usepackage{fvextra}   %
\usepackage{longtable} %
\usepackage{xltabular} %
\newcolumntype{Y}{>{\raggedright\arraybackslash}X}

\definecolor{tierZero}{HTML}{999999}   %
\definecolor{tierOne}{HTML}{56B4E9}    %
\definecolor{tierTwo}{HTML}{0072B2}    %
\definecolor{tierThree}{HTML}{D55E00}  %
\definecolor{tierFour}{HTML}{009E73}   %

\title{Selective QA over Conflicting Multi-Source Personal Memory:\\A Diagnostic Testbed and Method Comparison}

\author{%
  Tiancheng Yang\\
  \texttt{t77yang@uwaterloo.ca}
  \and
  Matthias Schonlau\\
  \texttt{schonlau@uwaterloo.ca}
  \and
  Ilia Sucholutsky\\
  \texttt{is3060@nyu.edu}
}
\date{}

\begin{document}

\maketitle

\begin{abstract}
Emerging personal AI agents are moving toward persistent, multi-source memory. This creates an evaluation problem: systems must decide how to use conflicting or incomplete evidence; they cannot just retrieve facts from one clean history. Existing benchmarks rarely show whether an error came from the evidence given to a method or from the method's conflict-resolution step. We study this as selective QA over conflicting multi-source personal memory: systems answer based on  conflicting, sometimes incomplete sources, or abstain when evidence is insufficient. 
We develop a benchmark containing 18 question templates across 8 reasoning types, 480 personas, 4 random seeds, and 34{,}560 instances, with controlled source distortions and deterministic ground truth. We evaluate the performance of  baselines without access to any source, access to a single source, structured fusion methods, and frontier LLMs.  The best trained fusion resolver reaches 80.3\% accuracy, while the strongest prompt-only LLM baseline reaches 70.0\%. 
With abstention, the same resolver reaches 85.3\% selective accuracy at 78.3\% coverage  
and the best LLM reaches 71.0\% selective accuracy at 95.4\% coverage. 
Different models have  different strengths across reasoning types.
We release the data, code, cached model outputs, and data-generating process for reuse.
\end{abstract}

\addtocontents{toc}{\protect\setcounter{tocdepth}{-10}}

\section{Introduction}\label{sec:intro}
\begin{figure}[t]
\centering
\includegraphics[width=0.88\linewidth]{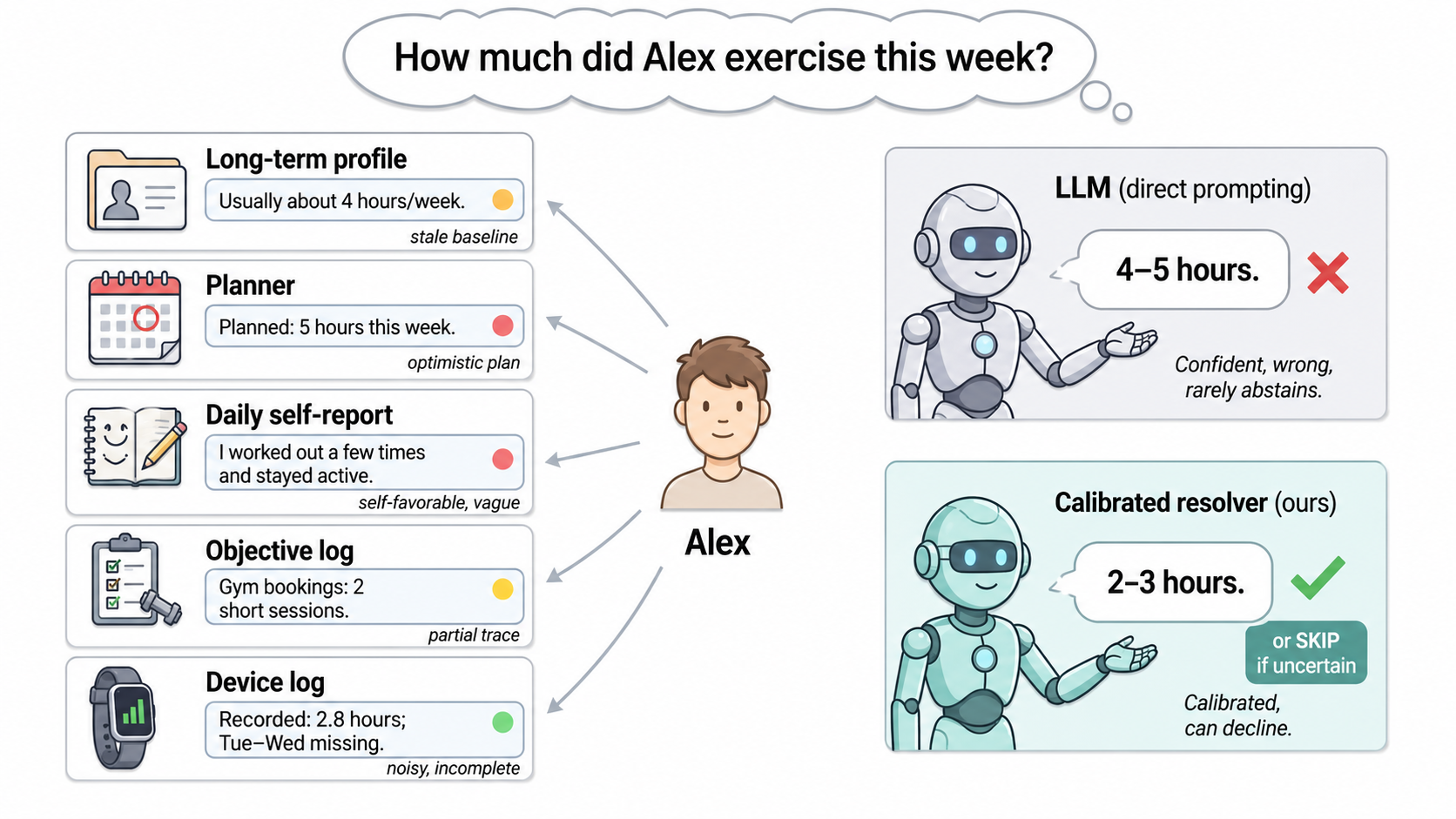}
\caption{A personal-memory agent sees five memory sources that disagree. Direct prompting can over-trust optimistic or vague sources, while a calibrated resolver can give a lower binned answer or abstain.}
\label{fig:hook}
\end{figure}

Personal AI agents are assistants that use a user's history, preferences, plans, and records to answer and act across sessions. Personal-agent systems and agent-memory architectures already point to a future setting in which this memory is layered: long-term profiles, recent logs, plans, self-reports, and device records \citep{openclaw2025,hermes2024,apple2024intelligence,doubao2025,gbrain2026,packer2024memgpt,zhang2024agent_memory_survey}. As these systems mature, they will collect more memory from more places rather than maintain one clean history. Preferences can drift, plans can be optimistic, self-reports can carry topic-dependent bias, and sensors can be missing or noisy \citep{koren2009collaborative,buehler1994planning_fallacy,paulhus2007self_report,brenner2014social_desirability,choi2011accelerometer_nonwear}. The exercise example in Figure~\ref{fig:hook} shows the resulting conflict at small scale: plans and self-reports can overstate activity, objective and device traces can be incomplete, and the resolver must decide whether to answer conservatively or abstain. This creates a deployment-relevant evaluation problem as well as an accuracy problem. If one black-box LLM both extracts memory facts and resolves conflicts, changing the model can also change the conflict policy. We use a resolver for the decision layer that maps the available memory representation, structured atoms (closed-class observations extracted from each source) or natural-language memory, to an answer or SKIP. Our benchmark separates extraction from resolution: an LLM can extract source atoms, while a structured resolver can learn source-to-ground-truth mappings from labels.

To our knowledge, conflict resolution across heterogeneous personal-memory sources has not yet been formalized as an evaluation target. Among coherent chat-history memory benchmarks, LongMemEval \citep{wu2024longmemeval} studies memory over a single user--assistant interaction history. Concurrent multi-source memory work such as LifeBench \citep{cheng2026lifebench} studies long-horizon retrieval and integration from diverse digital traces. Our setting targets a different failure mode: multiple sources can report conflicting evidence about the same latent behavior, so the system must learn source reliability and abstain when support is insufficient. Prior conflict-QA work studies cases where contexts, passages, or evidence sources disagree \citep{xie2024conflictqa,nachshoni2025natconfqa,econ2024,qacc2025}. Context-use studies ask whether QA systems rely on the provided context appropriately \citep{shaier2024desiderata}. Our benchmark targets a different setting: named personal-memory sources, such as profiles, plans, logs, and self-reports, can be biased in different ways for different questions. Reliability-aware RAG is related, but it usually learns a source-level reliability weight rather than how each source type tends to distort particular facts \citep{hwang2024rarag}. Cross-modal conflict benchmarks probe perception rather than textual evidence fusion \citep{tian2026crosscheck,popordanoska2025clash}. In multi-source personal memory, source reliability is part of the task. This lets us test whether a method can learn how each source maps to ground truth, instead of folding all conflict handling into a single prompted LLM. Existing QA benchmarks often force an answer, but personal-memory agents also need a calibrated option to abstain when sources contain irreconcilable conflict. Appendix~\ref{app:related} compares related benchmarks on multi-source evidence, directed bias, selective QA, resolver-versus-input decomposition, and synthetic data-generation support.

The testbed follows this separation explicitly. A synthetic data-generating process (DGP) creates latent daily events, projects them into five biased memory streams, renders those streams as natural-language memory, and asks resolvers to answer closed-class questions or SKIP. Structured methods first use an LLM extractor to turn each stream into these atoms; LLM baselines may instead read the natural-language memory directly. This design lets us vary the resolver and the input representation separately.

We make three contributions.

\begin{enumerate}[nosep,leftmargin=*]
  \item Task formalization. We define bias-aware selective QA over multi-source personal memory, where sources have directional distortions and resolvers may abstain under irreconcilable conflict (\S~\ref{sec:benchmark}; Appendix~\ref{app:related}).
  \item Testbed. We release a synthetic diagnostic testbed of 18 question templates $\times$ 480 personas $\times$ 4 seeds (34,560 instances), with a documented DGP, label rules, code, and runnable extension examples for new questions and evidence streams (\S~\ref{sec:benchmark}).
  \item Diagnostic evaluation. We compare a suite of structured and LLM resolvers plus a direct-readout reference, showing that learned source models, calibrated abstention, and type-specific failure signatures reveal where conflict-resolution systems break (\S~\ref{sec:experiments}).
\end{enumerate}

\section{Benchmark Design}\label{sec:benchmark}
We define selective QA over conflicting multi-source personal memory as follows. A persona $p$ is associated with five evidence streams $\mathcal{S} = \{s_1, \ldots, s_5\}$: long-term profile, planner, daily self-report, objective log, and device log. The streams describe the same 30-day window through different views: profile summaries can be stale, plans can be optimistic, self-reports can carry topic-dependent bias, objective logs can be noisy or incomplete, and device logs can have day-level and field-level missingness (full distortion model in Appendix~\ref{subsec:B-1}). Given a question $q$ with discrete answer space $\mathcal{V}_q$ ($|\mathcal{V}_q| \in \{3,4\}$), a method must produce either $v \in \mathcal{V}_q$ or SKIP. Answers are scored against a deterministic ground truth (GT) label defined by the question-specific rule. SKIP is distinct from factual edge labels in $\mathcal{V}_q$, such as \texttt{no\_plans}. We report both answer-only and selective variants in \S~\ref{sec:experiments}.

The benchmark spans 18 question templates across eight reasoning types (Table~\ref{tab:1}). Table~\ref{tab:1} groups templates by the evidence pattern a method must resolve. Seven types target different source-conflict patterns, while the control type checks calibration under minimal conflict. Topic coverage spans work, diet, social, sleep, and exercise; 15/18 questions use ordered answer scales and 3 use unordered categorical labels. Representative questions and the survey-design principles applied to all 18 are in Appendix~\ref{subsec:B-2}.

\begin{table}[b]
\centering
\caption{Question families in the diagnostic testbed. The eight reasoning types cover evidence-conflict patterns that can arise across personal-memory sources; each row states the pattern tested by the corresponding templates.}
\label{tab:1}
\small
\begin{tabularx}{\linewidth}{@{}llY@{}}
\toprule
Type & Question family & What this tests \\
\midrule
A & Source arbitration & Several streams report the same quantity with different values; the method must learn which source is reliable for that question. \\
B & Profile vs. behavior & Long-term profile or self-description claims are checked against aggregated behavior over the 30-day window. \\
C & Plan vs. reality & Planned goals are compared with actual daily execution. \\
D & Temporal trend & Early-window or stale summaries must be separated from late-window behavior. \\
E & Factor attribution & Several plausible factors are present; the method must identify the factor supported by the event-history rule. \\
F & Missing evidence & A missing observation can be informative, but it should not be confused with true absence. \\
G & Latent event annotation & Latent event attributes, such as voluntary versus required, are only partly visible through biased source labels. \\
Ctrl & Control & Minimal-conflict cases check calibration and basic answer handling. \\
\bottomrule
\end{tabularx}
\end{table}

\subsection{Data-Generating Process and Scale}\label{s2:sub1}

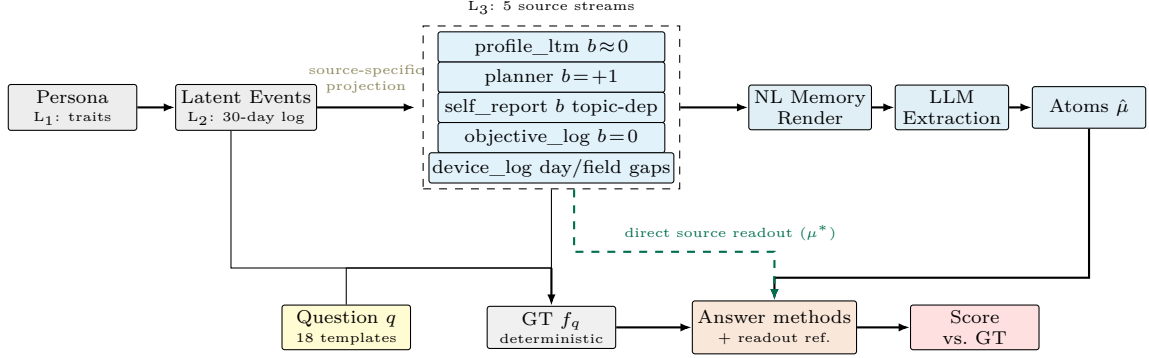
\begin{figure}[t]
\centering
\noindent\makebox[\linewidth][c]{%
\begin{tikzpicture}[
  every node/.style={font=\scriptsize},
  box/.style       ={draw, rounded corners=1pt, align=center, inner sep=2pt, minimum height=5.5mm},
  latent/.style    ={box, fill=tierZero!18, minimum width=17mm},
  src/.style       ={box, fill=tierOne!18,  minimum width=30mm, minimum height=4mm, inner sep=1.2pt},
  proc/.style      ={box, fill=tierTwo!12,  minimum width=15mm},
  oracle/.style    ={box, fill=tierFour!18, minimum width=17mm},
  gtbox/.style     ={box, fill=gray!12, minimum width=17mm},
  method/.style    ={box, fill=tierThree!15,minimum width=20mm, minimum height=7mm},
  arr/.style       ={-{Latex[length=1.5mm]}, thick},
  byparr/.style    ={-{Latex[length=1.5mm]}, thick, dashed, tierFour!70!black},
]

\node[latent]                          (persona) at (0, 0)
       {Persona\\[-1pt]{\tiny L$_1$: traits}};
\node[latent, right=5mm of persona]    (events)
       {Latent Events\\[-1pt]{\tiny L$_2$: 30-day log}};

\coordinate (srcAnchor) at ($(events.east) + (31mm, 0)$);
\node[src] (s1) at ($(srcAnchor) + (0, 8mm)$)  {profile\_ltm \hfill $b\!\approx\!0$};
\node[src] (s2) at ($(srcAnchor) + (0, 4mm)$)  {planner \hfill $b\!=\!{+}1$};
\node[src] (s3) at (srcAnchor)                 {self\_report \hfill $b$ topic-dep};
\node[src] (s4) at ($(srcAnchor) + (0,-4mm)$)  {objective\_log \hfill $b\!=\!0$};
\node[src] (s5) at ($(srcAnchor) + (0,-8mm)$)  {device\_log \hfill day/field gaps};
\node[draw, dashed, fit=(s1)(s5), inner sep=2pt] (sources) {};
\node[font=\tiny, anchor=south, yshift=0.3mm] at (sources.north) {L$_3$: 5 source streams};

\node[proc, right=9mm of sources]      (nl)    {NL Memory\\Render};
\node[proc, right=3mm of nl]           (ext)   {LLM\\Extraction};
\node[proc, right=3mm of ext]          (muhat) {Atoms $\hat\mu$};

\draw[arr] (persona) -- (events);
\draw[arr] (events.east) -- ([xshift=-1mm]sources.west)
  node[midway, above=0.6mm, font=\tiny, align=center, text=yellow!45!black] {source-specific\\projection};
\draw[arr] (sources.east) -- (nl.west);
\draw[arr] (nl) -- (ext);
\draw[arr] (ext) -- (muhat);

\coordinate (bypassY) at ($(persona.south) + (0,-16mm)$);
\coordinate (row2Y)   at ($(persona.south) + (0,-26mm)$);

\node[gtbox] (gt) at (sources |- row2Y) {GT $f_q$\\[-1pt]{\tiny deterministic}};
\node[box, fill=yellow!22, minimum width=17mm, left=10mm of gt] (q)
       {Question $q$\\[-1pt]{\tiny 18 templates}};
\node[method, right=10mm of gt] (methods) {Answer methods\\[-1pt]{\tiny + readout ref.}};
\node[box, fill=red!12, minimum width=17mm, right=7mm of methods] (eval)
       {Score\\vs.\ GT};

\draw[arr] (gt) -- (methods);
\draw[arr] (methods) -- (eval);

\draw[arr] (muhat.south) |- ($(methods.north) + (0,3mm)$) -- (methods.north);

\coordinate (gtJ)      at ($(gt.north) + (0,5mm)$);
\coordinate (L2out)    at ($(events.south) + (-2mm,0)$);
\draw (q.north)  |- (gtJ);
\draw (L2out) -- (L2out |- gtJ) -- (gtJ);
\draw (sources.south) -- (gtJ);
\draw[arr] (gtJ) -- (gt.north);

\coordinate (bypOrig)  at ($(sources.south) + (3mm,0)$);
\coordinate (bypStart) at (bypOrig |- bypassY);
\coordinate (bypEnd)   at (methods.north |- bypassY);
\draw[byparr] (bypOrig) -- (bypStart) -- (bypEnd) -- (methods.north);
\node[font=\tiny, tierFour!70!black, anchor=south, yshift=0.3mm]
      at ($(bypStart)!0.78!(bypEnd)$) {direct source readout ($\mu^\ast$)};

\end{tikzpicture}%
}
\caption{\textbf{Data-generating process (DGP) as a DAG.} \emph{Top row, generation:} persona (L$_1$) $\to$ latent 30-day event log (L$_2$) $\to$ 5 source-specific streams (L$_3$) with schema filtering, bias, noise, and missingness $\to$ NL memory $\to$ LLM-extracted atoms $\hat\mu$. \emph{Bottom row, evaluation:} ground truth $f_q(L_2,L_3,q)$ is computed deterministically with no circular dependency on extraction. The answer methods consume $\hat\mu$ and are scored against GT. The dashed green arrow is the direct source readout: $\hat\mu$ is replaced by atoms $\mu^\ast$ read directly from the structured $L_3$ source streams, isolating NL extraction noise from all upstream $L_2\to L_3$ projection loss. This readout is used for matched input experiments and for the Source Reachability reference in \S~\ref{s4:sub1}. An end-to-end walked example tracing one \texttt{(persona, question)} through every stage of this DAG is in Appendix~\ref{app:walked}.}
\label{fig:1}
\end{figure}

Real personal-memory data with verifiable behavioral ground truth is hard to obtain, so we use a synthetic data-generating process (DGP) following the precedent of CLEVR \citep{johnson2017clevr} and bAbI \citep{weston2016babi}. The DGP, combined with paired extracted-atom and direct-readout conditions, supports controlled comparisons between resolver and input quality. The DGP runs in four stages (persona traits ($L_1$) $\to$ latent event table ($L_2$) $\to$ biased source projection ($L_3$) $\to$ natural-language (NL) memory render), then a shared LLM extraction step produces per-source structured atoms; per-stage details and per-stream bias parameters are in Appendix~\ref{subsec:B-1} and Appendix~\ref{subsec:B-3}. Personas are balanced across three difficulty classes (\texttt{stable} / \texttt{temporal\_shift} / \texttt{stated\_vs\_revealed}, 160 each), and instances are formed by crossing each persona with the 18 question templates. The released full testbed contains 18 questions $\times$ 480 personas $\times$ 4 seeds = 34,560 instances. Figure~\ref{fig:1} shows how the DGP separates latent behavior, visible source projections, extracted atoms, GT computation, and the direct-readout reference. This lets the evaluation vary atom source and resolver separately while preserving a DAG with no extraction-to-GT path.

Throughout the paper, GT means the deterministic label $f_q(L_2,L_3,q)$. An atom is the closed-class value, or null, that one evidence stream provides for one question. We use $\hat\mu^m$ for atoms extracted from NL memory by extractor model $m$, and $\mu^\ast$ for atoms read directly from the structured $L_3$ source streams, with no NL rendering or LLM extraction. Unless stated otherwise, $\hat\mu$ means $\hat\mu^{\text{GPT}}$, the shared GPT-5.4 extraction used in the main tables; Appendix~\ref{subsec:F-6} repeats extraction with Gemini 3.1 Pro. Results on $\hat\mu$ and $\mu^\ast$ keep the resolver fixed and change only the atom source. When later tables say GPT-5.4 on $\mu^\ast$, the LLM is reading those direct-readout atoms as structured prompt input. Source Reachability is a GT-aided reference: it asks whether at least one direct-readout source atom exactly equals GT. We report it only as a diagnostic reference, not as a resolver or skip policy.

Figure~\ref{fig:1} separates latent behavior from the memory evidence available to methods. The top path shows what methods observe: the $L_2$ event table is transformed into five source-specific $L_3$ projections: an early-window or stated profile summary, aspirational plans, biased self-reports, sparse objective records, and device traces with day- and field-level missingness. These structured projections are rendered as NL memory and extracted into atoms $\hat\mu$. The bottom path shows GT and scoring: all GT labels depend on $q$ and $L_2$, and 9/18 templates also read $L_3$ annotations such as self-report labels or planner commitments; non-skipped answers are scored against GT. The dashed green path is the direct-readout reference path: it replaces NL extraction with $\mu^\ast$ from structured $L_3$. This removes NL extraction noise but does not read $L_2$ directly, so $L_2 \to L_3$ projection loss remains.

\subsection{Ground Truth, Splits, and Protocol}\label{s2:sub2}

Ground-truth labels are deterministic. The label rules use window selection, counting, ratios, thresholds, discretization, and categorical checks, with no ML or probabilistic component. Full per-template label functions are listed in Appendix~\ref{subsec:B-4}. The pipeline forms a DAG: $L_2$ generates $L_3$, $(q,L_2,L_3)$ feeds GT, and extraction never feeds GT. An internal independent audit implementation recovers all 34,560 labels exactly (Appendix~\ref{subsec:B-5}).

Per seed, 480 personas are split into 216 train / 48 dev / 96 calibration / 120 test, stratified by difficulty class. Train fits learnable parameters when a method has them. Calibration tunes SKIP thresholds for threshold-based selective variants via $F_{0.5}$ maximization; $F_{0.5}$ serves only as the threshold-selection criterion. We use $\beta = 0.5$ because wrong answers are costlier than abstentions. Dev supports development-time checks and tuning, and test is held out for reporting. All fitted parameters and SKIP thresholds are learned within each seed. Main test results pool the four 120-persona test splits, giving 480 seed-persona clusters and 8,640 question instances, and report 95\% bootstrap percentile CIs ($B = 2000$ resamples over seed-persona clusters).

\section{Evaluated Methods}\label{sec:methods}
We evaluate resolvers in four tiers and include the Source Reachability reference (Table~\ref{tab:3}); each tier isolates a specific modeling assumption and exposes a diagnostic contrast. We use $\hat\mu(s, q)$ as shorthand for the default GPT-5.4-extracted atom from source $s$ for question $q$ (a value from $\mathcal{V}_q$, or null), and $\mu^\ast(s, q)$ for the same slot read directly from the structured $L_3$ source streams. Methods marked $\pm$ in the Selective column have both answer-only and selective variants with SKIP; we report both in \S~\ref{sec:experiments}.

\begingroup\small
\begin{xltabular}{\linewidth}{@{}llcYc@{}}
\caption{Evaluated methods organized by assumption tier. Input: $\hat\mu$ = GPT-5.4-extracted atoms from NL memory unless stated otherwise; NL = raw natural-language memory; $\mu^\ast$ = direct-readout atoms from the structured $L_3$ source streams.}\label{tab:3}\\
\toprule
 & Method & Input & Mechanism & Sel. \\
\midrule
\endfirsthead
\toprule
 & Method & Input & Mechanism & Sel. \\
\midrule
\endhead
\bottomrule
\endfoot
\textbf{T0 $\cdot$ Source-Free} &  &  &  &  \\
 & Random & — & Uniform draw from $\mathcal{V}_q$ & — \\
 & Majority Class & — & Per-question mode from training labels & — \\
\textbf{T1 $\cdot$ Source Selection} &  &  &  &  \\
 & SSB & $\hat\mu$ & Best single source per question (train-selected)$\dagger$ & $\pm$ \\
\textbf{T2 $\cdot$ Multi-Source Fusion} &  &  &  &  \\
 & Majority Vote & $\hat\mu$ & Plurality vote across non-null sources & — \\
 & ArgRAG-style & $\hat\mu$ & Equal-weight support/attack adaptation$\ddagger$ & — \\
 & BCF & $\hat\mu$ & Forward bias model with per-source deflation weights & — \\
 & NBF & $\hat\mu$ & Bayesian posterior from per-source confusion matrices & $\pm$ \\
 & DSNBF & $\hat\mu$ & NBF with difficulty-stratified confusion matrices & $\pm$ \\
 & ABF & $\hat\mu$ & Soft-mixture abductive scoring (bias + identity paths) & $\pm$ \\
\textbf{T3 $\cdot$ End-to-End LLM} &  &  &  &  \\
 & LLM Direct & NL & LLM reads all 5 NL source documents, answers directly & $\pm$ \\
 & Schema Aware & NL & LLM Direct with source bias hints and reliability guidance & $\pm$ \\
\textbf{Ref. $\cdot$ Source Reachability} &  &  &  &  \\
 & Source Reachability & $\mu^\ast$ & Direct-readout reference; not deployable & ref. \\
\multicolumn{5}{@{}p{\linewidth}@{}}{\footnotesize $\dagger$ SSB requires source ranking on the train split but has no learned continuous parameters. $\ddagger$ Our ArgRAG adaptation \citep{zhu2025argrag} degenerates to Majority Vote under our five-source closed-class atom representation (Appendix~\ref{subsec:C-argrag}).}\\
\end{xltabular}
\endgroup

T0 provides source-free baselines, and T1 gives a single-source selection baseline. T2 fusion methods use structured multi-source signals: BCF applies a hand-specified ordinal source-bias prior with learned deflation weights, ABF uses the same prior inside a soft-mixture abductive kernel, and NBF/DSNBF learn empirical per-source confusion matrices from training labels. Four derived questions use per-question bias-prior overrides where the topic-level prior would invert the source semantics; Appendix Table~\ref{tab:C-bias-prior} gives the fixed prior and override resolution rule. Although DSNBF has difficulty-stratified parameters, it is not given the held-out test-set difficulty mixture or any test persona's true difficulty label; at test time it infers a per-persona difficulty posterior from that persona's observed atoms only, and true difficulty labels are used only after prediction for reporting breakdowns. T3 LLM methods bypass the extraction pipeline entirely and read raw NL memory. Schema-Aware extends LLM-Direct by injecting qualitative source-bias descriptions into the prompt (e.g., "the planner tends to overstate intentions; the self-report overreports sleep and exercise but underreports work hours"), supplying the LLM with the same kind of domain knowledge that fusion methods encode through learned parameters but without access to test data. Source Reachability is a GT-aided reference: it reads $\mu^\ast$ directly from structured $L_3$, preserves all $L_2 \to L_3$ projection artefacts, and asks whether GT is directly present in the visible source projections. Its 93.2\% / 6.8\% interpretation is in \S~\ref{s4:sub1}. Selective variants abstain via posterior margin (NBF/DSNBF), two-criterion thresholds on absolute and relative evidence (ABF), null fallback (SSB), or self-reported \texttt{would\_skip} (LLMs); SKIP thresholds for threshold-based variants are calibrated on a held-out split via $F_{0.5}$ maximization (Appendix~\ref{subsec:C-2}).

All T1/T2 methods share an LLM extraction stage that converts each NL memory document into structured atoms $\hat\mu(s, q)$; we use GPT-5.4 \citep{openai2026gpt54} at maximum reasoning effort, with Gemini 3.1 Pro Preview as a cross-extractor robustness check (Appendix~\ref{subsec:F-6}). Frozen no-API T1/T2 rows fit/calibrate on $\mu^\ast$ train/cal atoms and test on frozen GPT-5.4 $\hat\mu$ atoms; Appendix~\ref{subsec:C-3} gives cache details. T3 evaluates four LLM families (GPT-5.4, Gemini 3.1 Pro Preview, DeepSeek V3.2, Qwen3-235B-A22B-Instruct-2507, reported as Qwen3 235B in tables) under identical prompt templates with no per-family tuning. The 2$\times$2 crossing of resolver (DSNBF vs. GPT-5.4) $\times$ input quality ($\hat\mu$ vs.\ $\mu^\ast$) is performed on GPT-5.4 (\S~\ref{s4:sub2}); full prompt templates, schema-aware bias descriptions, and per-LLM configurations are in Appendix~\ref{subsec:C-3}.

Appendix~\ref{app:walked} traces one persona end-to-end through five-source extraction, conflict structure, and a cross-tier subset of the method predictions discussed above.

\section{Experiments}\label{sec:experiments}
\subsection{Main Results and Selective QA}\label{s4:sub1}

\begin{figure}[t]
\centering
\scriptsize
\noindent\makebox[\linewidth][c]{%
\begin{tikzpicture}[
  every node/.style={font=\scriptsize},
  ci/.style={line width=0.7pt},
  pt/.style={circle, draw=black, fill=black, inner sep=0pt, minimum size=3.0pt},
  llmpt/.style={font=\tiny, text=tierThree!90!black, inner sep=0pt},
  refline/.style={dashed, line width=0.5pt},
]

\def\xmin{25}
\def\xmax{95}
\def\xscale{0.19} %

\newcommand{\fpoint}[5]{%
  \draw[ci] ({(#2-\xmin)*\xscale}, #1) -- ({(#4-\xmin)*\xscale}, #1);
  \draw[ci] ({(#2-\xmin)*\xscale}, #1+0.05) -- ({(#2-\xmin)*\xscale}, #1-0.05);
  \draw[ci] ({(#4-\xmin)*\xscale}, #1+0.05) -- ({(#4-\xmin)*\xscale}, #1-0.05);
  \node[pt] at ({(#3-\xmin)*\xscale}, #1) {};
  \node[anchor=east] at (-0.15, #1) {#5};
  \node[anchor=west, font=\tiny] at ({(#4-\xmin)*\xscale + 0.08}, #1) {#3};
}
\newcommand{\lpair}[8]{%
  \draw[ci, gray!70] ({(#2-\xmin)*\xscale}, #1) -- ({(#4-\xmin)*\xscale}, #1);
  \draw[ci, gray!70] ({(#2-\xmin)*\xscale}, #1+0.05) -- ({(#2-\xmin)*\xscale}, #1-0.05);
  \draw[ci, gray!70] ({(#4-\xmin)*\xscale}, #1+0.05) -- ({(#4-\xmin)*\xscale}, #1-0.05);
  \node[llmpt] at ({(#3-\xmin)*\xscale}, #1) {D};
  \draw[ci, gray!70] ({(#5-\xmin)*\xscale}, #1) -- ({(#7-\xmin)*\xscale}, #1);
  \draw[ci, gray!70] ({(#5-\xmin)*\xscale}, #1+0.05) -- ({(#5-\xmin)*\xscale}, #1-0.05);
  \draw[ci, gray!70] ({(#7-\xmin)*\xscale}, #1+0.05) -- ({(#7-\xmin)*\xscale}, #1-0.05);
  \node[llmpt] at ({(#6-\xmin)*\xscale}, #1) {S};
  \node[anchor=east] at (-0.15, #1) {#8};
}

\draw[refline, gray] ({(70.0-\xmin)*\xscale}, -0.3) -- ({(70.0-\xmin)*\xscale}, 4.55)
  node[above, font=\tiny, gray] {Best LLM 70.0};
\draw[refline, gray] ({(82.3-\xmin)*\xscale}, -0.3) -- ({(82.3-\xmin)*\xscale}, 4.55)
  node[above, font=\tiny, gray] {DSNBF on $\mu^\ast$ 82.3};
	
\lpair{0.00}{41.1}{42.1}{43.0}{47.1}{48.0}{48.8}{Qwen3 (T3)}
\lpair{0.35}{55.0}{56.1}{57.1}{55.9}{56.9}{57.9}{DeepSeek (T3)}
\lpair{0.70}{67.6}{68.6}{69.6}{68.6}{69.7}{70.7}{GPT-5.4 (T3)}
\lpair{1.05}{66.8}{67.8}{68.7}{69.0}{70.0}{71.1}{Gemini (T3)}
	
\fpoint{1.60}{29.1}{30.1}{31.0}{Random (T0)}
\fpoint{1.95}{55.9}{57.1}{58.4}{Majority Class (T0)}
\fpoint{2.30}{67.6}{68.8}{69.9}{MV / ArgRAG-style (T2)}
\fpoint{2.65}{68.0}{69.2}{70.3}{BCF (T2)}
\fpoint{3.00}{70.9}{72.0}{72.9}{ABF (T2)}
\fpoint{3.35}{76.4}{77.3}{78.1}{SSB (T1)}
\fpoint{3.70}{78.9}{79.8}{80.6}{NBF (T2)}
\fpoint{4.05}{79.5}{80.3}{81.2}{DSNBF (T2)}

\draw[dotted, gray!60] (-2.3, 1.32) -- ({(\xmax-\xmin)*\xscale + 0.3}, 1.32);

\draw[->, line width=0.4pt] (0, -0.3) -- ({(\xmax-\xmin)*\xscale + 0.3}, -0.3);
\foreach \x in {30, 40, 50, 60, 70, 80, 90} {
  \draw ({(\x-\xmin)*\xscale}, -0.3) -- ({(\x-\xmin)*\xscale}, -0.4);
  \node[anchor=north, font=\tiny] at ({(\x-\xmin)*\xscale}, -0.4) {\x};
}
\node[anchor=north, font=\scriptsize] at ({(\xmax-\xmin)*\xscale/2}, -0.85) {Macro accuracy (\%, answer-only mode, 4-seed pooled)};

\end{tikzpicture}%
}
\caption{\textbf{Answer-only macro accuracy.} Points show the main comparison with 95\% persona-level bootstrap CIs. Filled circles: T0/T1/T2 methods. For each T3 LLM family, D and S mark Direct and Schema-Aware on the same row. Dashed lines mark the strongest LLM (70.0\%) and DSNBF on $\mu^\ast$ (82.3\%).}
\label{fig:2}
\end{figure}
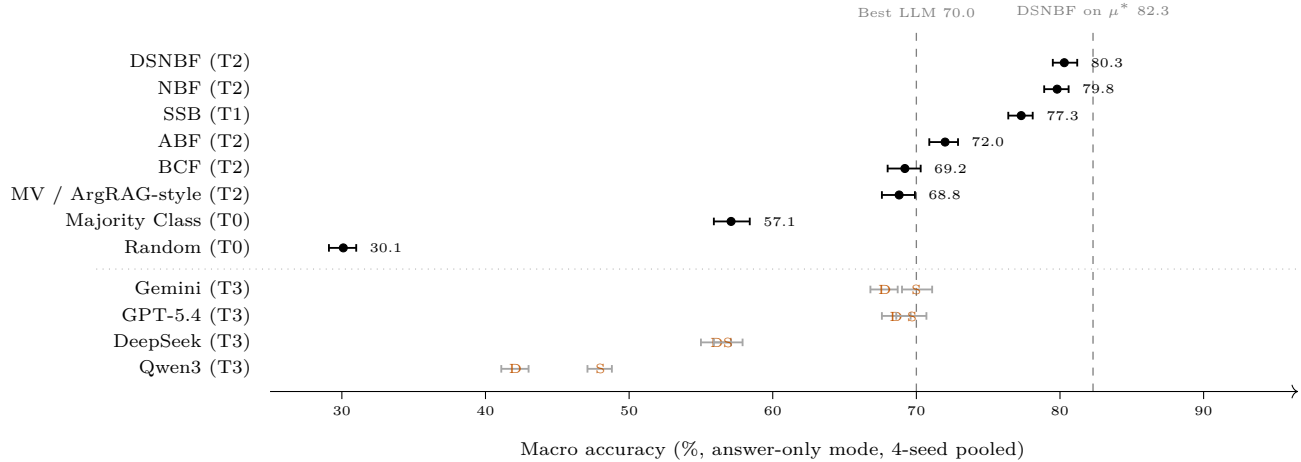

Table~\ref{tab:4} and Figure~\ref{fig:2} report macro accuracy (per-question accuracy averaged over 18 questions) under each method's canonical information condition. We pool four held-out test splits (480 seed-persona clusters; 8,640 question instances) and report 95\% bootstrap percentile CIs ($B = 2{,}000$).

\begin{table}[t]
\centering
\caption{Macro accuracy (\%, answer-only mode), 4-seed pooled; metric definitions are in Appendix~\ref{subsec:C-metrics}. T1/T2 $\hat\mu$ rows use GPT-5.4 test extractions after $\mu^\ast$ train/cal fitting (Appendix~\ref{subsec:C-3}); T3 reads NL memory (Direct / Schema, $\Delta$). Last column uses direct-readout $\mu^\ast$. T0 baselines: Random 30.1\%, Majority Class 57.1\%. For compactness, T3 CIs are plotted in Figure~\ref{fig:2}. ArgRAG-style denotes our equal-weight closed-class adaptation (Appendix~\ref{subsec:C-argrag}); Source Reachability is the direct-readout reference.}
\label{tab:4}
\small
\begin{tabular}{@{}llccc@{}}
\toprule
 & Method & Acc (\%) & 95\% CI & Acc on $\mu^\ast$ (\%) \\
\midrule
\textbf{T1} & SSB & 77.3 & [76.4, 78.1] & 79.0 \\
\textbf{T2} & Majority Vote / ArgRAG-style & 68.8 & [67.6, 69.9] & 69.5 \\
 & BCF & 69.2 & [68.0, 70.3] & 69.8 \\
 & NBF & \underline{79.8} & [78.9, 80.6] & \underline{82.1} \\
 & DSNBF & \textbf{80.3} & \textbf{[79.5, 81.2]} & \textbf{82.3} \\
 & ABF & 72.0 & [70.9, 72.9] & 72.0 \\
\textbf{T3} & GPT-5.4 (Direct / Schema, $\Delta$=+1.1) & 68.6 / 69.7 & — & — \\
 & Gemini 3.1 Pro (Direct / Schema, $\Delta$=+2.2) & 67.8 / 70.0 & — & — \\
 & DeepSeek V3.2 (Direct / Schema, $\Delta$=+0.8) & 56.1 / 56.9 & — & — \\
 & Qwen3-235B (Direct / Schema, $\Delta$=+5.9) & 42.1 / 48.0 & — & — \\
\textbf{Ref.} & Source Reachability & — & — & 93.2 \\
\bottomrule
\end{tabular}
\end{table}

\noindent The direct-readout column needs a careful reading because it combines three granularities. At the answer level, DSNBF scores 80.3\% on extracted atoms $\hat\mu$ and 82.3\% on direct-readout atoms $\mu^\ast$, so the aggregate extraction effect is about 2.0 pp. At the cell level, the GPT-5.4 extractor matches direct readout in 93.08\% of \texttt{(persona, seed, question, source)} cells. At the instance level, Source Reachability is a GT-aided probe: 93.2\% of instances have some direct-readout source atom equal to GT. The complementary 6.8\% marks cases where GT is absent verbatim from every direct-readout source; methods may still infer GT from distorted projections. Appendix~\ref{subsec:B-4} separates this reachability effect from extraction-path effects.

The main results suggest three things. First, naive aggregation can be worse than choosing a single reliable source: SSB (77.3\%) beats Majority Vote and BCF (both near 69\%) because, without bias correction, a biased majority can outvote an accurate minority. Second, methods that learn per-source confusion matrices pull clearly ahead. NBF and DSNBF reach about 80\%, and DSNBF leads the strongest prompt-only LLM baseline by 10.3 pp with non-overlapping CIs; the same lead holds against every LLM family, with a CI lower bound of at least 8.4 pp. Third, qualitative bias hints are a much weaker substitute for labeled training: Schema-Aware prompting buys an LLM 0.8--5.9 pp over Direct, while DSNBF leads Direct prompting by 11.7--38.2 pp and remains 10.3--32.3 pp above Schema-Aware prompting. A prompt that names each source's bias direction therefore carries much less of the source-to-GT mapping than a confusion matrix estimated from labels.

\begin{figure}[t]
\centering
\noindent\makebox[\linewidth][c]{%
\begin{tikzpicture}[
  every node/.style={font=\scriptsize},
  axisl/.style={font=\scriptsize},
  m1/.style={draw=tierOne!80!black,   fill=tierOne!50,   rectangle, minimum size=2.6mm, inner sep=0pt},
  m2/.style={draw=tierTwo!85!black,   fill=tierTwo!70,   circle,    minimum size=3.2mm, inner sep=0pt, line width=0.7pt},
	  m3/.style={draw=tierThree!90!black, fill=white, regular polygon, regular polygon sides=3, minimum size=4.6mm, inner sep=0pt, line width=0.8pt, font=\tiny, text=tierThree!90!black},
	  pareto/.style={dashed, gray!60!black, thick},
  pairlink/.style={gray!50, thin},
]

\def\px#1{\fpeval{(#1 - 60) * 0.18}}
\def\py#1{\fpeval{(#1 - 40) * 0.09}}

\draw[->, thick] (\px{60},\py{40}) -- (\px{103},\py{40}) node[right, font=\scriptsize] {coverage (\%)};
	\draw[->, thick] (\px{60},\py{40}) -- (\px{60},\py{91.5}) node[above, font=\scriptsize] {sel.\ acc.\ (\%)};

\foreach \x in {60,70,80,90,100} {
	  \draw[gray!25] (\px{\x},\py{40}) -- (\px{\x},\py{92});
  \node[axisl, below] at (\px{\x},\py{40}) {\x};
}
	\foreach \y in {40,50,60,70,80,90} {
	  \draw[gray!25] (\px{60},\py{\y}) -- (\px{102},\py{\y});
	  \node[axisl, left] at (\px{60},\py{\y}) {\y};
	}

\draw[pareto]
     (\px{77.5},\py{85.9})    %
  -- (\px{78.3},\py{85.3})    %
  -- (\px{82.9},\py{80.8})    %
  -- (\px{95.4},\py{71.0})    %
  -- (\px{99.2},\py{70.4});   %

\node[m2] (nbf)   at (\px{77.5},\py{85.9}) {};
\node[m2] (dsnbf) at (\px{78.3},\py{85.3}) {};
\node[m2] (abf)   at (\px{62.0},\py{83.0}) {};
\node[font=\tiny, anchor=south, yshift=0.6mm]                  at (nbf)   {NBF};
\node[font=\tiny, anchor=north west, xshift=0.7mm, yshift=-0.2mm] at (dsnbf) {DSNBF};
\node[font=\tiny, anchor=west, xshift=0.8mm] at (abf)   {ABF};

\node[m1] (ssb) at (\px{82.9},\py{80.8}) {};
\node[font=\tiny, anchor=west, xshift=0.6mm] at (ssb) {SSB};

\node[m3] (gptD) at (\px{92.3},\py{70.3}) {D};
\node[m3] (gptS) at (\px{95.4},\py{71.0}) {S};
\draw[pairlink] (gptD) -- (gptS);
\node[font=\tiny, anchor=north, yshift=-0.6mm] at ($(gptD)!0.5!(gptS)$) {GPT-5.4};

\node[m3] (gemD) at (\px{98.5},\py{68.4}) {D};
\node[m3] (gemS) at (\px{99.2},\py{70.4}) {S};
\draw[pairlink] (gemD) -- (gemS);
\node[font=\tiny, anchor=west, xshift=0.8mm] at ($(gemD)!0.5!(gemS)$) {Gemini};

\node[m3] (dsD)  at (\px{100.0},\py{56.1}) {D};
\node[m3] (dsS)  at (\px{100.0},\py{56.9}) {S};
\draw[pairlink] (dsD) -- (dsS);
\node[font=\tiny, anchor=west, xshift=0.8mm] at ($(dsD)!0.5!(dsS)$) {DeepSeek};

\node[m3] (qwD)  at (\px{99.3},\py{42.2}) {D};
\node[m3] (qwS)  at (\px{98.8},\py{48.0}) {S};
\draw[pairlink] (qwD) -- (qwS);
\node[font=\tiny, anchor=east, xshift=-0.8mm] at ($(qwD)!0.5!(qwS)$) {Qwen3};

	\end{tikzpicture}%
	}
			\caption{\textbf{Selective QA: coverage vs.\ selective accuracy.} Coverage is the answered fraction; top-right is better. Square = SSB, circles = fusion, triangles = LLM self-reported SKIP. Dashed grey segments mark Pareto-optimal deployable methods. Values are in Appendix~\ref{subsec:D-1}.}
	\label{fig:3}
	\end{figure}
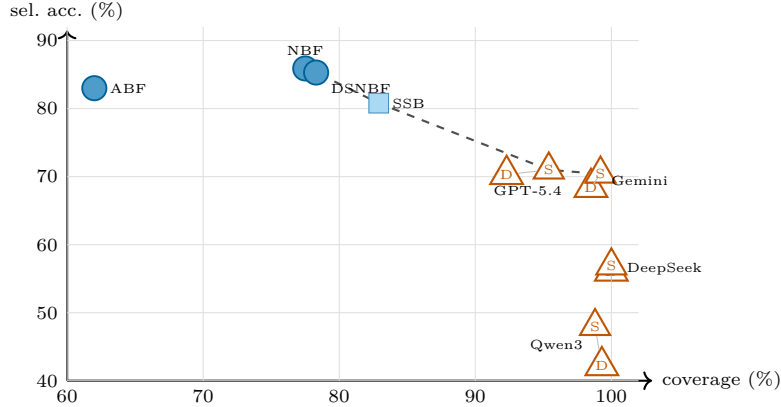

Selective QA asks how accurate a method is on the questions it chooses to answer; coverage is that answered fraction. DSNBF answers 78.3\% of the testbed and reaches 85.3\% selective accuracy, climbing to 88.8\% on direct-readout input. Calibration transfers cleanly to the held-out test split, with test $F_{0.5}$ between 80.4 and 85.3 across the four threshold-based fusion methods. LLM self-reported abstention behaves differently: selective-vs-answer-only gains stay under 2 pp, GPT-5.4 abstains on roughly 5--8\% of items, Gemini and Qwen3 stay below 2\%, and DeepSeek never abstains. Figure~\ref{fig:3} plots deployable SKIP variants. The split is partly an interface property: fusion exposes calibrated posterior margins, while a prompt-only LLM returns a binary flag with no continuous score for post-hoc calibration. Full selective and few-shot results are in Appendix~\ref{subsec:D-1} and \ref{subsec:D-3}.

\subsection{Factorial Decomposition (Resolver \texorpdfstring{$\times$}{x} Input)}\label{s4:sub2}

The testbed's factorial design enables a controlled decomposition that requires $\mu^\ast$ atoms, which are not normally available in naturalistic data without manual annotation or instrumentation. We cross a trained resolver (DSNBF-NoSkip vs. GPT-5.4) with input quality ($\hat\mu$ vs.\ $\mu^\ast$); to isolate the resolver gap from the extraction gap, we provide the LLM with the same structured atoms that fusion methods receive, removing NL comprehension as a confound.

The four cells are DSNBF on direct-readout atoms $\mu^\ast$ (82.3\%), GPT-5.4 on the same $\mu^\ast$ atoms (73.5\%), DSNBF on extracted atoms $\hat\mu$ (80.3\%), and GPT-5.4 on extracted atoms $\hat\mu$ (71.4\%). The diagonal contrast from DSNBF/$\mu^\ast$ to GPT/$\hat\mu$ is 10.9 pp, and a difference-of-differences attribution assigns about 81\% to the resolver and 19\% to the input. The interaction term is essentially zero, so the two effects are close to additive in this comparison. We read the 81/19 split as an evaluation-design result: DSNBF was trained on labeled personas, while GPT-5.4 was prompted without that training distribution. Two checks outside the 2$\times$2 point in the same direction: qualitative schema descriptions buy GPT only +1.1 pp, with overlapping CIs in Figure~\ref{fig:2}, and a seed-1 few-shot variant lifts GPT Direct by +6.5 pp but still trails DSNBF (Appendix~\ref{subsec:D-3}). Enriching the LLM's input narrows the gap to fusion but does not close it.

\subsection{Diagnostic Analysis by Reasoning Type}\label{s4:sub3}

The testbed's eight reasoning types serve as diagnostic probes (\S~\ref{sec:benchmark}). Table~\ref{tab:6} reports accuracy by type under matched $\mu^\ast$ input ($\Delta_{\text{matched}}$ = DSNBF on $\mu^\ast$ $-$ GPT-$\mu^\ast$, isolating the matched-input DSNBF-versus-GPT gap), alongside the strongest LLM and the Source Reachability reference by type.

\begin{table}[t]
\centering
\caption{Accuracy by reasoning type (\%, answer-only mode, 4-seed pooled). DSNBF and GPT-$\mu^\ast$ both read $\mu^\ast$ atoms; GPT-$\mu^\ast$ is GPT-5.4 reading them as structured prompt input. $\Delta_{\text{matched}}$ = DSNBF $-$ GPT-$\mu^\ast$, the matched-input DSNBF-versus-GPT gap. Best LLM = highest accuracy among 4 LLM families $\times$ 2 prompting regimes, each reading NL text. Ref. = Source Reachability from the source projections. Types ordered by DSNBF accuracy descending.}
\label{tab:6}
\small
\setlength{\tabcolsep}{3pt}
\begin{tabular}{@{}lccccclc@{}}
\toprule
Type & \# Q & DSNBF & GPT-$\mu^\ast$ & $\Delta_{\text{matched}}$ & 95\% CI & Best LLM & Ref. \\
\midrule
B $\cdot$ Ident & 2 & 96.8 & 97.0 & $-$0.2 & [$-$1.0, +0.6] & 94.5 (GPT-S) & 99.7 \\
A $\cdot$ Arbit & 3 & \textbf{87.5} & 68.5 & $+$19.0 & [17.0, 21.2] & 75.1 (Gem-S) & 94.1 \\
Ctrl & 2 & 85.7 & 75.7 & $+$10.0 & [6.9, 13.0] & 79.1 (GPT-D) & 96.0 \\
E $\cdot$ Factor & 2 & \textbf{83.1} & 70.6 & $+$12.5 & [9.6, 15.5] & 70.9 (GPT-S) & 95.0 \\
D $\cdot$ Temp & 2 & \textbf{82.2} & 76.2 & $+$5.9 & [3.4, 8.4] & 72.5 (GPT-D) & 90.0 \\
C $\cdot$ P-R & 2 & 81.6 & 73.4 & $+$8.1 & [4.9, 11.4] & 69.8 (GPT-D) & 94.8 \\
F $\cdot$ Miss & 3 & 77.9 & 71.5 & $+$6.5 & [4.0, 8.8] & 68.5 (GPT-S) & 92.0 \\
G $\cdot$ Annot & 2 & 63.5 & 58.5 & $+$5.0 & [0.8, 9.2] & 49.9 (Gem-S) & 83.7 \\
\bottomrule
\end{tabular}
\end{table}

At the type level, the results suggest three things. First, DSNBF leads on 7/8 types, with the largest matched-input gaps on Arbitration ($+$19.0 pp [17.0, 21.2]) and Factor attribution ($+$12.5 pp [9.6, 15.5]); Identity is the exception ($-$0.2 pp), with both methods near the Source Reachability reference ($\geq$96.8\%). This is directionally consistent with \S~\ref{s4:sub2}. Second, Annotation (G) is the hardest type because the latent attribute lies behind the source projections and must be inferred indirectly. DSNBF reaches 63.5\%, the best LLM 49.9\%, and the reference 83.7\%, suggesting that learned source-GT mappings help more than qualitative bias hints. Third, model families fail in different places, with stable type-specific signatures. GPT-5.4 and Gemini split by type: under Schema-Aware prompting, GPT-5.4 exceeds Gemini by 13.1 pp on Identity, while Gemini exceeds GPT-5.4 by 15.5 pp on Annotation (Appendix~\ref{subsec:E-1}). Full failure distributions and seed-stable signatures are in Appendix~\ref{subsec:E-3}; the bias-modeling gradient by difficulty class is in Appendix~\ref{subsec:E-2}.

\subsection{Robustness Summary}\label{s4:sub4}

Robustness checks reduce the concern that fusion only fits one fixed synthetic setting. NBF and DSNBF estimate source-to-GT confusion matrices from labeled train personas and predict from the observed atom table (whether $\hat\mu$ or $\mu^\ast$; Appendix~\ref{subsec:C-1}). Across seeds, fusion $\sigma \leq$0.8 pp; training-size curves saturate with data; rankings stay stable across nine DGP bias$\times$dropout variants (mean Kendall $\tau = 0.77$); swapping GPT for Gemini extraction costs DSNBF only 0.7 pp downstream; and DSNBF trained only on the default projection setting stays within 1.6 pp of DSNBF refit on each shifted variant on average (Appendix~\ref{subsec:F-1}, \ref{subsec:F-2}, \ref{subsec:F-4}, \ref{subsec:F-6}, \ref{subsec:F-7}). Taken together, these checks support the internal robustness claim: the main result holds across seeds, parameter settings, and extractor models. Real-user bias patterns remain an external-validity question.

\section{Discussion}\label{sec:discussion}
\phantomsection\label{s5:sub1}
The main result is that trained fusion methods are better at deciding which source to trust when sources disagree. In the matched-input comparison, GPT-5.4 and DSNBF receive the same structured atoms, but DSNBF still performs better, suggesting that the improvement is driven by the resolver rather than by the quality of data extraction from natural language.  In particular, prompt-only LLMs make a substantial part of their errors at the conflict-resolution step in this benchmark. 

The benchmark is useful because it makes these failure points observable. Source Reachability checks whether the answer was visible in {\it any} source. The extraction audit checks whether the LLM recovered the intended structured atoms from the natural-language memories. The resolver comparison then tests what happens after those atoms have been fixed. Selective QA adds a final check: whether a method can skip cases where the evidence is too weak. These three measurements are combined together into a single end-to-end accuracy metric. The released testbed also supports extensions with new questions or evidence streams.

\phantomsection\label{s5:sub2}
The benchmark has three main limitations. First, it is synthetic by design. Its source distortions are monotone and topic-dependent, while real personal data may include changing habits, changing sensor coverage, adversarial reports, or user-specific source biases. Second, the benchmark has 18 question templates, which limits within-type power even though the data-generating process can support more label functions. Third, the strongest fusion methods require labeled training personas, so their results should not be read as claims about cheaper deployment. Their value here is diagnostic: an LLM can extract source atoms, while a separate fusion layer resolves conflicts.

\phantomsection\label{s5:sub3}
A diagnostic benchmark for personal-memory agents can improve reliability by making conflict handling and abstention explicit before deployment. The same line of work also raises privacy risks if used to infer sensitive habits from personal records or to normalize surveillance style aggregation. We therefore release only synthetic data and cached artifacts. Real deployment would require user consent, data minimization, privacy controls, and validation on the target population.

\section{Conclusion}\label{sec:conclusion}
We introduced selective QA over conflicting multi-source personal memory: systems must reconcile biased evidence streams or abstain when evidence is insufficient. The core contribution is a diagnostic testbed that separates source coverage, extraction, resolver behavior, and abstention, so failures can be localized rather than collapsed into one leaderboard score.

The results show why this separation matters. Under the primary test condition, trained fusion reaches 80.3\% macro accuracy versus 70.0\% for the strongest prompt-only LLM baseline, and selective fusion reaches 85.3\% accuracy at 78.3\% coverage. We release data, code, cached artifacts, and extension examples for new questions, sources, and real-user validation; the reproduction repository is available at \url{https://github.com/TianchengY/multisource-membench}, and the dataset release is hosted at \url{https://huggingface.co/datasets/ytc1997/multisource-membench}.

\bibliographystyle{plainnat}
\bibliography{references}

\newpage
\appendix

\addtocontents{toc}{\protect\setcounter{tocdepth}{2}}
{\makeatletter\section*{Appendix Contents}\@starttoc{toc}\makeatother}

\section{Related Work}\label{app:related}
\subsection{Comparison Table with Existing Conflict-Related Benchmarks}\label{subsec:A-1}

Table~\ref{tab:A-1} summarises how our testbed compares to representative conflict-related and personal-memory benchmarks along five structural dimensions. The remaining subsections (\S~\ref{subsec:A-2}--\S~\ref{subsec:A-8}) discuss each line of literature in more depth, beginning with long-term memory benchmarks most likely to be confused with ours.

\begin{table}[ht]
\centering
\caption{Comparison with existing conflict-related benchmarks. "Typed evidence streams" = multiple named source or modality streams, not merely multiple retrieved passages. "Directed bias model" = sources have known, learnable distortion profiles. "Selective QA" = principled ANSWER/SKIP with calibrated thresholds; prompted refusal alone is marked separately. "Resolver $\times$ Input decomposition" = factorial design that isolates resolver from input quality.}
\label{tab:A-1}
\scriptsize
\setlength{\tabcolsep}{2pt}
\begin{tabularx}{\linewidth}{@{}>{\hsize=1.55\hsize}Y>{\hsize=0.95\hsize}Y>{\hsize=0.8\hsize}Y>{\hsize=1.05\hsize}Y>{\hsize=0.65\hsize}Y>{\hsize=1.0\hsize}Y@{}}
\toprule
Benchmark & Typed streams & Directed bias & Selective QA & Resolver $\times$ Input & Synthetic DGP \\
\midrule
Adaptive Chameleon / counter-memory setting \citep{xie2024conflictqa} & $\times$ & $\times$ & $\times$ (unknown filtered) & $\times$ & $\times$ \\
ECon \citep{econ2024} & $\times$ & $\times$ & $\times$ & $\times$ & partial (generated evidence conflicts) \\
NatConfQA \citep{nachshoni2025natconfqa} & $\times$ (passages) & $\times$ & $\times$ & $\times$ & $\times$ \\
QACC \citep{qacc2025} & $\times$ (passages) & $\times$ & $\times$ & $\times$ & $\times$ \\
ConflictQA \citep{zhao2026conflictqa} & $\checkmark$ (text + knowledge graph) & $\times$ & $\times$ & $\times$ & partial (generated conflicts) \\
LaMP \citep{salemi2024lamp} & $\times$ & $\times$ & $\times$ & $\times$ & $\times$ \\
LongMemEval \citep{wu2024longmemeval} & $\times$ (chat) & $\times$ & partial (abstention) & $\times$ & partial (edited chats) \\
LifeBench \citep{cheng2026lifebench} & $\checkmark$ (digital traces) & $\times$ & $\times$ (unanswerable labels) & $\times$ & $\checkmark$ \\
CrossCheck-Bench \citep{tian2026crosscheck} & $\checkmark$ (2 modal) & $\times$ & $\times$ & $\times$ & partial (controlled contradictions) \\
CLASH \citep{popordanoska2025clash} & $\checkmark$ (2 modal) & $\times$ & $\times$ & $\times$ & partial (controlled contradictions) \\
\textbf{Ours} & \textbf{$\checkmark$ (5 text)} & \textbf{$\checkmark$} & \textbf{$\checkmark$ (calibrated)} & \textbf{$\checkmark$} & \textbf{$\checkmark$} \\
\bottomrule
\end{tabularx}
\end{table}

\subsection{Long-Term Memory Benchmarks and Agent Memory}\label{subsec:A-2}\label{app:longmemeval}

The benchmark most likely to be confused with ours is concurrent LifeBench \citep{cheng2026lifebench}, which studies long-horizon memory over diverse digital traces such as chats, app records, and health records. Its central question is whether a memory system can retrieve and integrate information scattered across heterogeneous artifacts over time. Our benchmark targets source conflict: several streams can make incompatible claims about the same latent behavior. The system must learn source reliability and abstain when support is insufficient. LifeBench emphasizes source diversity; our benchmark isolates source disagreement.

This distinction changes the evaluation object. LifeBench primarily stresses memory retrieval, temporal integration, and non-declarative memory over a long horizon. We instead fix five typed personal-memory streams and make disagreement systematic: profile, planner, self-report, objective log, and device log can be stale, optimistic, distorted, incomplete, or unavailable in source-specific ways. This lets us measure conflict resolution directly through closed-class GT, calibrated selective QA, and a resolver-versus-input decomposition.

LongMemEval \citep{wu2024longmemeval} is complementary in another direction. It evaluates chat-assistant memory over coherent user--assistant histories, with questions testing information extraction, multi-session reasoning, temporal reasoning, knowledge updates, and abstention. Its own construction notes that sampled surrounding sessions ``avoid providing conflicting information that would invalidate the question'' \citep[Appendix A.2]{wu2024longmemeval}. LongMemEval therefore asks whether a system can recall from a coherent log; our testbed studies arbitration when source streams are intentionally inconsistent.

MemoryAgentBench \citep{hu2025memoryagentbench} evaluates memory agents through incremental multi-turn interactions, including selective forgetting when later evidence contradicts earlier facts. This is complementary to our setting: it tests whether an agent updates memory over time, whereas our benchmark fixes several typed personal-memory streams at once and evaluates source-specific conflict resolution, resolver-versus-input decomposition, and calibrated abstention.

Broader agent-memory systems make a similar distinction clear. MemGPT \citep{packer2024memgpt} introduces layered memory management, Generative Agents \citep{park2023generative} simulate daily-life memory with retrieval and reflection, and Reflexion \citep{shinn2023reflexion} uses memory for verbal self-improvement. These systems study how agents store, retrieve, and reuse memory. Our evaluation target is the separate problem of resolving conflicts across systematically biased personal-memory streams.

\subsection{Knowledge Conflicts}\label{subsec:A-3}

\citet{xu2024knowledge_conflicts} survey knowledge conflicts in LLMs, categorizing them as context-memory, inter-context, or intra-memory discrepancies. \citet{xie2024conflictqa} construct a controlled QA setting for parametric-memory versus counter-memory conflicts, while ECon \citep{econ2024} constructs a benchmark for evidence-conflict detection and resolution. \citet{chen2022confidence} study confidence calibration under knowledge conflicts, finding that QA models are poorly calibrated when retrieved passages disagree. NatConfQA \citep{nachshoni2025natconfqa} introduces fine-grained evaluation of conflict detection across retrieved documents, and QACC \citep{qacc2025} studies open-domain QA with conflicting contexts. Very recent ConflictQA \citep{zhao2026conflictqa} studies faithful reasoning when textual evidence and knowledge graph evidence conflict. \citet{shaier2024adaptiveqa} study QA with source citations in ambiguous settings where multiple valid answers may be supported by different sources.

ConflictQA is close to our setting because it also uses heterogeneous evidence sources, but its sources are external RAG artifacts rather than personal-memory streams. It does not model source-specific personal bias, calibrated selective QA, or a resolver-versus-input decomposition. More broadly, these settings use retrieved passages or contexts. Our testbed models named personal-memory streams with learnable, question-specific directional distortions and controlled source-disagreement profiles.

\subsection{RAG Conflict Resolution}\label{subsec:A-4}

\citet{gao2024rag} survey the Naive / Advanced / Modular RAG paradigms and their retrieval, generation, and augmentation components. Most standard RAG formulations treat retrieved passages as exchangeable evidence. Reliability-aware variants such as RA-RAG \citep{hwang2024rarag} estimate scalar source reliability, which differs from learning question-specific mappings from typed personal-memory streams to GT labels.

ArgRAG \citep{zhu2025argrag} applies argumentation graphs to resolve retrieved document contradictions. Under our closed-class, equal-weight adaptation, ArgRAG reduces to plurality voting; richer ArgRAG variants with document-specific strengths are outside this comparison. FactReasoner \citep{marinescu2025factreasoner} builds probabilistic factuality models for long-form generation, atomizing claims and scoring them against retrieved evidence. These methods target retrieved-evidence conflict or long-form factuality; our setting fixes typed, directionally biased streams and evaluates closed-class source-conflict resolution.

\subsection{Cross-Modal Conflict Benchmarks}\label{subsec:A-5}

CrossCheck-Bench \citep{tian2026crosscheck} evaluates vision-language models on compositional image-text contradictions and diagnoses failure modes across controlled difficulty tiers. CLASH \citep{popordanoska2025clash} pairs COCO images with captions containing controlled object-level and attribute-level contradictions, evaluated via multiple-choice and open-ended QA. These benchmarks address perceptual conflict detection between vision and language modalities; our work addresses conflicts among multiple textual evidence streams with controlled systematic biases.

\subsection{Selective Prediction and Abstention}\label{subsec:A-6}

\citet{geifman2017selective} formalize selective classification with coverage-accuracy trade-offs. \citet{kamath2020selective} apply it to QA. \citet{cole2023selective} study selective answering under question ambiguity using sampling-based confidence. \citet{wen2025abstention} survey \~{}100 abstention methods across the LLM lifecycle. Our testbed extends selective prediction to multi-source settings where evidence insufficiency is an explicit driver of abstention, alongside model uncertainty.

\subsection{Synthetic Benchmarks}\label{subsec:A-7}

Diagnostic benchmark design often uses synthetic or controlled tasks: CLUTRR \citep{sinha2019clutrr} for systematic generalization, bAbI \citep{weston2016babi} for reasoning primitives, and BigBench-Hard \citep{suzgun2023bigbench} for controlled difficulty. Our DGP follows this tradition, providing a fully documented generation process that supports documented extensions to new question types and sandboxed source configurations.

\subsection{Agent Evaluation Benchmarks}\label{subsec:A-8}

AgentBoard \citep{ma2024agentboard} and TaskBench \citep{shen2024taskbench} evaluate agent capabilities in tool-use and planning domains. InfiBench \citep{li2024infibench} evaluates code-LLM question answering, and AgentIF \citep{qin2025agentif} benchmarks instruction following in agentic scenarios (tool use, system prompts, multi-step plans). These benchmarks evaluate planning, tool use, and instruction following; our benchmark focuses on multi-source conflict resolution and selective abstention over personal memory.

\section{Benchmark Details and Ground-Truth Verification}\label{app:benchmark}
\subsection{Evidence Streams and Distortion Model}\label{subsec:B-1}

Table~\ref{tab:B-1} lists the five evidence streams and their distortion characteristics. We encode directional bias as $b$: $b = +1$ indicates systematic overreporting (the source reports more favorably than reality), $b = -1$ indicates systematic underreporting, and $b = 0$ indicates no directional bias.

\begin{table}[t]
\centering
\caption{Evidence streams and distortion model.}
\label{tab:B-1}
\small
\begin{tabularx}{\linewidth}{@{}lYYl@{}}
\toprule
Stream & Content & Distortion & Bias $b$ \\
\midrule
Long-term profile & Self-descriptions, habits, goals & Anchored to early-window mean; stale for shifting personas & $\approx 0$ \\
Planner & Daily intentions, schedules & Starts from habit parameters; optimistic & $+1$ \\
Self-report & Self-recorded events, explanations & Topic-dependent: $+1$ (sleep, diet, exercise), $-1$ (work, social) & varies \\
Objective log & Timestamps, purchase records & Small additive noise; most accurate & $0$ (fixed) \\
Device/activity log & Sleep, movement, and device-observed work-session signals & Day-level missingness plus field-level dropout and noise & $0$ \\
\bottomrule
\end{tabularx}
\end{table}

The self-report's directional bias pattern reflects a well-documented phenomenon in survey methodology: respondents tend to overreport socially desirable behaviors (sleep quality, healthy eating, exercise frequency) and underreport undesirable ones (overtime work, social isolation) \citep{paulhus2007self_report,brenner2014social_desirability}. The planner's positive bias mirrors the planning fallacy \citep{buehler1994planning_fallacy}, where intended behavior is systematically more favorable than realized behavior. The long-term profile's early-window anchoring reflects a design choice: the profile is generated from the persona's initial behavioral parameters and may not be updated promptly as behavior shifts. Device/activity-log missingness combines difficulty-dependent day-level dropout with field-level missingness, including an approximately 50\% dropout rate for the work-session field, to simulate non-wear, connectivity loss, battery limits, and partial sensor coverage \citep{choi2011accelerometer_nonwear}.

\paragraph{Bias-parameter grounding.} The literature above motivates the direction of each distortion, not an exact empirical effect-size calibration. We choose coarse bias signs ($-1,0,+1$), schema gaps, and missingness levels to create a controlled stress test in which source conflict is common enough to diagnose methods. The DGP perturbation grid in Appendix~\ref{subsec:F-4} then checks whether the method ordering is tied to one fixed bias magnitude.

\subsection{Question Specifications}\label{subsec:B-2}

\subsubsection{Full 18-Question Specification}

Table~\ref{tab:B-2} lists all 18 questions with topic, time window, ordered/categorical answer-space classification, and answer space. Table~\ref{tab:B-3} gives the full question text and targeted failure mode. The complete reasoning-type taxonomy and per-question design rationale are released in \texttt{configs/questions.yaml} in the code artifact.

\begingroup\small
\begin{xltabular}{\linewidth}{@{}lclccY@{}}
\caption{Question metadata and answer-space classification for all 18 templates. Reasoning type prefixes: A (Source Arbitration), B (Identity-Behavior), C (Plan-Reality), D (Trend), E (Factor-Attribution), F (Missing-Data), G (Annotation), Ctrl (Control). Type column reports answer-space classification (ord = ordinal with monotone bias direction; nom = unordered categorical). $K$ is the number of answer options.}\label{tab:B-2}\\
\toprule
Q & Type & Topic & $K$ & Win. (d) & Answer space (display order; "+" denotes edge option) \\
\midrule
\endfirsthead
\toprule
Q & Type & Topic & $K$ & Win. (d) & Answer space (display order; "+" denotes edge option) \\
\midrule
\endhead
\bottomrule
\endfoot
A1 & ord & Sleep & 3 & 30 & fewer\_than\_10 < 10\_to\_19 < 20\_or\_more nights with $\geq 7$h sleep \\
A2 & ord & Work & 3 & 30 & 0\_to\_3 < 4\_to\_7 < 8\_or\_more days worked $> 9$h \\
A3 & ord & Diet & 3 & 30 & less\_than\_40 < 40\_to\_69 < 70\_or\_more percent home-cooked meals \\
B2 & ord & Exercise & 4 & 30 & more\_than\_1\_below < within\_1\_day < more\_than\_1\_above (vs. profile freq); + no\_frequency\_described \\
B3 & nom & Work & 3 & 30 & matches / does\_not\_match / no\_approach\_described (weekend work vs. profile) \\
C2 & ord & Social & 4 & 14 & below\_25\_pct < 25\_to\_50\_pct < above\_50\_pct of social plans realized; + no\_plans \\
C3 & nom & Sleep & 4 & 14 & within\_20min\_more\_than\_50pct / later\_more\_than\_50pct / earlier\_more\_than\_50pct / no\_targets \\
D1 & ord & Social & 3 & 30 & decreased < stayed\_same < increased social activity rate (first 14d vs. last 16d) \\
D2 & ord & Diet & 3 & 30 & within\_1 < differs\_more\_than\_1 meal/day vs. profile baseline; + no\_baseline \\
E1 & nom & Sleep & 4 & 30 & work\_activity / social\_activity / no\_single\_factor / no\_late\_nights (factor on post-midnight bedtimes) \\
E2 & ord & Exercise & 3 & 30 & no\_fewer\_than\_30 < between\_30\_60 < yes\_more\_than\_60 percent of skipped-exercise days had $> 8.5$h work \\
F1 & ord & Social & 4 & 30 & 0\_to\_3 < 4\_to\_6 < 7\_or\_more unplanned social days; + no\_social\_activities \\
F2 & ord & Exercise & 3 & 30 & inactive\_confirmed < both\_occurred < yes\_tracker\_missing (on no-workout-recorded days) \\
F3 & ord & Work & 3 & 30 & truly\_off < both\_occurred < yes\_worked\_despite\_no\_entry (on missing-timesheet days) \\
G1 & ord & Exercise & 4 & 30 & incidental\_movement\_70plus < mix < deliberate\_exercise\_70plus (exercise intent share); + no\_activity \\
G2 & ord & Social & 4 & 30 & obligatory\_70plus < mix < voluntary\_70plus (social motivation share); + no\_meetings \\
Ctrl1 & ord & Diet & 3 & 7 & 0\_to\_1\_days < 2\_to\_3\_days < 4\_or\_more meals eaten outside per week \\
Ctrl2 & ord & Sleep & 3 & 7 & 0\_nights < 1\_to\_2 < 3\_or\_more nights with sleep $< 6$h \\
\end{xltabular}
\endgroup

Composition: 15 ordinal (with monotone bias direction along the answer scale) and 3 unordered categorical templates (B3, C3, E1; bidirectional or categorical without natural ordering). Topic distribution: Work (3), Diet (3), Social (4), Sleep (4), Exercise (4). Edge-case options (no\_frequency\_described, no\_plans, no\_baseline, no\_late\_nights, no\_social\_activities, no\_activity, no\_meetings, no\_targets, no\_approach\_described) represent factual edge cases ("the prerequisite condition does not exist for this persona"), not uncertainty; they are deterministic GT labels and do not participate in ordinal encoding. The table prints human-readable answer spaces. For BCF/ABF, the code also defines a per-question bias-shift order: the ordered rank axis along which a $+1$ or $-1$ source prior moves a candidate label. This order is part of the released YAML specification and can differ from the display order for derived-count questions; Table~\ref{tab:C-bias-prior} documents the fixed prior, override rule, and rationale.

\subsubsection{Full Question List}

\begingroup\small
\begin{xltabular}{\linewidth}{@{}lYY@{}}
\caption{All 18 questions in the testbed: full prompt text and targeted multi-source failure mode. Question metadata, answer-space size, ordered/categorical classification, and answer options are listed in Table~\ref{tab:B-2}.}\label{tab:B-3}\\
\toprule
ID & Question & Targeted failure mode \\
\midrule
\endfirsthead
\toprule
ID & Question & Targeted failure mode \\
\midrule
\endhead
\bottomrule
\endfoot
A1 & "In the past 30 days, how many nights did this person sleep 7 hours or more?" & Self-report systematically overreports sleep; device log is most accurate but has missingness and partial field coverage; profile is anchored to early-window averages. Equal weighting or trusting self-report $\to$ systematic overestimation. \\
A2 & "In the past 30 days, on how many days did this person work more than 9 hours?" & Self-report systematically underreports work hours; planner is optimistic. Without objective\_log as cross-check, biased channels dominate $\to$ underestimation of overtime. \\
A3 & "In the past 30 days, on what share of meal occasions did this person eat home-cooked food?" & Self-report inflates home-cooked share; objective payments only catch delivery/restaurant transactions. Trusting self-report $\to$ systematic overestimation; trusting payments alone $\to$ undercount of cash purchases. \\
B2 & "This person's profile describes exercising a certain number of days per week. In the past 30 days, how did their actual exercise frequency compare to that description?" & Identity-behavior bridging requires aggregating sparse exercise events and comparing to a stated profile claim. Methods that anchor on the profile description miss the gap when behavior diverges. \\
B3 & "This person's profile describes a specific approach to weekend work. In the past 30 days, did their actual weekend work pattern match that description?" & LLMs trust the profile identity claim (e.g., "strict work-life boundary") even when objective\_log shows contradictory weekend work. \\
C2 & "In the past 14 days, on days when this person planned a social activity, what share of those plans were actually realized?" & Plan-reality alignment requires both a plan signal (planner) and a reality signal (events). LLMs treat planner absence as "no plan", conflating NO\_RECORD with planned-but-empty. \\
C3 & "In the past 14 days, on days when this person set a target bedtime, did they go to bed within 20 minutes of that target?" & Requires circular time arithmetic across midnight, paired plan-reality observations, and handling device-log missingness. LLMs miss the $\pm$20 min boundary or trust biased self\_report. \\
D1 & "Comparing the first 14 days to the last 16 days of the 30-day window: did this person's number of social activities per day increase, decrease, or stay the same?" & Each source has a directional bias (self-report under-reports casual social, planner is aspirational, objective only catches paid social). LLMs select a narrative trend instead of a windowed quantitative comparison. \\
D2 & "In the past 30 days, did this person's daily home-cooked meal count and total meal count differ from the averages described in their profile by more than 1 meal/day?" & Profile baseline is stale for temporal\_shift personas. LLMs anchor on the profile description rather than computing a baseline-vs-recent delta. \\
E1 & "In the past 30 days, on nights when this person went to bed after midnight, which co-occurring factor appeared on more than 50\% of those nights?" & Factor attribution requires a per-night multi-factor check. LLMs pick the most salient narrative factor (typically work\_stress) without ruling out alternatives. \\
E2 & "In the past 30 days, on days when this person had planned to exercise but did not, did they work more than 8.5 hours that same day?" & Requires per-skip-day cross-checking of work hours against the planned-but-skipped subset. LLMs select the most frequent excuse instead of testing each candidate factor. \\
F1 & "In the past 30 days, on how many days did this person attend social activities when the planner showed no social intent?" & Requires distinguishing NO\_RECORD planner days from "planner present but no intent". LLMs conflate the two, treating any planner absence as "no plan". \\
F2 & "In the past 30 days, on days when the fitness tracker recorded no workout, did other sources (self-report or payment records) indicate that this person exercised?" & Device-log missingness means some "no workout" days reflect missing data, not missing exercise. LLMs treat absence of a tracker entry as evidence of inactivity. \\
F3 & "In the past 30 days, on days when the work timesheet had no entry, did other sources (self-report, planner, or device/activity log) indicate that this person worked?" & LLMs conflate "no timesheet entry" with "did not work", missing the cross-source pattern that confirms work occurred despite the gap. \\
G1 & "In the past 30 days, on days when this person was physically active, was the activity deliberate exercise or incidental movement (such as walking to work or household chores)?" & Latent intent attribute only partially observable from biased projections. Each source has a directional bias (self-report toward intentional, planner aspirational); qualitative prompting does not recover the per-event mapping in our experiments. \\
G2 & "In the past 30 days, when this person attended social activities, were those activities voluntary (by choice) or obligatory (e.g., work events, family commitments)?" & Voluntary/obligatory is a latent qualitative attribute. Self-report carries the primary signal but is biased toward voluntary framing; planner records aspirational intent, not motivation. LLMs conflate activity wording with motivation. \\
Ctrl1 & "In the past week, how many days did this person eat food prepared outside the home?" & Calibration baseline: low cross-source conflict, high agreement. All methods should answer correctly; failures expose extraction or counting bugs rather than reasoning limits. \\
Ctrl2 & "In the past week, how many nights did this person sleep less than 6 hours?" & Calibration baseline (sleep counterpart of Ctrl1): high source agreement on short-sleep nights. Acts as a control anchor across topics. \\
\end{xltabular}
\endgroup

The question templates use survey-style safeguards to reduce ambiguity: fixed time windows, closed answer spaces, explicit numeric boundaries, and edge labels for missing prerequisites. All wording uses neutral language without value-laden terms.

\subsection{Synthetic Data-Generating Process}\label{subsec:B-3}

The DGP runs in four stages.

\begin{itemize}[nosep,leftmargin=*]
  \item Stage 1: Personas. Generates 480 personas across three difficulty classes (\texttt{stable}, \texttt{temporal\_shift}, \texttt{stated\_vs\_revealed}; 160 each), and samples behavioral parameters from domain-specific distributions over five topics.
  \item Stage 2: Latent event table. Produces a 30-day latent event table for each persona recording actual daily behavior.
  \item Stage 3: Source projection. Projects $L_2$ into five visible source files through source-specific schemas, bias/noise, quantization, and missingness, using seeded random number generators.
  \item Stage 4: NL render. Renders each structured source into a natural-language (NL) memory document via deterministic templates, ensuring exact reproducibility and removing NL variance as a confounding variable.
\end{itemize}

A shared LLM extraction step then recovers per-source structured atoms from the NL memory. By design, the LLM performs only factual extraction; all cross-source reasoning is handled by downstream resolvers (\S~\ref{sec:methods}). The DGP is method-agnostic: any new method can be evaluated on the released data without re-running generation or LLM calls. We release the DGP code, generated data across four seeds, pre-rendered NL memories, and pre-extracted structured atoms in the code and dataset artifacts.

\subsection{GT Construction Notes}\label{subsec:B-4}

For the 9 questions that measure source-reality discrepancies (B2-B3, C2-C3, D2, E2, F1-F3), $f_q$ additionally reads the generated source outputs as immutable inputs. A direct-readout gap arises when a visible source schema drops, coarsens, or biases the relevant $L_2$ field enough to move it across an answer boundary. This is intentional: systematic reporting biases can make recorded behavior differ from actual behavior, and some latent attributes, such as whether a social event was voluntary, are only partially visible in the source projections. We quantify this recoverability gap with the Source Reachability reference in \S~\ref{s4:sub1} and by type in Appendix~\ref{subsec:E-1}. The statistic measures exact visibility of GT in direct source readouts; inference methods can still recover some remaining cases by learning how biased source values map back to GT.

Table~\ref{tab:B4-label-functions} gives the label function for each template. Answer labels and edge labels are listed in Table~\ref{tab:B-2}; the table below specifies which fields are read and how one label is selected. For G1/G2, GT is a DGP-latent event attribute, such as intentional exercise or voluntary social activity, while visible sources expose biased projections of that attribute.

\begingroup\small
\begin{xltabular}{\linewidth}{@{}l l Y@{}}
\caption{Per-template GT label functions. $L_2$ denotes the latent 30-day event table. $L_3$ denotes generated source projections that are read only for templates whose rule needs a profile claim, a plan, or source availability.}\label{tab:B4-label-functions}\\
\toprule
Q & GT inputs & Label function \\
\midrule
\endfirsthead
\toprule
Q & GT inputs & Label function \\
\midrule
\endhead
\bottomrule
\endfoot
A1 & $L_2$ sleep & Count 30 nights with sleep duration $\geq 7$h. Use \texttt{20\_or\_more} for $\geq 20$, \texttt{10\_to\_19} for 10--19, and \texttt{fewer\_than\_10} otherwise. \\
A2 & $L_2$ work & Count days with work hours $>9$ among weekdays and weekends that were worked. Use \texttt{8\_or\_more} for $\geq 8$, \texttt{4\_to\_7} for 4--7, and \texttt{0\_to\_3} otherwise. \\
A3 & $L_2$ diet & Compute home-cooked meals divided by all meals over 30 days. Use \texttt{70\_or\_more} for ratio $\geq .70$, \texttt{40\_to\_69} for ratio $\geq .40$, and \texttt{less\_than\_40} otherwise. \\
B2 & $L_2$ exercise, $L_3$ profile & Read the profile exercise frequency. If absent, return \texttt{no\_frequency\_described}. Otherwise compare actual exercise days per week with the profile value; within $\pm 1$ day is \texttt{within\_1\_day}, below by more than 1 is \texttt{more\_than\_1\_below}, and above by more than 1 is \texttt{more\_than\_1\_above}. \\
B3 & $L_2$ work, $L_3$ profile & Read the profile after-hours work style. If absent, return \texttt{no\_approach\_described}. For \texttt{strict\_boundary}, weekend work ratio $\leq .15$ matches. For \texttt{flexible}, zero weekend work over at least four off-days does not match. All remaining profile styles match when weekend work ratio $< .50$. \\
C2 & $L_2$ social, $L_3$ planner & In the last 14 days, find days with planner social intent. If none, return \texttt{no\_plans}. Among planned days, compute the share with at least one latent social activity; $>.50$ is \texttt{above\_50\_pct}, $\geq .25$ is \texttt{25\_to\_50\_pct}, and lower is \texttt{below\_25\_pct}. \\
C3 & $L_2$ sleep, $L_3$ planner & In the last 14 days, pair planner target bedtime with latent bedtime. Times before noon are treated as after midnight for bedtime arithmetic. If no target is present, return \texttt{no\_targets}. If more than half of paired nights are later than target by $>20$ minutes, return \texttt{later\_more\_than\_50pct}; if more than half are earlier by $>20$ minutes, return \texttt{earlier\_more\_than\_50pct}; otherwise return \texttt{within\_20min\_more\_than\_50pct}. \\
D1 & $L_2$ social & Compare social activity rate in the first 14 days with the last 16 days. A late-minus-early rate change $>.15$ is \texttt{increased}; $<-.15$ is \texttt{decreased}; otherwise it is \texttt{stayed\_same}. \\
D2 & $L_2$ diet, $L_3$ profile & Read profile diet baselines for meals per day and home-cooked meals. If either is missing, return \texttt{no\_baseline}. Compute recent 30-day means and sum the absolute deviations in home-cooked meals/day and total meals/day; values $>1$ are \texttt{differs\_more\_than\_1}, otherwise \texttt{within\_1}. \\
E1 & $L_2$ sleep and context tags & Find nights with bedtime after midnight. If none, return \texttt{no\_late\_nights}. Among late nights, count \texttt{work\_stress} tags and \texttt{social} tags. A factor present on more than half of late nights selects \texttt{work\_activity} or \texttt{social\_activity}; otherwise return \texttt{no\_single\_factor}. \\
E2 & $L_2$ exercise and work, $L_3$ planner & Find days where exercise was planned but did not occur. If there are at most two such days, return \texttt{between\_30\_60}. Otherwise compute the share with work hours $>8.5$ or overtime; $>.60$ is \texttt{yes\_more\_than\_60}, $<.30$ is \texttt{no\_fewer\_than\_30}, and the middle range is \texttt{between\_30\_60}. \\
F1 & $L_2$ social, $L_3$ planner & If there are no latent social days, return \texttt{no\_social\_activities}. Otherwise count social days with no planner record or no social intent. Counts 0--3, 4--6, and 7 or more map to the three non-edge labels. \\
F2 & $L_2$ exercise, $L_3$ device log & Inspect days where the device did not detect a workout. Missingness includes unavailable-device days with actual exercise and available-device misses where actual exercise occurred; days with no actual exercise count as confirmed inactivity. Mixed counts give \texttt{both\_occurred}; more missingness than inactivity gives \texttt{yes\_tracker\_missing}; otherwise return \texttt{inactive\_confirmed}. \\
F3 & $L_2$ work, $L_3$ objective log & Inspect days with no available objective work record. If no such day exists, return \texttt{truly\_off}. Otherwise split the missing-record days by latent work hours $>0$ versus no work. Mixed counts give \texttt{both\_occurred}; more worked days than off days gives \texttt{yes\_worked\_despite\_no\_entry}; otherwise return \texttt{truly\_off}. \\
G1 & $L_2$ exercise intent & Among latent active days, compute the share marked intentional. If there are no active days, return \texttt{no\_activity}. A share $>.70$ is \texttt{deliberate\_exercise\_70plus}; a share $<.30$ is \texttt{incidental\_movement\_70plus}; otherwise return \texttt{mix}. \\
G2 & $L_2$ social motivation & Among latent social days, compute the share not marked \texttt{supporting\_other}. If there are no social days, return \texttt{no\_meetings}. A share $>.70$ is \texttt{voluntary\_70plus}; a share $<.30$ is \texttt{obligatory\_70plus}; otherwise return \texttt{mix}. \\
Ctrl1 & $L_2$ diet & In the last 7 days, count days with any meal outside the home, defined as total meals greater than home-cooked meals. Counts 0--1, 2--3, and 4 or more map to the three answer labels. \\
Ctrl2 & $L_2$ sleep & In the last 7 days, count nights with sleep duration $<6$h. Counts 0, 1--2, and 3 or more map to the three answer labels. \\
\end{xltabular}
\endgroup

Table~\ref{tab:B4-source-reachability-complement} asks a narrow diagnostic question: after splitting test instances by whether GT appears as an exact match in any direct-readout source atom, how often does each method still predict GT? The three accuracy columns are instance accuracies on the full test set, the GT-present slice, and the no-exact-source-match slice.

\begin{table}[t]
\centering
\caption{Diagnostic instance accuracy (\%) by direct source reachability. Full is all 8,640 test instances. GT-present is the 8,049 instances where at least one direct-readout source atom $\mu^\ast_s$ exactly equals GT. No exact match is the remaining 591 instances, i.e., the 6.8\% Source Reachability complement. Source Reachability itself is omitted because this slice is defined by its miss condition. These are instance accuracies, not the macro accuracies in Table~\ref{tab:4}.}
\label{tab:B4-source-reachability-complement}
\begingroup\small
\begin{tabularx}{\linewidth}{@{}lYlccc@{}}
\toprule
Group & Method & Input & Full & GT-present & No exact match \\
\midrule
T0 & Majority Class & $\hat\mu$ & 57.1 & 59.6 & 22.7 \\
T1 & SSB & $\hat\mu$ & 77.2 & 81.9 & 13.2 \\
T2 & Majority Vote / ArgRAG-style & $\hat\mu$ & 68.8 & 73.4 & 5.8 \\
 & BCF & $\hat\mu$ & 69.2 & 73.1 & 15.2 \\
 & NBF & $\hat\mu$ & 79.8 & 82.6 & 42.1 \\
 & \textbf{DSNBF} & $\hat\mu$ & 80.3 & 82.3 & 53.5 \\
 & ABF & $\hat\mu$ & 72.0 & 76.5 & 10.7 \\
 & NBF & $\mu^\ast$ & 82.0 & 86.0 & 27.7 \\
 & DSNBF & $\mu^\ast$ & 82.3 & 85.7 & 36.0 \\
T3 & GPT-5.4 Struct-LLM & $\hat\mu$ & 71.4 & 76.2 & 5.9 \\
 & GPT-5.4 Struct-LLM & $\mu^\ast$ & 73.5 & 78.8 & 1.2 \\
 & GPT-5.4 Direct & NL memory & 68.6 & 72.4 & 17.3 \\
 & GPT-5.4 Schema & NL memory & 69.7 & 74.1 & 9.8 \\
 & Gemini 3.1 Direct & NL memory & 67.8 & 71.9 & 12.2 \\
 & Gemini 3.1 Schema & NL memory & 70.0 & 74.3 & 11.5 \\
 & DeepSeek V3.2 Direct & NL memory & 56.1 & 58.1 & 28.9 \\
 & DeepSeek V3.2 Schema & NL memory & 56.9 & 59.3 & 23.9 \\
 & Qwen3 235B Direct & NL memory & 42.1 & 41.8 & 46.4 \\
 & Qwen3 235B Schema & NL memory & 48.0 & 47.1 & 59.2 \\
\bottomrule
\end{tabularx}
\endgroup
\end{table}

Table~\ref{tab:B4-source-reachability-complement} gives two checks on the 6.8\% complement. The no-exact-source-match slice is hard for methods that mostly trust visible labels, confirming that it is a real source-reachability gap. At the same time, DSNBF with extracted atoms still answers 53.5\% of these instances correctly, so exact visibility of GT is not required for recovery. The $\mu^\ast$ row is lower (36.0\%) because the slice is defined by the absence of an exact $\mu^\ast$ match; the extracted-atom row therefore mixes learned source-to-GT inference with extraction-path changes. Since the slice is label-distribution shifted, we use it as a diagnostic companion and keep Table~\ref{tab:4} as the main aggregate comparison.

\subsection{Independent Reproduction of Ground Truth}\label{subsec:B-5}

As an internal audit, a separate implementation, written independently and without reference to the released GT computation code, reimplements the full label generation pipeline: latent event table construction, source projection, and the 18 question-specific label functions $f_q(L_2,L_3,q)$. The audit re-reads the raw event tables and structured source JSON files, recomputes each closed-class label with separate question-specific functions, and compares only final answer labels. Running this independent implementation on all 4 seeds $\times$ 480 personas $\times$ 18 questions (34,560 labels) produces 100\% exact agreement with the released GT labels. This establishes that the released labels are a stable, deterministic consequence of the formal task definition and DGP specification, not an artifact of one specific code path or implementation choice. This is an internal reproducibility check for the formal GT definition, not a reproduction command; external validity of the latent target remains a design question for real-user follow-up.

\subsection{Walked-through Example: One Persona, One Question}\label{app:walked}

We applied a deterministic selection protocol (\S\ref{app:walked-protocol}) to all 18 question templates on the test split (120 personas $\times$ 4 seeds $=$ 480 \texttt{(persona, seed)} pairs per question) and chose A3 (Home-Cooked Share, Diet topic), whose qualifying pool contained $N=43$ pairs satisfying all five hard filters; the displayed instance was selected by the deterministic protocol in Appendix~\ref{app:walked-protocol}. The numbered paragraphs below correspond to the boxes of Figure~\ref{fig:1}, walking one instance through every stage of the DAG.

\paragraph{Stage 0 (Figure~\ref{fig:1}, Question $q$). A3.} ``In the past 30 days, on what share of meal occasions did this person eat home-cooked food?'' Answer space (3 ordinal bins): \texttt{less\_than\_40} $<$ \texttt{40\_to\_69} $<$ \texttt{70\_or\_more}. Ground truth: \texttt{40\_to\_69} (52 of 92 meals home-cooked, $\rho = 0.565$), computed deterministically from the latent event table.

\paragraph{Stage 1 (Figure~\ref{fig:1}, Persona L$_1$).} \texttt{bench\_shift\_123\_taylor\_ellis}, seed 1, difficulty class \texttt{temporal\_shift}: a 34-year-old vegetarian accountant whose stated profile claims roughly 68\% of meals home-cooked, but whose realised 30-day behavior shifts toward more delivery.

\paragraph{Stage 2 (Figure~\ref{fig:1}, Latent Events L$_2$).}
The 30-day meal log contains 92 events; 52 are home-cooked. Eight representative days:
\begin{center}
\small
\begin{tabular}{lcccl}
\toprule
Date & Day & Meals & Home-cooked & Context \\
\midrule
2026-01-03 & Sat & 3 & 2 & social, gallery opening \\
2026-01-04 & Sun & 3 & 3 & social \\
2026-01-05 & Mon & 3 & 2 & routine workday \\
2026-01-06 & Tue & 3 & 1 & overtime (9.8\,h) \\
2026-01-12 & Mon & 3 & 2 & routine workday \\
2026-01-20 & Tue & 2 & 1 & high stress \\
2026-01-30 & Fri & 3 & 2 & post-shift slack \\
2026-02-01 & Sun & 3 & 0 & post-shift slack \\
\bottomrule
\end{tabular}
\end{center}
The drop from ${\sim}70\%$ home-cooked in days 1--13 to ${\sim}40\%$ in days 14--30 instantiates this persona's \texttt{temporal\_shift} difficulty class.

\paragraph{Stage 3 (Figure~\ref{fig:1}, 5 Source Streams L$_3$).}
Each source projects the same $L_2$ table through its DGP-defined schema. Diet-relevant fields:
\begin{center}
\small
\begin{tabular}{lp{8.6cm}}
\toprule
Source & Diet-relevant fields (selected) \\
\midrule
\texttt{profile\_ltm}        & \texttt{\{meals\_per\_day:\,3.1, home\_cooked\_mean:\,2.1, dietary\_restriction:\,veg\}} (anchored on days 1--13 only; stale by 17 days) \\
\texttt{planner}             & 30 daily entries: \texttt{\{home\_cooked\_priority:\,true\}} every day; no quantitative share \\
\texttt{daily\_self\_report} & 30 daily entries: self-reported \texttt{\{meals,\,home\_cooked\}} counts; aggregate share $0.720$ (vs.\ true $0.565$) \\
\texttt{objective\_log}      & 30 daily entries: payments (\texttt{food\_delivery}, \texttt{coffee}) + timesheet; no grocery or cooking field \\
\texttt{device\_log}         & 30 daily entries: sleep + activity tracker only; no diet field by schema \\
\bottomrule
\end{tabular}
\end{center}

\paragraph{Stage 4 (Figure~\ref{fig:1}, NL Memory Render). Excerpts (single day, 2026-01-06).}
The deterministic renderer turns each $L_3$ source into prose. One day per source:
\begin{itemize}[nosep,leftmargin=*]
  \item \texttt{profile\_ltm}: ``Diet: averages 3.1 meals per day, about 2.1 of those are home-cooked.''
  \item \texttt{planner}: ``2026-01-06 (Tue). Home-cooked food prioritized. Work capped at 8.0 hours, finish by 18:00.''
  \item \texttt{daily\_self\_report}: ``2026-01-06. Diet: 3 meals total, 3 home-cooked. Work: 9.8 hours, overtime.''
  \item \texttt{objective\_log}: ``2026-01-06. Payments: 2 coffee purchases, 1 food-delivery purchase. Calendar: work block 09:00--18:24.''
  \item \texttt{device\_log}: ``2026-01-06. Sleep: bed 23:39, wake 07:58, 7.8 hours. Activity: 19 active minutes.'' (no diet content)
\end{itemize}

\paragraph{Stage 5 (Figure~\ref{fig:1}, LLM Extraction). NL $\to \hat\mu$.}
The extractor (GPT-5.4, max reasoning) reads each source's NL render in isolation and emits structured atoms per question; no cross-source reasoning and no GT access. For A3 it returns a JSON of the form \texttt{\{"A3":\,"$\langle$bin$\rangle$"\}} per source, where $\langle$bin$\rangle \in \{\texttt{less\_than\_40}, \texttt{40\_to\_69}, \texttt{70\_or\_more}\}$. For \texttt{daily\_self\_report}, the 30-day NL text implies a self-reported share $\rho_{\text{self}} = 0.720$, and the emitted atom is consistent with binning that aggregate as \texttt{70\_or\_more}. The emitted atoms across the five sources are tabulated next.

\paragraph{Stage 6 (Figure~\ref{fig:1}, Atoms $\hat\mu$). Source extracted atoms.}
\begin{center}
\begin{tabular}{llll}
\toprule
Source & Note & $\hat\mu$ & Spec bias $b_s$ (Diet) \\
\midrule
\texttt{profile\_ltm}        &     & \texttt{40\_to\_69}   & 0    \\
\texttt{planner}             &     & \texttt{70\_or\_more} & $+1$ \\
\texttt{daily\_self\_report} &     & \texttt{70\_or\_more} & $+1$ \\
\texttt{objective\_log}      & $\dagger$ & \texttt{70\_or\_more} & 0 \\
\texttt{device\_log}         &     & null (schema absence) & 0    \\
\bottomrule
\end{tabular}
\end{center}

\noindent Note $\dagger$: For A3, \texttt{objective\_log} records delivery purchases but not grocery or cooking events, so it is an incomplete proxy for home-cooked share. The resulting \texttt{70\_or\_more} atom reflects information missing from this source rather than LLM extraction error; aggregate extraction faithfulness is audited in Appendix~\ref{app:audit}.

\paragraph{Stage 6$'$ (Figure~\ref{fig:1}, direct source readout). $\mu^\ast$ atoms and extraction faithfulness.}
The Source Reachability reference reads the structured $L_3$ fields directly via a hand-coded Python readout and emits atoms $\mu^\ast$ with zero LLM involvement. Walking through the same day for \texttt{daily\_self\_report}: the readout pulls the JSON field \texttt{\{"meals":\,3,\,"home\_cooked":\,3\}} for 2026-01-06, repeats over the 30 dated entries, sums to $(93, 67)$, computes $\rho_{\text{self}} = 67/93 = 0.720$, and bins to \texttt{70\_or\_more}. The two extraction paths are reading the same underlying $L_3$ projection in different surface forms; for this instance $\mu^\ast = \hat\mu$ for all five sources: \texttt{(40\_to\_69, 70\_or\_more, 70\_or\_more, 70\_or\_more, null)}. NL extraction introduces no atom-level noise here. This is the dominant pattern globally: across the test split, $\hat\mu = \mu^\ast$ in $93.08\%$ of \texttt{(persona, seed, question, source)} cells (Appendix~\ref{app:audit}).

\paragraph{Stage 7 (Figure~\ref{fig:1}, GT $f_q(L_2,L_3,q)$). Ground truth derivation.}
The label function $f_{A3}$ is deterministic: count home-cooked meal events in the 30-day latent event table, divide by total meals, assign to bin.
\[
\rho \;=\; \frac{\#\{\text{home\_cooked}\}}{\#\{\text{meals}\}} \;=\; \frac{52}{92} \;=\; 0.5652 \;\in\; [0.40,\,0.70) \quad\Longrightarrow\quad \texttt{40\_to\_69}.
\]
The independent audit implementation (\S\ref{subsec:B-5}) computes the same value bit-for-bit; both populations write to \texttt{ground\_truth.json} as \texttt{\{"A3":\,\{"answer":\,"40\_to\_69", "derivation\_detail":\,"home=52/total=92 ratio=0.565"\}\}}.

\paragraph{Per-source distortion taxonomy on this instance.} To separate which mechanism produced each source's answer, we decompose the five sources below:

\begin{center}
\begin{tabular}{llp{6.3cm}}
\toprule
Source & Distortion type & Mechanism on this instance \\
\midrule
\texttt{profile\_ltm}        & same-bin coincidence
  & Stated $67.7\%$ and true $56.5\%$ both map to \texttt{40\_to\_69} bin ($[0.40, 0.70)$); correct by coincidence under this 3-bin quantization. \\
\texttt{planner}             & quantization + bias
  & Schema carries only intent boolean (``home-cooked prioritized'': yes/no), no quantitative share; under $b = +1$ the channel defaults to high-share projection, yielding \texttt{70\_or\_more}. \\
\texttt{daily\_self\_report} & boundary-crossing bias
  & Self-reported share $0.720$ exceeds the $0.70$ cutoff by $0.020$; with true $\rho = 0.565$, bias $b \approx +0.155$ is just sufficient to cross the bin boundary. \\
\texttt{objective\_log}      & schema projection gap
  & NL render projects only delivery events; grocery and cooking events, though present in $L_2$, are absent from this channel. Extractor infers $1 - \text{delivery\_share}$, yielding \texttt{70\_or\_more}. $\hat\mu = \mu^\ast$ here. \\
\texttt{device\_log}         & schema absence
  & Channel carries no diet fields by DGP design. Output is null on this instance (not a dropout artefact). \\
\bottomrule
\end{tabular}
\end{center}

\noindent The five rows instantiate five distinct distortion mechanisms: same-bin coincidence, boolean quantization, boundary-crossing bias, schema projection gap, and schema absence. MV and SSB do not model these mechanisms jointly; a learned per-source confusion matrix (NBF, DSNBF) absorbs all five jointly, which is why it is the first fusion family to recover GT on this instance.

\paragraph{Stage 8 (Figure~\ref{fig:1}, resolvers + reference). Method predictions.}
\begin{center}
\small
\begin{tabular}{llllc}
\toprule
Group & Method & Prediction & Skip? & Correct? \\
\midrule
T0 & Random                              & \texttt{70\_or\_more}       & n/a & $\times$ \\
T0 & Majority-Class                      & \texttt{40\_to\_69}         & n/a & \checkmark (train prior) \\
\midrule
T1 & SSB                                 & \texttt{70\_or\_more}       & n/a & $\times$ \\
T1 & SSB+SKIP                            & \texttt{70\_or\_more}       & no  & $\times$ \\
\midrule
T2 & Majority-Vote                       & \texttt{70\_or\_more}       & n/a & $\times$ \\
T2 & BCF (4p)                            & \texttt{70\_or\_more}       & n/a & $\times$ \\
T2 & NBF                                 & \texttt{40\_to\_69}         & n/a & \checkmark \\
T2 & NBF+SKIP                            & skip               & yes & skipped \\
T2 & DSNBF ($m = 0.30$)                  & \texttt{40\_to\_69}         & n/a & \checkmark \\
T2 & DSNBF+SKIP ($m{<}\theta_E{=}0.39$)  & skip               & yes & skipped \\
\midrule
T3 & GPT-5.4 LLM-Direct                  & \texttt{40\_to\_69}         & n/a & \checkmark \\
T3 & GPT-5.4 LLM+SKIP                    & skip               & yes & skipped \\
T3 & GPT-5.4 Schema-Aware                & \texttt{70\_or\_more}       & no  & $\times$ \\
T3 & GPT-5.4 on $\mu^\ast$ & \texttt{70\_or\_more}    & no  & $\times$ \\
T3 & Gemini 3.1 Pro Direct               & \texttt{70\_or\_more}       & n/a & $\times$ \\
T3 & Gemini 3.1 Pro Schema-Aware         & \texttt{70\_or\_more}       & no  & $\times$ \\
\midrule
Ref. & Source Reachability      & \texttt{40\_to\_69}         & n/a & \checkmark (reference) \\
\bottomrule
\end{tabular}
\end{center}
\noindent Variants marked \texttt{+SKIP} are the selective (abstaining) counterparts; NBF/DSNBF+SKIP abstain when posterior margin $p(\text{top1}) - p(\text{top2}) < \theta_E$ (per-method calibrated threshold, Appendix~\ref{subsec:D-2}). LLM+SKIP abstains when the answering LLM reports \texttt{would\_skip=true}. ``n/a'' marks methods without a skip mechanism.

\paragraph{Stage 9 (Figure~\ref{fig:1}, Score vs.\ GT). How this instance enters the aggregate metrics.}
For each non-skipping resolver, the prediction is compared against GT (\texttt{40\_to\_69}) and contributes one binary correctness sample to the per-question accuracy in Table~\ref{tab:4}: NBF, DSNBF, Majority-Class, and GPT Direct each contribute a $1$, while SSB, MV, BCF, GPT Schema, GPT-5.4 on $\mu^\ast$, Gemini Direct, and Gemini Schema each contribute a $0$. Source Reachability is reported separately as a GT-aided direct-readout reference; this instance is reachable because GT appears in the $\mu^\ast$ readout. For methods with SKIP, the same instance feeds the selective-QA pair $(\text{coverage}, \text{selective accuracy})$ in Table~\ref{tab:D-1}. NBF+SKIP, DSNBF+SKIP, and GPT-LLM+SKIP skip this item, lowering coverage while leaving the selective-accuracy numerator and denominator unchanged. $F_{0.5}$ is used only on the calibration split to choose SKIP thresholds; once fixed, the thresholds determine the reported test coverage and selective-accuracy cells. The instance therefore enters Tables~\ref{tab:4}, \ref{tab:6}, \ref{tab:D-1}, and Tables~\ref{tab:E2a}--\ref{tab:E2b-skip} once at the (method, A3, \texttt{temporal\_shift}) coordinate.

\paragraph{Takeaway.} This single instance illustrates five mechanisms. First, the \texttt{planner} and \texttt{daily\_self\_report} atoms both shift one bin upward relative to GT, matching their positive Diet bias. Second, \texttt{objective\_log} also lands on \texttt{70\_or\_more} even though its spec-level bias is $b_{\text{obj}} = 0$. The source records only delivery transactions here, and $\hat\mu_{\text{obj}} = \mu^\ast_{\text{obj}}$, so the error comes from structured projection loss at $L_2 \to L_3$ rather than from LLM extraction or a mean source-bias parameter.

DSNBF's per-$(s,q)$ confusion matrix is estimated from the empirical joint distribution of $\hat\mu$ and GT, so it absorbs projection loss, quantization loss, and bias jointly. This accounts for its margin over methods that model only one axis of distortion, such as SSB and Majority Vote.

Third, Schema-Aware prompting flips GPT from a correct \texttt{40\_to\_69} to an incorrect \texttt{70\_or\_more}, matching the aggregate overcorrection pattern in \S~\ref{s4:sub3}. The matched-input GPT row (GPT-5.4 on $\mu^\ast$, \S~\ref{s4:sub2}) gives the same incorrect answer, placing the error after extraction for this example.

Fourth, selective abstention pays a coverage cost on this item. NBF+SKIP, DSNBF+SKIP, and GPT-5.4 LLM+SKIP all abstain, while their answer-only counterparts choose the GT bin. The aggregate coverage/accuracy trade-off is reported in Appendix~\ref{subsec:D-2}. The skip mechanisms also differ: NBF+SKIP and DSNBF+SKIP use calibrated posterior thresholds, GPT-5.4 LLM+SKIP uses a self-reported flag, and SSB+SKIP fires only when the chosen source returns no atom.

Fifth, the two majority baselines diverge. Source-level Majority Vote follows the three biased sources to \texttt{70\_or\_more}, while population-level Majority Class happens to match GT through the training-set modal label. Among fusion methods, NBF and DSNBF recover GT through learned per-source reliability. LLM Direct answers split across model families, with GPT correct and Gemini wrong; Table~\ref{tab:4} gives the aggregate per-method result.

\subsection{Selection Protocol for the Walked-through Example}\label{app:walked-protocol}

The instance shown in \S~\ref{app:walked} was selected by the following deterministic protocol applied to all 18 question templates on the test split (120 personas $\times$ 4 seeds $=$ 480 \texttt{(persona, seed)} pairs per question). Hard filters: (i) at least 4 of 5 sources produce a non-empty extracted atom, and the present sources include \texttt{planner}, \texttt{daily\_self\_report}, and \texttt{objective\_log}; (ii) DSNBF $\arg\max == \mathrm{GT}$; (iii) at least one frontier T3-LLM (GPT-5.4 Direct or Gemini 3.1 Pro Direct) $\arg\max \neq \mathrm{GT}$; (iv) the source extracted atoms span at least two distinct values; (v) DSNBF posterior margin $p(\text{top1}) - p(\text{top2}) \in [0.08, 0.35]$, excluding both trivial and marginal cases. Soft preferences (used to rank within the qualifying pool): planner positive bias visible in the extracted atom; \texttt{daily\_self\_report} directional bias visible in the source text; cross-family LLM divergence; at least one other fusion method also correct. Final tiebreak: lexicographic sort on \texttt{(persona\_id, seed, q\_id)}. The qualifying pool size for A3 was $N = 43$.

\subsection{Atom Extraction Faithfulness Audit}\label{app:audit}

To quantify how often LLM extraction recovers exactly what the direct structured readout produces, we compare $\hat\mu$ against $\mu^\ast$ at the \texttt{(persona, seed, question, source)} cell level on the full test split (120 personas $\times$ 4 seeds $\times$ 18 questions $\times$ 5 sources $= 43{,}200$ cells). Both arms see the same $L_3$ structured projections; only the atom-level extraction path differs ($\hat\mu$ via GPT-5.4 reading NL, $\mu^\ast$ via deterministic Python reading the structured JSON). $\hat\mu = \mu^\ast$ in $\mathbf{93.08\%}$ of cells. The remaining $6.92\%$ are concentrated in three questions (A2, G2, F2; per-question equality $81$--$84\%$) and one source (\texttt{objective\_log}; equality $86.5\%$).

\begin{center}
\small
\begin{tabular}{lr}
\toprule
Source & $P(\hat\mu = \mu^\ast)$ \\
\midrule
\texttt{profile\_ltm}        & $91.24\%$ \\
\texttt{planner}             & $96.28\%$ \\
\texttt{daily\_self\_report} & $96.20\%$ \\
\texttt{objective\_log}      & $86.46\%$ \\
\texttt{device\_log}         & $95.20\%$ \\
\midrule
Overall                      & $\mathbf{93.08\%}$ \\
\bottomrule
\end{tabular}
\end{center}
\noindent Cells where both $\hat\mu$ and $\mu^\ast$ are null count as equal; this primarily affects \texttt{device\_log} (no diet schema by design) and \texttt{objective\_log} on questions outside its observation domain. Equality rate is uniform across seeds (per-seed range $92.66$--$93.94\%$) and across difficulty classes (\texttt{stable} $93.02\%$, \texttt{temporal\_shift} $93.81\%$, \texttt{stated\_vs\_revealed} $92.40\%$).

The audit supports two claims used in the main text. First, LLM extraction is faithful on average: for $93\%$ of cells the NL$\to\hat\mu$ step produces the same atom that a hand-coded JSON readout would. Second, for GT/source mismatches, the audit shows that most are unlikely to be caused by NL extraction alone, because $\hat\mu$ usually matches $\mu^\ast$. The Source Reachability reference (Table~\ref{tab:4}) bypasses extraction noise but inherits every upstream projection artefact. See the number legend in \S~\ref{s4:sub1} for the distinction among 80.3, 82.3, 93.08, 93.2, and 6.8. The pure extraction contribution is captured by the $\sim$2 pp factorial input effect in \S~\ref{s4:sub2}, which holds the resolver fixed and swaps $\hat\mu$ for $\mu^\ast$. Because $\hat\mu$ in the main tables denotes the frozen GPT-5.4 extraction cache, this audit uses GPT-5.4; Appendix~\ref{subsec:F-6} reports the corresponding Gemini 3.1 Pro cross-extractor comparison, including per-question extraction accuracy.

\section{Method Details}\label{app:methods}
\subsection{Fusion Method Formulas}\label{subsec:C-1}

Bias Corrected Fusion (BCF). BCF treats each candidate answer as a hypothesis. For each hypothesis $v$, a forward bias model predicts what each source would report, $g_s(v)$, by shifting $v$ along the source's bias direction and clamping to the ordinal range. The candidate with the most source prediction matches wins:

$$\hat{y} = \arg\max_v \sum_{s:\mu_s \neq \text{null}} (1 - \delta_s)\cdot \mathbb{1}[\mu_s = g_s(v)]$$

where $\delta_s \in [0, 0.5]$ are per-source deflation weights (4 learnable, $\delta_{\text{obj}} = 0$ fixed), grid-searched on the training split. Because the shift mechanism assumes ordinal structure, BCF degenerates to Majority Vote on the 3 unordered categorical questions. Table~\ref{tab:C-bias-prior} gives the fixed directional prior used by the forward bias model. A shift of $+1$ or $-1$ means one step along the per-question bias-shift order in the released YAML specification; the signs are semantic priors fixed before evaluation, not empirical fits. Table~\ref{tab:B-2} prints human-readable answer spaces, which need not be identical to this bias-shift order for every derived-count question.

\begin{table}[t]
\centering
\caption{Fixed source-bias prior used by BCF and the bias-aware path of ABF. The prior is hand-specified a priori from common-sense source semantics and answer-scale direction; it is not estimated from train, calibration, or test labels.}
\label{tab:C-bias-prior}
\begingroup\small
\begin{tabularx}{\linewidth}{@{}lY@{}}
\toprule
Component & Prior \\
\midrule
Default source/topic prior &
\texttt{planner} has shift $+1$; \texttt{profile\_ltm}, \texttt{objective\_log}, and \texttt{device\_log} have shift $0$; \texttt{daily\_self\_report} follows topic-level defaults ($+1$ for sleep, diet, and exercise; $-1$ for work and social), including control questions unless an override is listed below. \\
Per-question overrides &
A2: \texttt{daily\_self\_report} $=+1$; C2: \texttt{daily\_self\_report} $=+1$; F1: \texttt{planner} $=-1$ and \texttt{daily\_self\_report} $=+1$; G2: \texttt{daily\_self\_report} $=+1$. \\
Resolution rule &
Per-question overrides take precedence over source/topic defaults; unspecified source-question pairs fall back to the default prior, and then to $0$. \\
Scope &
This prior is used only by BCF and ABF. NBF and DSNBF instead learn source-to-GT confusion matrices from train labels. \\
\bottomrule
\end{tabularx}
\endgroup
\end{table}

\noindent Override rationale. The defaults are topic-level priors, but four derived questions reverse the usual topic semantics. A2 counts long-work days; because its bias-shift order is reversed, underreporting work requires a $+1$ shift toward fewer overtime days, not toward the high-work label. C2 asks which planned social activities were actually realized; retrospective self-report tends to emphasize completed plans, shifting toward higher realization rather than the generic ``less social'' prior. F1 asks about unplanned social activity: planner records should shift toward fewer unplanned events, while self-report can reveal activities that occurred despite no prior plan. G2 asks whether social activities were voluntary or obligatory; self-report idealization shifts toward voluntary framing. These overrides therefore encode the intended source semantics for each answer scale, without using held-out labels.

Naive Bayes Fusion (NBF). NBF drops BCF's assumption that bias acts as a fixed ordinal shift and instead learns the full empirical relationship between ground truth and each source's report. For each source $s$, a confusion matrix $C_s \in \mathbb{R}^{|\mathcal{V}| \times |\mathcal{V}|}$ records $C_s[v, v'] = P(\mu_s = v' | y = v)$, estimated with Laplace smoothing on the training split. Given a new instance, the posterior is:

$$\hat{y} = \arg\max_v \; P(v) \prod_{s:\mu_s \neq \text{null}} C_s[v,\; \mu_s]$$

The class prior $P(v)$ is estimated as the empirical ground-truth marginal on the 216-persona training split with Laplace smoothing ($\alpha=1$), re-estimated per seed; calibration and test splits are not accessed during parameter estimation. SKIP thresholds are selected later on the calibration split. This yields $5 \times |\mathcal{V}|^2$ free parameters per question.

Difficulty Stratified NBF (DSNBF). DSNBF observes that sources distort differently across persona difficulty classes (\texttt{stable}, \texttt{temporal\_shift}, \texttt{stated\_vs\_revealed}) and learns a separate confusion matrix $C_s^{(d)}$ for each class. To avoid overfitting when per-class counts are small, each $C_s^{(d)}$ is regularized toward the global matrix via a Dirichlet prior with strength $\eta$. At test time the persona's difficulty class is unobserved; DSNBF infers it from the pattern of source observations across all 18 questions using Bayesian marginal likelihood, then blends global and difficulty-stratified posteriors with a mixing weight $g_w$.

Information boundary for DSNBF. The difficulty stratification is learned from labeled training personas only. The method estimates a training prior $\pi_{\mathrm{train}}(d)$ and per-difficulty source matrices from the 216-persona train split; $\eta$, the emission temperature $T$, the difficulty-estimation temperature $T_{\mathrm{diff}}$, the global/difficulty mixing weight $g_w$, and any SKIP margin threshold are selected on the calibration split. No statistic of the held-out test-set difficulty distribution is passed to DSNBF. For a test persona $i$, the runtime call receives only its observed atom table $\mu_i$ and computes
\[
\tilde p_i(d)
\propto
\pi_{\mathrm{train}}(d)
\prod_q
\sum_{y \in \mathcal V_q}
P_d(y)
\prod_{s:\mu_i(s,q)\neq\mathrm{null}}
C_{s,q}^{(d)}[y,\mu_i(s,q)]^{T_{\mathrm{diff}}}.
\]
This posterior is persona-local: it may infer that an individual test persona looks stable, temporal-shifted, or stated-vs-revealed from its source pattern, but it never observes the aggregate counts of stable/temporal-shift/stated-vs-revealed personas in the test split, nor the true difficulty label used later for per-difficulty reporting. In the evaluation code, \texttt{evaluate\_method} calls \texttt{prepare\_persona(inst["mu"])} before prediction; \texttt{inst["difficulty"]} is consumed only after the prediction to accumulate reporting cells.

Persona identifiers are bookkeeping keys only. They are used to join files, cache outputs, and group bootstrap resamples, but they are not passed as model features or included in LLM prompt text for the target persona. Evaluated prediction calls receive only the source atom table (T1/T2), the rendered NL memory (T3), or the structured atom grid used for the matched-input diagnostic; they do not receive \texttt{persona\_id}, seed names, split paths, or difficulty-prefix strings such as \texttt{bench\_stable}, \texttt{bench\_shift}, or \texttt{bench\_stated}. Difficulty labels and ID prefixes are used for generation, training/calibration labels, selection of few-shot exemplars, and reporting strata, but not as target-persona prediction inputs. External evaluators should preserve this boundary by stripping or ignoring identifiers before invoking prediction code; using ID prefixes as features would be leakage rather than a valid benchmark submission.

Abductive Bias Fusion (ABF). ABF scores each candidate answer $v$ via a soft-mixture kernel that blends two alignment paths. The bias-aware path compares each source's observation $\mu_s$ against the bias-shifted prediction $g_s(v)$ (as in BCF, using the fixed prior and override rule in Table~\ref{tab:C-bias-prior}). The bias-free path compares $\mu_s$ directly against $v$ without any bias correction. The combined score is:

$$\text{ExplScore}(v) = \sum_s w_s \big[\pi \cdot \kappa_\alpha(d_{\text{bias}}(v, \mu_s, \delta_s)) + (1-\pi) \cdot \kappa_\alpha(d_{\text{id}}(v, \mu_s))\big]$$

where $\alpha$ controls kernel bandwidth and $\pi$ controls the bias-identity mixture. ABF has 6 learnable parameters (4 bias offsets $\delta_s$ + $\alpha$ + $\pi$), fitted by MLE on the training split.

\subsection{Selective (SKIP) Mechanisms}\label{subsec:C-2}

Methods marked $\pm$ in Table~\ref{tab:3} support a selective mode where the system may abstain instead of committing an answer. The SKIP decision differs by method family.

\begin{itemize}[nosep,leftmargin=*]
  \item Bayesian fusion (NBF, DSNBF). SKIP is triggered when the posterior margin (the gap between the top two posterior probabilities) falls below a calibrated threshold $\theta$; instances with fewer than two active sources always answer.
  \item ABF. A two-criterion rule: if the best ExplScore falls below $\theta_E$ (insufficient absolute evidence) or if the gap to the second-best candidate falls below $\theta_\Delta$ (ambiguous winner), the method abstains.
  \item SSB. Skips when its best source returns null or, under a stricter variant, when fewer than $\theta_{\text{agree}}$ fraction of active sources agree with the chosen answer.
  \item LLM methods. The model itself decides: each prompt requests both an answer-only label and a binary \texttt{would\_skip} indicator; no external threshold is applied.
\end{itemize}

Source Reachability is a GT-aided diagnostic row. It reports whether some direct-readout source observation matches GT, and we use it only to measure how much of the benchmark is exactly visible in the source projections.

SKIP thresholds for threshold-based variants are calibrated on a held-out calibration split by maximizing $F_{0.5}$ ($\beta = 0.5$, weighting precision above recall).

\subsection{Evaluation Metrics}\label{subsec:C-metrics}

All answer-only accuracy numbers reported as ``macro accuracy'' use a macro-over-question reduction. Let $\mathcal{Q}$ be the 18 question templates, $\mathcal{I}_q$ the evaluated personas for question $q$, $y_{i,q}$ the deterministic GT label, and $\tilde a_{i,q}$ the method's answer in answer-only mode. For selective methods, $\tilde a_{i,q}$ is the answer that would have been returned before abstention when the method records one; otherwise a SKIP without a raw answer counts as incorrect. The reported forced macro accuracy is
\[
\mathrm{Acc}_{\mathrm{macro}}
= \frac{1}{|\mathcal{Q}|}
\sum_{q\in\mathcal{Q}}
\frac{1}{|\mathcal{I}_q|}
\sum_{i\in\mathcal{I}_q}
\mathbb{1}[\tilde a_{i,q}=y_{i,q}].
\]
Coverage is micro-averaged over evaluated instances: $\mathrm{Cov}=\sum_{q,i}\mathbb{1}[\text{answered}_{i,q}]/\sum_q|\mathcal{I}_q|$. Selective accuracy is macro-over-question on the answered subset:
\[
\mathrm{SelAcc}_{\mathrm{macro}}
= \frac{1}{|\mathcal{Q}_{\mathrm{ans}}|}
\sum_{q\in\mathcal{Q}_{\mathrm{ans}}}
\frac{\sum_{i\in\mathcal{I}_q}\mathbb{1}[\text{answered}_{i,q}\wedge a_{i,q}=y_{i,q}]}
{\sum_{i\in\mathcal{I}_q}\mathbb{1}[\text{answered}_{i,q}]},
\]
where $\mathcal{Q}_{\mathrm{ans}}$ excludes question templates for which a method abstains on every evaluated persona. Thresholds are selected on the calibration split by maximizing $F_{0.5}$. In the released code, this uses $P=\mathrm{SelAcc}_{\mathrm{macro}}$, $R=(\mathrm{SelAcc}_{\mathrm{macro}}\cdot \mathrm{Cov})/\mathrm{Acc}_{\mathrm{macro}}$, and
\[
F_\beta = \frac{(1+\beta^2)PR}{\beta^2P+R},\qquad \beta=0.5.
\]
Main-table confidence intervals pool all four held-out test seeds and use percentile bootstrap resampling over seed-persona clusters; appendix tables marked ``4-seed mean'' instead average the stated cell metric across seeds.

\subsection{LLM Configurations and Information Conditions}\label{subsec:C-3}

T1 and T2 methods share a common extraction stage in which an LLM converts each NL memory document into structured atoms $\hat\mu^m(s,q)$ for all (source, question) pairs. We use GPT-5.4 \citep{openai2026gpt54} as the primary extraction model with reasoning effort set to maximum, so $\hat\mu$ denotes $\hat\mu^{\mathrm{GPT}}$ unless stated otherwise. The extraction prompt constrains the LLM to factual recovery only: it reports what each source says without cross-source reasoning or conflict resolution. As a cross-extractor robustness check, we also extract with Gemini 3.1 Pro Preview and feed the resulting atoms into fusion methods (Appendix~\ref{subsec:F-6}). A direct $L_3$ readout uses hand-coded Python functions that read the structured JSON fields directly, with zero LLM involvement.

\paragraph{Released-cache boundary.} The input column in Table~\ref{tab:3} denotes prediction-time input. In the API-free released reproduction, fitted T1/T2 rows labeled $\hat\mu$ fit source-selection, fusion, and SKIP parameters on deterministic train/calibration direct-readout atoms $\mu^\ast$, then evaluate on frozen GPT-5.4 held-out test extractions $\hat\mu$. Rows labeled $\mu^\ast$ use direct-readout atoms for train, calibration, and test. Thus the reported $\hat\mu$ fusion rows measure a structured resolver trained from clean source-to-GT supervision and deployed on extracted test atoms; refitting from extracted train/calibration atoms would require regenerating non-test extractions with live APIs and is not part of the frozen no-API reproduction path.

We evaluate four LLM families under identical prompt templates with no per-family tuning:

\begin{itemize}[nosep,leftmargin=*]
  \item GPT-5.4 \citep{openai2026gpt54} with reasoning effort \texttt{xhigh} (maximum).
  \item Gemini 3.1 Pro Preview \citep{google2026gemini31}, API model \texttt{gemini-3.1-pro-preview}, with \texttt{thinking\_level}/\texttt{thinkingLevel}=\texttt{high}; also feeds extraction into all fusion methods as a cross-extractor robustness check (Appendix~\ref{subsec:F-6}).
  \item DeepSeek V3.2 \citep{deepseek2025v32}, 2025-12 release, strong open-weight model.
  \item Qwen3-235B-A22B-Instruct-2507 \citep{qwen2025qwen3,qwen2025qwen3instruct2507}, 2025-07 release, largest instruction-tuned variant.
\end{itemize}

All four families run LLM-Direct and Schema-Aware across all four seeds. GPT-5.4 additionally serves as the LLM arm of the matched input 2$\times$2 decomposition (\S~\ref{s4:sub2}) and as the base for a supplementary few-shot experiment on seed 1 (Appendix~\ref{subsec:D-3}). The four families span two proprietary (GPT, Gemini) and two open-weight (DeepSeek, Qwen3), reducing the chance that the resolver-versus-LLM gap is an artifact of one model family. Each LLM call returns an answer label from the question's answer space $\mathcal{V}_q$ plus a binary \texttt{would\_skip} indicator, allowing us to evaluate both answer-only and selective modes from a single API call.

We define four information conditions to support the factorial decomposition in \S~\ref{s4:sub2}:

\begin{enumerate}[nosep,leftmargin=*]
  \item NL. The LLM reads raw NL memory (LLM-Direct).
  \item Schema-Aware. The LLM receives NL memory augmented with source-bias descriptions and reliability guidance.
  \item $\hat\mu$ input. A method reads structured atoms $\hat\mu$ extracted from NL memory.
  \item $\mu^\ast$ input. A method reads structured atoms $\mu^\ast$ directly from the structured $L_3$ source streams.
\end{enumerate}

The 2$\times$2 crossing of resolver (DSNBF vs. GPT-5.4) and input quality ($\hat\mu$ vs.\ $\mu^\ast$) enables the gap decomposition reported in \S~\ref{s4:sub2}. We interpret the split as a supervised-resolver versus prompted-LLM design result.

\subsection{ArgRAG Adaptation}\label{subsec:C-argrag}

ArgRAG builds a Quantitative Bipolar Argumentation Framework (QBAF) from a claim and retrieved evidence passages. The original algorithm labels evidence as support, contradict, or irrelevant, drops irrelevant evidence, adds support/attack edges, and predicts the claim from its final strength under QE gradual semantics \citep{zhu2025argrag}. Our input is different: each source provides a closed-class atom, not a free-text passage. To keep this a deterministic T2 method, our label-level adaptation scores each candidate answer $v$ as the claim ``the answer is $v$.'' A non-null source atom equal to $v$ supports that claim. A non-null atom with another label attacks it. Null atoms are irrelevant. We add no evidence-evidence edges and no source-specific weights.

Under this equal-weight, no evidence-edge adaptation, all evidence nodes have equal base strength. QE semantics is monotone increasing in supporters and decreasing in attackers. If $N$ sources report non-null atoms, candidate $v$ has $n_v$ supporters and $N-n_v$ attackers. A candidate with more votes therefore has weakly higher claim strength. Ties use the same seeded tie rule as Majority Vote. The adapted ArgRAG procedure therefore returns the plurality label for both ordinal and unordered categorical answer spaces.

\subsection{Verbatim Prompts}\label{subsec:C-4}

All four LLM families (T3) consume the same instruction-style prompt. Each persona is processed by a single chat-completion call: the system message contains the task instructions and the per-question answer-space specifications, the user message contains the persona's complete NL memory render (the same five-section document used by extraction, \S~\ref{sec:methods} and Appendix~\ref{subsec:C-3}), and the model returns one JSON object covering all 18 questions. We reproduce the exact prompt content below.

\subsubsection{LLM-Direct System Prompt (Verbatim)}

\begin{Verbatim}[fontsize=\scriptsize,frame=single,framesep=2mm,xleftmargin=2mm,breaklines=true,breakanywhere=true]
You will answer 18 survey questions about a single persona. The user message
contains the persona's complete natural-language memory document, organized
into five sections:

  ## Long-Term Background and Habits  (profile/background)
  ## Plans and Intentions             (planner)
  ## Daily Self-Reports               (daily journal)
  ## Objective Records                (timesheet/measurements)
  ## Device and Activity Records      (wearable/tracker)

For EACH of the 18 questions listed below:
  1. Read the question text carefully.
  2. Consider evidence from ALL relevant source sections.
  3. When sources disagree, use your judgment to determine the most likely
     true answer.
  4. Select exactly one answer label from the answer space (forced answer
     — you MUST pick one).
  5. Also decide: if you had the option to abstain (because evidence is too
     conflicting or insufficient for a confident judgment), would you?
     Record this as would_skip (true/false).

# Answering Principles
- Synthesize across sources. Different sources may tell different stories.
- No single source is presumed correct. Every source has potential biases.
- Forced answer required. Pick the best-supported answer even if ambiguous.
- The skip decision is separate from the forced answer.
- Use exact labels with underscores ("20_or_more", not "20 or more").
- Time windows matter (7, 14, or 30 days as specified per question).
- Cross-midnight times: treat 00:10 as 24:10.

# Output Format
Return a single JSON object with the schema given below. No prose outside
the JSON. No markdown fences.

# Survey Questions
Answer ALL 18 questions. For each question, select exactly one answer label
from the answer space.

## A1 — Sleep Duration Count (Sleep, 30 days)
In the past 30 days, how many nights did this person sleep 7 hours or more?
- fewer_than_10 (0–9 nights)
- 10_to_19 (10–19 nights)
- 20_or_more (20–30 nights)

## A2 — Long Work Days Count (Work, 30 days)
In the past 30 days, on how many days did this person work more than 9 hours?
- 0_to_3 (0–3 days)
- 4_to_7 (4–7 days)
- 8_or_more (8+ days)

## A3 — Home-Cooked Share (Diet, 30 days)
In the past 30 days, on what share of meal occasions did this person eat home-cooked food?
- less_than_40 (<40%
- 40_to_69 (40%
- 70_or_more (70%

## B2 — Exercise Frequency vs Profile (Exercise, 30 days)
This person's profile describes exercising a certain number of days per week.
In the past 30 days, how did their actual exercise frequency compare?
- more_than_1_below (>1 day/week less than profile)
- within_1_day (within +/-1 day/week of profile)
- more_than_1_above (>1 day/week more than profile)
- no_frequency_described (profile has no exercise frequency)

## B3 — Weekend Work vs Profile (Work, 30 days)
This person's profile describes a specific approach to weekend work.
In the past 30 days, did the actual weekend work pattern match that description?
- matches
- does_not_match
- no_approach_described

## C2 — Social Plan Realization (Social, 14 days)
In the past 14 days, on days when a social activity was planned, what share
of those plans were actually realized?
- below_25_pct (<25%
- 25_to_50_pct (25%
- above_50_pct (>50%
- no_plans (no social plans in planner)

## C3 — Bedtime vs Target (Sleep, 14 days)
In the past 14 days, on days when a target bedtime was set, did the person
go to bed within 20 minutes of that target?
- within_20min_more_than_50pct
- later_more_than_50pct
- earlier_more_than_50pct
- no_targets

## D1 — Social Trend (Social, 30 days)
Comparing the first 14 days to the last 16 days of the 30-day window: did
the number of social activities per day increase, decrease, or stay the same?
- decreased
- stayed_same
- increased

## D2 — Diet vs Baseline (Diet, 30 days)
In the past 30 days, did the person's daily home-cooked meal count and total
meal count differ from the profile averages by more than 1 meal/day?
- within_1
- differs_more_than_1
- no_baseline

## E1 — Late Night Cause (Sleep, 30 days)
In the past 30 days, on nights when bedtime was after midnight, which
co-occurring factor appeared on more than 50%
- work_activity (overtime/unusually long work hours on >50%
- social_activity (social activity present on >50%
- no_single_factor (neither factor exceeds 50%
- no_late_nights

## E2 — Exercise Skip Cause (Exercise, 30 days)
In the past 30 days, on days when exercise was planned but not done, did the
person work more than 8.5 hours that same day?
- no_fewer_than_30 (<30%
- between_30_60 (30%
- yes_more_than_60 (>60%

## F1 — Unplanned Social Count (Social, 30 days)
In the past 30 days, on how many days did the person attend social
activities when the planner showed no social intent?
- 0_to_3
- 4_to_6
- 7_or_more
- no_social_activities

## F2 — Tracker Missing vs Inactive (Exercise, 30 days)
In the past 30 days, on days when the fitness tracker recorded no workout,
did other sources indicate that the person exercised?
- inactive_confirmed
- both_occurred
- yes_tracker_missing

## F3 — Timesheet Missing vs Off (Work, 30 days)
In the past 30 days, on days when the work timesheet had no entry, did
other sources indicate that the person worked?
- truly_off
- both_occurred
- yes_worked_despite_no_entry

## G1 — Deliberate vs Incidental Exercise (Exercise, 30 days)
In the past 30 days, on days when the person was physically active, was the
activity deliberate exercise or incidental movement?
- incidental_movement_70plus
- mix
- deliberate_exercise_70plus
- no_activity

## G2 — Voluntary vs Obligatory Social (Social, 30 days)
In the past 30 days, when the person attended social activities, were those
activities voluntary or obligatory?
- obligatory_70plus
- mix
- voluntary_70plus
- no_meetings

## Ctrl1 — Outside Meals (Diet, 7 days)
In the past week, how many days did the person eat food prepared outside
the home?
- 0_to_1_days
- 2_to_3_days
- 4_or_more

## Ctrl2 — Short Sleep Nights (Sleep, 7 days)
In the past week, how many nights did the person sleep less than 6 hours?
- 0_nights
- 1_to_2
- 3_or_more

[Output schema: JSON object mapping each question ID to
{"answer": <one valid label>, "would_skip": <true_or_false>}.]
\end{Verbatim}

\subsubsection{Schema-Aware System Prompt (Delta from Direct)}

The Schema-Aware variant is identical to LLM-Direct except for two additions inserted into the system prompt: (i) a global Source Bias Characteristics table (Table~\ref{tab:E-1}), and (ii) per-question augmentations attached to each of the 18 question entries. Each per-question augmentation contains three fields: a Source Relevance row indicating which of the five sections are informative for the question, an optional Cross-Source Context Instruction (e.g., "compare against the planner's target bedtimes"), and a Reasoning Hint that paraphrases how a knowledgeable analyst would compute or judge the answer. The user message (the NL memory render) and the output schema are identical to LLM-Direct. These additions are qualitative descriptions of bias direction and reasoning structure only; they do not expose DGP bias magnitudes, dropout rates, learned confusion matrices, GT labels, train/test split statistics, or any artifacts produced during DSNBF fitting.

\begin{table}[t]
\centering
\caption{Source bias characteristics inserted into the Schema-Aware system prompt.}
\label{tab:E-1}
\small
\begin{tabularx}{\linewidth}{@{}lY@{}}
\toprule
Source & General Bias Pattern \\
\midrule
Profile/Background & May be outdated or idealized. People describe themselves as they want to be, not always as they are. Stated habits (exercise frequency, diet quality) are often more favorable than actual behavior. \\
Planner & Optimistic by nature — plans often exceed what actually happens. Social plans, exercise plans, and healthy eating intentions frequently go unrealized. \\
Daily Self-Report & Direction varies by topic. People tend to over-report diet quality and exercise effort, but under-report social isolation and work-related stress. Self-reports are most accurate for routine events but less reliable for unusual or embarrassing events. \\
Objective Records & Most factually accurate when present, but may have gaps (days with no entry). An absent record does NOT necessarily mean inactivity — the recording system itself may have failed. \\
Device/Activity Log & Precise when working, but prone to data dropout (some days may have no data). A missing device record should not be interpreted as evidence of inactivity. \\
\bottomrule
\end{tabularx}
\end{table}

Schema-Aware per-question augmentations. Summary of the per-question prompt block. Src columns: Pr=profile\_ltm, Pl=planner, Sr=daily\_self\_report, Ob=objective\_log, Dv=device\_log ($\checkmark$ = relevant). Ctx = cross-source context instruction (blank if none). Hint = computation/judgment hint summarized from the prompt.

\begingroup\small
\begin{xltabular}{\linewidth}{@{}lcccccYY@{}}
\toprule
Q & Pr & Pl & Sr & Ob & Dv & Ctx & Hint (summary) \\
\midrule
\endfirsthead
\toprule
Q & Pr & Pl & Sr & Ob & Dv & Ctx & Hint (summary) \\
\midrule
\endhead
\bottomrule
\endfoot
A1 & $\checkmark$ & $\checkmark$ & $\checkmark$ &  & $\checkmark$ & — & Count nights with sleep duration $\geq$ 7h in 30d. Classify into fewer\_than\_10, 10\_to\_19, or 20\_or\_more. For device data with missing days, extrapolate to 30 days. Profile avg $\geq$ 7h $\to$ most nights $\geq$ 7h. When sources disagree, use bias characteristics to judge. \\
A2 & $\checkmark$ & $\checkmark$ & $\checkmark$ & $\checkmark$ & $\checkmark$ & — & Count days with work hours > 9 in 30d, including worked weekends; do not count non-work weekend days as long-work days. Use day-of-week only to interpret weekend context. \\
A3 & $\checkmark$ & $\checkmark$ & $\checkmark$ & $\checkmark$ &  & — & Calculate share of home-cooked meals among all meals in 30d. For objective\_log, use food-delivery payment frequency as inverse proxy. \\
B2 & $\checkmark$ & $\checkmark$ & $\checkmark$ & $\checkmark$ & $\checkmark$ & Profile states exercise frequency as N days/week; compare observed frequency from other sources. & Compute observed days/week from behavioral sources over 30d. Compare to profile. Classify more\_than\_1\_below / within\_1\_day / more\_than\_1\_above / no\_frequency\_described. \\
B3 & $\checkmark$ &  & $\checkmark$ & $\checkmark$ & $\checkmark$ & Profile describes after-hours work style; check whether other sources match it. & Use day-of-week to identify weekend days. For self-report: look for explicit "worked on weekend" markers. For device\_log: check work activity on weekend days. \\
C2 &  & $\checkmark$ & $\checkmark$ & $\checkmark$ &  & Identify planned social days from planner; check realization in other sources. & Identify days with social plans (last 14d). For each planned day, check whether self-report or objective records show social activity. Classify below\_25\_pct / 25\_to\_50\_pct / above\_50\_pct / no\_plans. \\
C3 &  &  & $\checkmark$ &  & $\checkmark$ & Compare against planner target bedtimes. & Last 14d: compare actual bedtime to planner target. Cross-midnight: treat 00:10 as 24:10. Within 20 min = compliant. \\
D1 &  & $\checkmark$ & $\checkmark$ & $\checkmark$ &  & — & Split 30d window: first 14 vs last 16. Count activities in each half and compute rate (count/days). Small fluctuations $\to$ stayed\_same. \\
D2 & $\checkmark$ &  & $\checkmark$ &  &  & Compare against profile meal baseline. & Estimate 30-day average daily meals and home-cooked meals from available behavioral evidence, accounting for self-report bias. Compare to profile baseline. Combined deviation > 1 meal/day $\to$ differs\_more\_than\_1. \\
E1 &  & $\checkmark$ & $\checkmark$ & $\checkmark$ & $\checkmark$ & — & Find nights with bedtime after midnight. For each, check overtime (>8.5h work) or social activity that day. Regular work (non-overtime) does NOT count. \\
E2 & $\checkmark$ & $\checkmark$ & $\checkmark$ & $\checkmark$ & $\checkmark$ & Identify planned exercise days from planner; check skip reasons. & Find days when exercise was planned but not done (self-report/device show no exercise). On those skip days, check work hours > 8.5h. \\
F1 & $\checkmark$ & $\checkmark$ & $\checkmark$ & $\checkmark$ &  & Compare social activities against planner social intent. & Count days with social activity (self-report/objective) where the planner showed NO social intent. \\
F2 & $\checkmark$ & $\checkmark$ & $\checkmark$ & $\checkmark$ & $\checkmark$ & Compare exercise evidence from all sources against device\_log coverage. & Focus on days when device log shows "Records unavailable" or no workout detected. If self-report or other sources show exercise on those days $\to$ tracker missed it. \\
F3 & $\checkmark$ & $\checkmark$ & $\checkmark$ &  & $\checkmark$ & Compare work evidence from all sources against objective\_log availability. & Focus on days when objective log shows "Records unavailable". If self-report shows work hours > 0, or device/activity log shows >30 min focused work $\to$ timesheet was just missing. \\
G1 &  & $\checkmark$ & $\checkmark$ &  & $\checkmark$ & — & Distinguish planned workouts from incidental movement. Duration can be a cue, but intent language and workout markers are more direct evidence of deliberate exercise. \\
G2 & $\checkmark$ & $\checkmark$ & $\checkmark$ & $\checkmark$ &  & — & OBLIGATORY = supporting someone else or attending family/work obligations. VOLUNTARY = personal choice (dinner with friends, club, coffee). Weight self-report for activity descriptions, planner for intent. \\
Ctrl1 & $\checkmark$ & $\checkmark$ & $\checkmark$ &  &  & — & Last 7d: count days with at least one meal eaten outside the home. \\
Ctrl2 &  & $\checkmark$ & $\checkmark$ &  & $\checkmark$ & — & Last 7d: count nights with sleep duration < 6 hours. \\
\end{xltabular}
\endgroup

The table above summarizes the per-question Schema-Aware prompt block; the full prompt source files are released alongside the dataset.

\subsubsection{Output Schema (Verbatim)}

Both LLM-Direct and Schema-Aware return one JSON object per persona with the same schema:

\begin{Verbatim}[fontsize=\scriptsize,frame=single,framesep=2mm,xleftmargin=2mm,breaklines=true,breakanywhere=true]
{
  "answers": {
    "A1": {"answer": "<label>", "would_skip": <true|false>},
    ...
    "Ctrl2": {"answer": "<label>", "would_skip": <true|false>}
  }
}
\end{Verbatim}

Acceptance rules: (i) \texttt{answers} has exactly 18 keys; (ii) \texttt{answer} is always a valid label from that question's answer space (never null, even when \texttt{would\_skip} is true); (iii) labels use exact underscored strings; (iv) the JSON validates against the schema before any downstream evaluation. Records that fail validation are not coerced into labels; they are tracked as missing-output errors (\S~\ref{subsec:C-5}).

\subsubsection{Few-Shot Variant (Supplementary)}

The few-shot experiment on seed 1 in \S~\ref{s4:sub1} augments the Direct prompt with $k=3$ exemplars per question, drawn from the training split via a frozen, deterministic rule:

\begin{Verbatim}[fontsize=\scriptsize,frame=single,framesep=2mm,xleftmargin=2mm,breaklines=true,breakanywhere=true]
For each question qid, partition the 216 training personas by difficulty class
(stable / temporal_shift / stated_vs_revealed). Within each class, sort by
sha256("{qid}|{seed}|{class}|{persona_id}"). Pick the first persona whose GT
answer differs from already-selected exemplars; if all share the same label,
fall back to the first by hash order. This yields 3 exemplars per question
covering the 3 difficulty classes with label-diversity tie-breaking.
\end{Verbatim}

The class labels and persona identifiers in this rule are consumed only by the deterministic exemplar-selection script. The prompt shown to the model redacts exemplar IDs and difficulty labels, and the target persona's ID is never included. Each exemplar consists of the full NL memory of a training persona followed by the question text and the final GT answer label. We do not provide a separate worked-solution trace, model-generated rationale, or skip demonstration; the model can still infer the reasoning pattern from the memory--label pairs. The selection rule was frozen before the pilot run and applied identically across all 18 questions on seed 1.

\subsubsection{Extraction Task Template}

The shared extraction stage produces one frozen atom bundle per persona. The task reads the persona's complete five-section NL memory render together with the extraction specification: the source-question map, context map, answer spaces, edge-case rules, and per-question extraction hints. It then writes all source-question atoms in one JSON file. The operative task template is:

\begin{Verbatim}[fontsize=\scriptsize,frame=single,framesep=2mm,xleftmargin=2mm,breaklines=true,breakanywhere=true]
SYSTEM: <empty>

USER:
Read the extraction specification and the NL memory render for the current persona.
The memory render contains five source sections:
  profile_ltm, planner, daily_self_report, objective_log, device_log.

Extract mu(source, question) for ALL 18 questions x 5 sources.
Follow the source-question map exactly:
  - if a source is not mapped to a question, write null;
  - if a mapped source has no relevant evidence, write null;
  - otherwise write exactly one valid answer label from that question's
    answer space.

For questions with a context-map entry, read the specified context source
only as instructed. Do not otherwise merge evidence across sources.

Write a JSON object whose top-level keys are question IDs:
{
  "A1": {
    "profile_ltm": <label_or_null>,
    "planner": <label_or_null>,
    "daily_self_report": <label_or_null>,
    "objective_log": <label_or_null>,
    "device_log": <label_or_null>
  },
  ...
}
\end{Verbatim}

Each cell contains either a label from that question's answer space or \texttt{null}. Null observations are passed to fusion methods as missing inputs and are never coerced into a guess. The extraction task is factual recovery only: it reports what each source says under the allowed context rules and does not resolve conflicts or access GT labels.

\subsection{Inference Parameters}\label{subsec:C-5}

All LLM calls use \texttt{temperature = 0.0}; no maximum-output budget is tuned or reported as a method setting. Reasoning-capable models use the serving API's native control when available: OpenAI calls set \texttt{reasoning\_effort=xhigh}, Gemini calls set \texttt{thinking\_level}/\texttt{thinkingLevel}=\texttt{high}, and DeepSeek V3.2 and Qwen3 calls omit reasoning-control fields. Other decoding parameters (\texttt{top\_p}, \texttt{top\_k}, \texttt{stop}, \texttt{seed}) are left at the API default; we do not specify them in our requests.

Calls are deterministically cached by \texttt{SHA256(model | system\_prompt | user\_prompt)} and replayed from disk across re-evaluations, so reported numbers are exactly reproducible from the released cache. Failed or empty responses are retried with exponential backoff up to 6 attempts; persistent failures are logged and carried as missing-output errors.

\begin{table}[t]
\centering
\caption{Inference parameters per role. T3 calls return all 18 answers in a single JSON object.}
\label{tab:E-2}
\small
\setlength{\tabcolsep}{2pt}
\begin{tabularx}{\linewidth}{@{}YYccYl@{}}
\toprule
Role & Model identifier (vendor) & Temp. & \shortstack{Reasoning /\\ thinking control} & Retry policy & Cache key \\
\midrule
Extraction & gpt-5.4 (OpenAI) & 0.0 & \texttt{reasoning\_effort=xhigh} & 6 attempts, exponential backoff & SHA256(model, sys, user) \\
LLM-Direct (T3) & gpt-5.4 (OpenAI) & 0.0 & \texttt{reasoning\_effort=xhigh} & 6 attempts, exponential backoff & SHA256(model, sys, user) \\
LLM-Direct (T3) & gemini-3.1-pro-preview (Google Gemini API) & 0.0 & \texttt{thinking\_level=high} & 6 attempts, exponential backoff & SHA256(model, sys, user) \\
LLM-Direct (T3) & deepseek-v3.2 (DeepSeek) & 0.0 & omitted/not set & 6 attempts, exponential backoff & SHA256(model, sys, user) \\
LLM-Direct (T3) & qwen3-235b-a22b-2507 (Alibaba) & 0.0 & omitted/not set & 6 attempts, exponential backoff & SHA256(model, sys, user) \\
Schema-Aware (T3) & same four models & 0.0 & (same as Direct) & (same as Direct) & SHA256(model, sys, user) \\
Few-shot (\S~\ref{s4:sub1}) & gpt-5.4 (OpenAI) & 0.0 & \texttt{reasoning\_effort=xhigh} & 6 attempts, exponential backoff & SHA256(model, sys, user) \\
Cross-extractor check (Appendix~\ref{subsec:F-6}) & gemini-3.1-pro-preview (Google Gemini API) & 0.0 & \texttt{thinking\_level=high} & 6 attempts, exponential backoff & SHA256(model, sys, user) \\
\bottomrule
\end{tabularx}
\end{table}

Vendor-specific notes: GPT-5.4 uses OpenAI's \texttt{reasoning\_effort} parameter, while Gemini 3.1 Pro Preview uses \texttt{thinking\_level} in the released config and \texttt{thinkingLevel} in the Gemini API REST field. DeepSeek V3.2 and Qwen3-235B-A22B-Instruct-2507 omit reasoning-control fields. We did not tune these controls per family; the configuration is identical across the four LLMs apart from the model identifier and API-specific reasoning/thinking controls.

\subsection{Compute Footprint}
\label{subsec:compute-footprint}

The original LLM stages used vendor APIs, but the released reproduction path replays frozen outputs and makes no live API calls. On an Apple M1 Pro laptop with 16 GB RAM, the full released reproduction takes about two hours, including structured fusion, parameter fitting, threshold calibration, and bootstrap confidence intervals.

Table~\ref{tab:api_inventory} reports the original LLM API workload (4 seeds $\times$ 480 personas for extraction; 4 seeds $\times$ 120 test personas for resolution, except where noted).

\begin{table}[t]
\centering
\caption{API call inventory and measured token summary.}
\label{tab:api_inventory}
\footnotesize
\begin{tabularx}{\linewidth}{@{}YYrr@{}}
\toprule
Stage & Provider / API & Calls & \multicolumn{1}{r}{\shortstack[r]{Avg.\ input / output tok.\ per call\\(incl.\ reasoning tok.\ if any)}} \\
\midrule
GPT-5.4 extraction (T1/T2 input) & OpenAI API & 1{,}920 & 12{,}922 / 42{,}870 \\
Gemini 3.1 Pro extraction (cross-extractor robustness, Appendix~\ref{subsec:F-6}) & Google API & 1{,}920 & 15{,}315 / 16{,}771 \\
GPT-5.4 LLM-Direct + Schema-Aware (T3) & OpenAI API & 960 & 9{,}838 / 45{,}620 \\
GPT-5.4 on atoms ($\hat\mu$ + $\mu^\ast$, \S~\ref{s4:sub2}) & OpenAI API & 960 & 3{,}636 / 14{,}862 \\
Gemini 3.1 Pro LLM-Direct + Schema-Aware (T3) & Google API & 960 & 13{,}347 / 11{,}564 \\
DeepSeek V3.2 LLM-Direct + Schema-Aware (T3) & OpenRouter API & 960 & 11{,}252 / 464 \\
Qwen3-235B LLM-Direct + Schema-Aware (T3) & OpenRouter API & 960 & 12{,}361 / 408 \\
Few-shot pilot (seed 1, per-question, App.~\ref{subsec:D-3}) & OpenAI API & 2{,}160 & 30{,}373 / 8{,}946 \\
\midrule
\textbf{Total} &  & \textbf{10{,}800} & \textbf{--} \\
\bottomrule
\end{tabularx}
\end{table}

Measured token values are reported where exported usage logs are available. For the Gemini extraction row, the released logs cover 480 observed test-split calls; the table reports the measured per-call token averages from those calls, while the call inventory reflects the paper's full 1{,}920-call workload. OpenAI token values are measured from the successful completed calls in the released 33-call batch sample; the matched-input row pools GPT calls over $\hat\mu$ and $\mu^\ast$ atom inputs. The table counts original API calls, not the number of JSON files in the released cache. The cache stores the outputs needed for API-free reproduction: test-split GPT atoms under \texttt{extracted\_atoms/}, Gemini cross-extractor atoms under \texttt{method\_outputs/gemini\_p2/*/extract/}, frozen T3/few-shot/structured-LLM predictions under \texttt{method\_outputs/}, and deterministic aggregate JSONs under \texttt{benchmark/results/}.

Each extraction call processes a single persona's complete NL memory and returns structured atoms $\hat\mu(\text{source}, \text{question})$ for all 18 questions $\times$ 5 sources in one structured-output JSON. Each LLM-Direct, Schema-Aware, and matched input GPT-5.4 resolution call processes a single persona and returns all 18 answers and per-question \texttt{would\_skip} flags in one JSON. Few-shot calls are issued per (persona, question) to allow per-question demonstration selection. All calls are deterministically cached by SHA256(model $|$ system\_prompt $|$ user\_prompt); re-running with unchanged inputs incurs no additional billed API usage.

\section{Selective QA Details and Few-Shot Supplementary}\label{app:selective}
\subsection{Full Selective QA Table}\label{subsec:D-1}

Table~\ref{tab:D-1} reports selective accuracy and coverage for all deployable methods with SKIP (4-seed pooled), plus the Source Reachability reference row. Sel. Acc and Cov report each method under its canonical input (GPT-5.4-extracted $\hat\mu$ for T1/T2, NL for T3). Fusion methods are additionally evaluated under $\mu^\ast$ input. Figure~\ref{fig:3} visualizes the deployable SKIP methods; Source Reachability stays table-only as an interpretive reference (\S~\ref{s4:sub1}).

\begin{table}[t]
\centering
\caption{Selective QA results. The Sel. Acc and Cov columns report each deployable method under its canonical input. For T1/T2, the last two columns report selective accuracy and coverage on $\mu^\ast$. The Source Reachability row is a GT-aided direct-readout reference for interpretation only.}
\label{tab:D-1}
\small
\begin{tabular}{@{}llcccc@{}}
\toprule
 & Method & Sel. Acc (\%) & Cov (\%) & Sel. Acc on $\mu^\ast$ (\%) & Cov on $\mu^\ast$ (\%) \\
\midrule
\textbf{T1} & SSB & 80.8 & 82.9 & 81.6 & 83.6 \\
\textbf{T2} & NBF & \textbf{85.9} & 77.5 & 88.5 & 77.9 \\
 & DSNBF & 85.3 & 78.3 & \textbf{88.8} & 77.2 \\
 & ABF & 83.0 & 62.0 & 83.4 & 60.2 \\
\textbf{T3} & GPT-5.4 Direct & 70.3 & 92.3 & — & — \\
 & GPT-5.4 Schema & 71.0 & 95.4 & — & — \\
 & Gemini 3.1 Direct & 68.4 & 98.5 & — & — \\
 & Gemini 3.1 Schema & 70.4 & 99.2 & — & — \\
 & DeepSeek V3.2 Direct & 56.1 & 100.0 & — & — \\
 & DeepSeek V3.2 Schema & 56.9 & 100.0 & — & — \\
 & Qwen3 235B Direct & 42.2 & 99.3 & — & — \\
 & Qwen3 235B Schema & 48.0 & 98.8 & — & — \\
\textbf{Ref.} & Source Reachability & — & — & 100.0 & 93.2 \\
\bottomrule
\end{tabular}
\end{table}

\subsection{Selective QA Analysis}\label{subsec:D-2}

Fusion methods achieve high selective accuracy at moderate coverage. DSNBF reaches 85.3\% selective accuracy while covering 78.3\% of the testbed; under $\mu^\ast$ input this climbs to 88.8\%. All fusion SKIP thresholds were calibrated by maximizing $F_{0.5}$ on the held-out calibration split (\S~\ref{sec:benchmark} and Appendix~\ref{subsec:C-2}). Test set $F_{0.5}$ (answer precision = selective accuracy, answer recall = fraction of answer-only correct instances that are answered) confirms effective calibration transfer: NBF 85.3, DSNBF 84.9, SSB 81.9, ABF 80.4. LLM self-reported abstention yields lower $F_{0.5}$: GPT-5.4 Schema 75.0, GPT-5.4 Direct 74.1.

On the 78.3\% of instances DSNBF chooses to answer, it achieves 85.3\%, compared to 80.3\% in answer-only mode. The \~{}22\% of skipped instances are disproportionately difficult, consistent with the posterior identifying harder cases. ABF achieves comparable selective accuracy (83.0\%) but at much lower coverage (62.0\%); its calibration is more conservative. Moving from extracted input to $\mu^\ast$ input improves selective accuracy for all fusion methods, with DSNBF gaining \~{}3.5 pp and NBF \~{}2.6 pp, while ABF gains only 0.4 pp and answers slightly fewer instances.

LLM methods abstain rarely and gain little from doing so. GPT-5.4 Direct covers 92.3\% but achieves only 70.3\% selective accuracy, barely above its answer-only macro accuracy of 68.6\%. GPT-5.4 Schema shows the same pattern: 71.0\% at 95.4\% coverage. Gemini abstains even less: Gemini Direct covers 98.5\% at 68.4\% selective accuracy. The open-weight families push this pattern further: DeepSeek V3.2 never abstains (100.0\% coverage on both Direct and Schema), while Qwen3 abstains on fewer than 1.2\% of instances.

\paragraph{Protocol asymmetry.} Fusion methods use thresholds tuned on a held-out calibration split (\S~\ref{sec:methods}); LLM methods return a self-reported binary \texttt{would\_skip} flag. We therefore read Table~\ref{tab:D-1} as a comparison of the interfaces evaluated here: fusion exposes posterior margins that can be calibrated after training, whereas the prompt-only interface gives a skip decision without a continuous confidence score. In deployment settings where $\mathrm{Cost}(\mathrm{wrong}) \gg \mathrm{Cost}(\mathrm{skip})$, that tunable uncertainty interface is a practical advantage.

Source Reachability is the only reference row. Its 93.2\% value is the GT-aided direct-readout reachability rate: the fraction of instances where some $\mu^\ast$ source atom exactly matches GT. The complementary 6.8\% has no exact visible GT match, but learned inference can still recover some of those cases. Table~\ref{tab:D-1} keeps the row next to selective methods only as an interpretive reference; its structure by type is analyzed in \S~\ref{s4:sub3} and Appendix~\ref{subsec:E-1}.

\subsection{Few-Shot Supplementary Check}\label{subsec:D-3}

A few-shot variant run on seed 1 (GPT-5.4, 3 exemplars per question) achieves 75.4\% answer-only accuracy, the highest among all NL-input LLM configurations in our experiments. The +6.5 pp gain over GPT-5.4 Direct (seed 1: 68.9\%) is the largest prompt-level improvement among NL-input configurations. The result still trails the main DSNBF result (80.3\%, 4-seed), with the seed-scope caveat, so we treat it as a supplementary consistency check; the primary finding remains the four-seed comparison.

This prompt-side check also clarifies the scope of stronger prompting. The locked scoring format requests only labels and skip flags; we did not add externalized chain-of-thought traces. GPT-5.4 and Gemini 3.1 Pro Preview already use vendor-maximum reasoning/thinking controls. Self-consistency would multiply LLM calls and falls outside the cost-matched single-call regime. The matched $\mu^\ast$ input condition tests direct source readout; tools that compute new statistics would define a different information condition. In the structured-input setting evaluated here, GPT-5.4 trails fusion by 8.9 pp.

The few-shot variant contains no abstention demonstrations, so the model relies solely on written instructions for skip decisions. It produces a 2.6\% skip rate, yielding 76.1\% selective accuracy at 97.4\% coverage, only 0.7 pp above its answer-only accuracy. Among the skipped instances, answer-only accuracy drops to 46.4\%, suggesting that the model identifies some difficult cases but abstains too rarely to materially improve selective performance. The largest remaining gaps to DSNBF occur on F-Missing-Data (63.1\% vs. 77.9\%, $-$14.8 pp) and G-Annotation (53.8\% vs. 63.5\%, $-$9.7 pp), the two types where examples provide weaker support for bias-aware reasoning.

\section{Extended Diagnostic Tables}\label{app:diagnostic}
\subsection{Per-Type Accuracy}\label{subsec:E-1}

Table~\ref{tab:E1} expands Table~\ref{tab:6} with the full set of 13 comparison columns: three T2 fusion methods (DSNBF, NBF, SSB), the matched input GPT reference (GPT-$\mu^\ast$, GPT-5.4 reading the same $\mu^\ast$ atoms as structured prompt input), eight T3 LLM settings (four families $\times$ two prompting regimes), and the Source Reachability reference by type. Rows are ordered by DSNBF accuracy descending. Bold marks the best method entry per type.

\begin{table}[t]
\centering
\caption{Full per-type macro accuracy (\%, answer-only mode, 4-seed pooled). In this diagnostic table, T1/T2 methods use $\mu^\ast$ input. Most T3 methods use NL input; GPT-$\mu^\ast$ is the matched input GPT-5.4 comparator on $\mu^\ast$. Bold = best method entry per type. Types ordered by DSNBF accuracy descending. GPT = GPT-5.4; Gem = Gemini 3.1; DS = DeepSeek V3.2; QW = Qwen3 235B. D = Direct; S = Schema. The T3 columns are the per-type marginal of Table~\ref{tab:E2b}. The T2 columns are matched-input $\mu^\ast$ summaries computed separately; Table~\ref{tab:E2a} reports the canonical GPT-extracted $\hat\mu$ T2 per-difficulty grid. We keep this summary table for navigation and to expose the Source Reachability reference and T1 column not shown in the per-difficulty tables.}
\label{tab:E1}
\scriptsize
\setlength{\tabcolsep}{2.5pt}
\begin{tabular}{@{}lcccccccccccccc@{}}
\toprule
Type & \# Q & DSNBF & NBF & SSB & GPT-$\mu^\ast$ & GPT-S & GPT-D & Gem-D & Gem-S & DS-D & DS-S & QW-D & QW-S & Ref. \\
\midrule
B $\cdot$ Ident & 2 & 96.8 & 96.8 & 96.4 & \textbf{97.0} & 94.5 & 81.9 & 71.6 & 81.4 & 77.4 & 74.0 & 58.1 & 55.4 & 99.7 \\
A $\cdot$ Arbit & 3 & \textbf{87.5} & 86.9 & 76.7 & 68.5 & 73.3 & 73.2 & 72.0 & 75.1 & 60.9 & 61.0 & 49.6 & 53.4 & 94.1 \\
Ctrl & 2 & 85.7 & 86.1 & \textbf{86.5} & 75.7 & 76.5 & 79.1 & 76.2 & 77.7 & 57.0 & 51.7 & 48.5 & 47.2 & 96.0 \\
E $\cdot$ Factor & 2 & \textbf{83.1} & 80.9 & 76.6 & 70.6 & 70.9 & 70.8 & 69.6 & 70.2 & 55.7 & 59.7 & 10.4 & 20.5 & 95.0 \\
D $\cdot$ Temp & 2 & \textbf{82.2} & 77.7 & 76.7 & 76.2 & 68.6 & 72.5 & 70.7 & 68.1 & 61.5 & 66.5 & 52.3 & 54.8 & 90.0 \\
C $\cdot$ P-R & 2 & 81.6 & \textbf{82.8} & 81.9 & 73.4 & 69.4 & 69.8 & 69.8 & 69.3 & 46.5 & 48.2 & 57.9 & 57.8 & 94.8 \\
F $\cdot$ Miss & 3 & 77.9 & \textbf{78.6} & 78.3 & 71.5 & 68.5 & 64.2 & 64.3 & 67.4 & 43.5 & 47.8 & 25.8 & 46.0 & 92.0 \\
G $\cdot$ Annot & 2 & 63.5 & \textbf{65.7} & 60.8 & 58.5 & 34.4 & 37.5 & 47.9 & 49.9 & 49.9 & 48.6 & 38.5 & 47.0 & 83.7 \\
\bottomrule
\end{tabular}
\end{table}

Figure~\ref{fig:pertype-heatmap} gives a colored-table view of the same values, so type-specific failure signatures can be read without scanning the full numeric table.
\definecolor{heatLow}{HTML}{F4F6F8}
\definecolor{heatHigh}{HTML}{6BAED6}
\definecolor{heatRef}{HTML}{009E73}

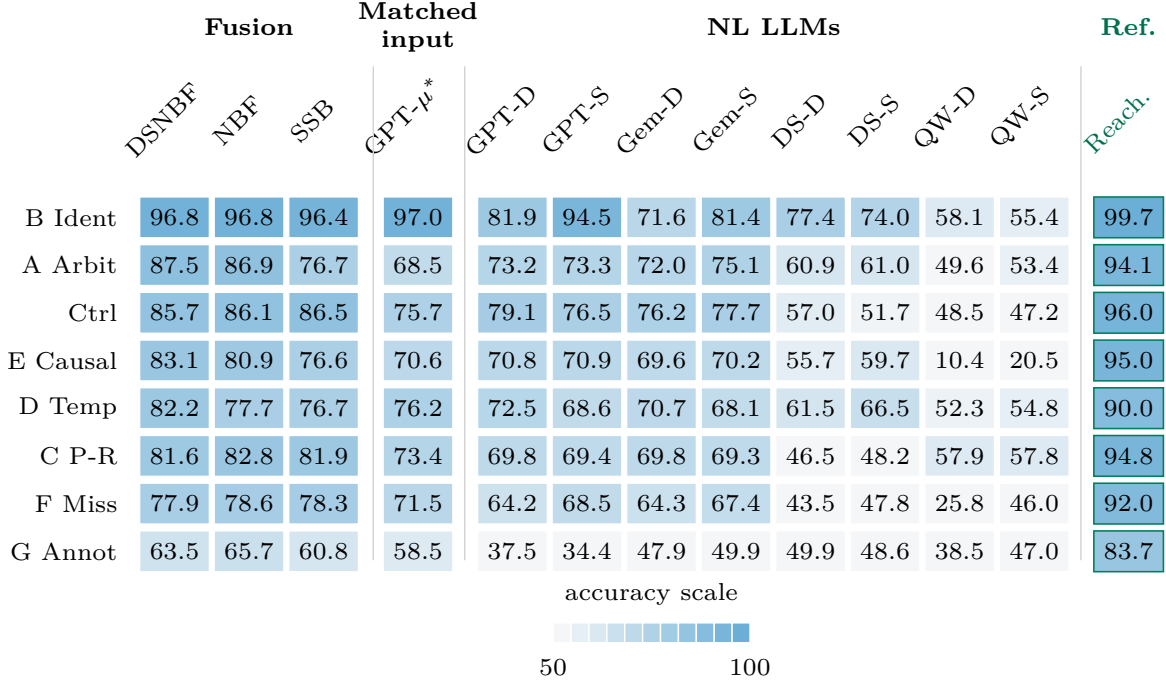
\begin{figure}
\centering
\noindent\makebox[\linewidth][c]{%
\resizebox{0.94\linewidth}{!}{%
\begin{tikzpicture}[
  every node/.style={font=\scriptsize},
  heatcell/.style={rectangle, minimum width=7.0mm, minimum height=4.1mm,
                   inner sep=0pt, draw=white, line width=0.35pt, text=black},
  refcell/.style={rectangle, minimum width=7.0mm, minimum height=4.1mm,
                  inner sep=0pt, draw=heatRef!80!black, line width=0.45pt, text=black},
  grouplabel/.style={font=\scriptsize\bfseries, align=center},
  collabel/.style={font=\scriptsize, rotate=45, anchor=south, align=center, xshift=0.45mm},
  rowlabel/.style={font=\scriptsize, anchor=east},
]

\newcommand{\pthmcell}[4]{%
  \pgfmathtruncatemacro{\mix}{round(max(0,min(100,(#3-50)*2)))}%
  \node[heatcell, fill=heatHigh!\mix!heatLow] at (#1,#2) {#4};%
}
\newcommand{\pthmrefcell}[4]{%
  \pgfmathtruncatemacro{\mix}{round(max(0,min(100,(#3-50)*2)))}%
  \node[refcell, fill=heatHigh!\mix!heatLow] at (#1,#2) {#4};%
}

\def\xD{2.00}
\def\xN{2.75}
\def\xS{3.50}
\def\xGm{4.45}
\def\xGD{5.40}
\def\xGS{6.15}
\def\xGeD{6.90}
\def\xGeS{7.65}
\def\xDSD{8.40}
\def\xDSS{9.15}
\def\xQWD{9.90}
\def\xQWS{10.65}
\def\xRef{11.60}

\node[grouplabel] at (2.75,1.22) {Fusion};
\node[grouplabel] at (4.45,1.22) {Matched\\input};
\node[grouplabel] at (8.03,1.22) {NL LLMs};
\node[grouplabel, text=heatRef!70!black] at (11.60,1.22) {Ref.};

\node[collabel] at (\xD,0.10) {DSNBF};
\node[collabel] at (\xN,0.10) {NBF};
\node[collabel] at (\xS,0.10) {SSB};
\node[collabel] at (\xGm,0.10) {GPT-$\mu^\ast$};
\node[collabel] at (\xGD,0.10) {GPT-D};
\node[collabel] at (\xGS,0.10) {GPT-S};
\node[collabel] at (\xGeD,0.10) {Gem-D};
\node[collabel] at (\xGeS,0.10) {Gem-S};
\node[collabel] at (\xDSD,0.10) {DS-D};
\node[collabel] at (\xDSS,0.10) {DS-S};
\node[collabel] at (\xQWD,0.10) {QW-D};
\node[collabel] at (\xQWS,0.10) {QW-S};
\node[collabel, text=heatRef!70!black] at (\xRef,0.10) {Reach.};

\draw[gray!35] (4.00,0.82) -- (4.00,-4.15);
\draw[gray!35] (4.92,0.82) -- (4.92,-4.15);
\draw[gray!35] (11.12,0.82) -- (11.12,-4.15);

\node[rowlabel] at (1.55,-0.70) {B Ident};
\node[rowlabel] at (1.55,-1.18) {A Arbit};
\node[rowlabel] at (1.55,-1.66) {Ctrl};
\node[rowlabel] at (1.55,-2.14) {E Causal};
\node[rowlabel] at (1.55,-2.62) {D Temp};
\node[rowlabel] at (1.55,-3.10) {C P-R};
\node[rowlabel] at (1.55,-3.58) {F Miss};
\node[rowlabel] at (1.55,-4.06) {G Annot};

\pthmcell{\xD}{-0.70}{96.8}{96.8}
\pthmcell{\xN}{-0.70}{96.8}{96.8}
\pthmcell{\xS}{-0.70}{96.4}{96.4}
\pthmcell{\xGm}{-0.70}{97.0}{97.0}
\pthmcell{\xGD}{-0.70}{81.9}{81.9}
\pthmcell{\xGS}{-0.70}{94.5}{94.5}
\pthmcell{\xGeD}{-0.70}{71.6}{71.6}
\pthmcell{\xGeS}{-0.70}{81.4}{81.4}
\pthmcell{\xDSD}{-0.70}{77.4}{77.4}
\pthmcell{\xDSS}{-0.70}{74.0}{74.0}
\pthmcell{\xQWD}{-0.70}{58.1}{58.1}
\pthmcell{\xQWS}{-0.70}{55.4}{55.4}
\pthmrefcell{\xRef}{-0.70}{99.7}{99.7}

\pthmcell{\xD}{-1.18}{87.5}{87.5}
\pthmcell{\xN}{-1.18}{86.9}{86.9}
\pthmcell{\xS}{-1.18}{76.7}{76.7}
\pthmcell{\xGm}{-1.18}{68.5}{68.5}
\pthmcell{\xGD}{-1.18}{73.2}{73.2}
\pthmcell{\xGS}{-1.18}{73.3}{73.3}
\pthmcell{\xGeD}{-1.18}{72.0}{72.0}
\pthmcell{\xGeS}{-1.18}{75.1}{75.1}
\pthmcell{\xDSD}{-1.18}{60.9}{60.9}
\pthmcell{\xDSS}{-1.18}{61.0}{61.0}
\pthmcell{\xQWD}{-1.18}{49.6}{49.6}
\pthmcell{\xQWS}{-1.18}{53.4}{53.4}
\pthmrefcell{\xRef}{-1.18}{94.1}{94.1}

\pthmcell{\xD}{-1.66}{85.7}{85.7}
\pthmcell{\xN}{-1.66}{86.1}{86.1}
\pthmcell{\xS}{-1.66}{86.5}{86.5}
\pthmcell{\xGm}{-1.66}{75.7}{75.7}
\pthmcell{\xGD}{-1.66}{79.1}{79.1}
\pthmcell{\xGS}{-1.66}{76.5}{76.5}
\pthmcell{\xGeD}{-1.66}{76.2}{76.2}
\pthmcell{\xGeS}{-1.66}{77.7}{77.7}
\pthmcell{\xDSD}{-1.66}{57.0}{57.0}
\pthmcell{\xDSS}{-1.66}{51.7}{51.7}
\pthmcell{\xQWD}{-1.66}{48.5}{48.5}
\pthmcell{\xQWS}{-1.66}{47.2}{47.2}
\pthmrefcell{\xRef}{-1.66}{96.0}{96.0}

\pthmcell{\xD}{-2.14}{83.1}{83.1}
\pthmcell{\xN}{-2.14}{80.9}{80.9}
\pthmcell{\xS}{-2.14}{76.6}{76.6}
\pthmcell{\xGm}{-2.14}{70.6}{70.6}
\pthmcell{\xGD}{-2.14}{70.8}{70.8}
\pthmcell{\xGS}{-2.14}{70.9}{70.9}
\pthmcell{\xGeD}{-2.14}{69.6}{69.6}
\pthmcell{\xGeS}{-2.14}{70.2}{70.2}
\pthmcell{\xDSD}{-2.14}{55.7}{55.7}
\pthmcell{\xDSS}{-2.14}{59.7}{59.7}
\pthmcell{\xQWD}{-2.14}{10.4}{10.4}
\pthmcell{\xQWS}{-2.14}{20.5}{20.5}
\pthmrefcell{\xRef}{-2.14}{95.0}{95.0}

\pthmcell{\xD}{-2.62}{82.2}{82.2}
\pthmcell{\xN}{-2.62}{77.7}{77.7}
\pthmcell{\xS}{-2.62}{76.7}{76.7}
\pthmcell{\xGm}{-2.62}{76.2}{76.2}
\pthmcell{\xGD}{-2.62}{72.5}{72.5}
\pthmcell{\xGS}{-2.62}{68.6}{68.6}
\pthmcell{\xGeD}{-2.62}{70.7}{70.7}
\pthmcell{\xGeS}{-2.62}{68.1}{68.1}
\pthmcell{\xDSD}{-2.62}{61.5}{61.5}
\pthmcell{\xDSS}{-2.62}{66.5}{66.5}
\pthmcell{\xQWD}{-2.62}{52.3}{52.3}
\pthmcell{\xQWS}{-2.62}{54.8}{54.8}
\pthmrefcell{\xRef}{-2.62}{90.0}{90.0}

\pthmcell{\xD}{-3.10}{81.6}{81.6}
\pthmcell{\xN}{-3.10}{82.8}{82.8}
\pthmcell{\xS}{-3.10}{81.9}{81.9}
\pthmcell{\xGm}{-3.10}{73.4}{73.4}
\pthmcell{\xGD}{-3.10}{69.8}{69.8}
\pthmcell{\xGS}{-3.10}{69.4}{69.4}
\pthmcell{\xGeD}{-3.10}{69.8}{69.8}
\pthmcell{\xGeS}{-3.10}{69.3}{69.3}
\pthmcell{\xDSD}{-3.10}{46.5}{46.5}
\pthmcell{\xDSS}{-3.10}{48.2}{48.2}
\pthmcell{\xQWD}{-3.10}{57.9}{57.9}
\pthmcell{\xQWS}{-3.10}{57.8}{57.8}
\pthmrefcell{\xRef}{-3.10}{94.8}{94.8}

\pthmcell{\xD}{-3.58}{77.9}{77.9}
\pthmcell{\xN}{-3.58}{78.6}{78.6}
\pthmcell{\xS}{-3.58}{78.3}{78.3}
\pthmcell{\xGm}{-3.58}{71.5}{71.5}
\pthmcell{\xGD}{-3.58}{64.2}{64.2}
\pthmcell{\xGS}{-3.58}{68.5}{68.5}
\pthmcell{\xGeD}{-3.58}{64.3}{64.3}
\pthmcell{\xGeS}{-3.58}{67.4}{67.4}
\pthmcell{\xDSD}{-3.58}{43.5}{43.5}
\pthmcell{\xDSS}{-3.58}{47.8}{47.8}
\pthmcell{\xQWD}{-3.58}{25.8}{25.8}
\pthmcell{\xQWS}{-3.58}{46.0}{46.0}
\pthmrefcell{\xRef}{-3.58}{92.0}{92.0}

\pthmcell{\xD}{-4.06}{63.5}{63.5}
\pthmcell{\xN}{-4.06}{65.7}{65.7}
\pthmcell{\xS}{-4.06}{60.8}{60.8}
\pthmcell{\xGm}{-4.06}{58.5}{58.5}
\pthmcell{\xGD}{-4.06}{37.5}{37.5}
\pthmcell{\xGS}{-4.06}{34.4}{34.4}
\pthmcell{\xGeD}{-4.06}{47.9}{47.9}
\pthmcell{\xGeS}{-4.06}{49.9}{49.9}
\pthmcell{\xDSD}{-4.06}{49.9}{49.9}
\pthmcell{\xDSS}{-4.06}{48.6}{48.6}
\pthmcell{\xQWD}{-4.06}{38.5}{38.5}
\pthmcell{\xQWS}{-4.06}{47.0}{47.0}
\pthmrefcell{\xRef}{-4.06}{83.7}{83.7}

\node[anchor=south, font=\scriptsize] at (6.80,-4.72) {accuracy scale};
\foreach \i in {0,...,10} {
  \pgfmathtruncatemacro{\mix}{\i*10}
  \fill[draw=white, fill=heatHigh!\mix!heatLow] ({5.81 + 0.18*\i},-4.98) rectangle ({5.99 + 0.18*\i},-4.78);
}
\node[font=\scriptsize, anchor=north] at (5.81,-5.03) {50};
\node[font=\scriptsize, anchor=north] at (7.79,-5.03) {100};

\end{tikzpicture}%
}%
}
\caption{\textbf{Per-type diagnostic heatmap.} Colored-table view of Table~\ref{tab:E1}; cells show answer-only macro accuracy (\%). The color scale is clipped to 50--100, with darker cells indicating higher accuracy. Columns separate structured fusion, the matched-input GPT-$\mu^\ast$ comparator, NL-input LLMs, and the Source Reachability reference.}
\label{fig:pertype-heatmap}
\end{figure}

\subsection{Difficulty-Class Breakdown}\label{subsec:E-2}

\begin{table}[t]
\centering
\caption{T2 fusion methods: per-type $\times$ per-difficulty-class accuracy (\%, 4-seed mean, GPT-5.4-extracted $\hat\mu$). svr = stated\_vs\_revealed; ts = temporal\_shift. Types in Table~\ref{tab:E1} order.}
\label{tab:E2a}
\small
\setlength{\tabcolsep}{5pt}
\begin{tabular}{@{}llcccccc@{}}
\toprule
Type & Diff. & DSNBF & NBF & ABF & SSB & BCF & MV \\
\midrule
\multirow{3}{*}{Ident} & stable & 89.1 & 88.4 & 89.1 & 88.8 & 85.3 & 90.0 \\
 & ts & 87.2 & 87.2 & 87.2 & 86.6 & 84.7 & 84.1 \\
 & svr & 97.8 & 97.5 & 95.0 & 95.6 & 85.3 & 87.2 \\
\cmidrule(l){1-8}
\multirow{3}{*}{Arbit} & stable & 85.2 & 86.5 & 70.6 & 70.6 & 69.8 & 72.9 \\
 & ts & 84.0 & 83.1 & 66.5 & 71.0 & 60.8 & 63.7 \\
 & svr & 81.5 & 83.1 & 46.2 & 81.5 & 46.0 & 43.5 \\
\cmidrule(l){1-8}
\multirow{3}{*}{Ctrl} & stable & 88.4 & 88.1 & 70.9 & 85.9 & 82.2 & 74.4 \\
 & ts & 81.6 & 83.4 & 78.1 & 80.6 & 77.2 & 76.9 \\
 & svr & 84.7 & 85.0 & 85.0 & 87.2 & 84.1 & 85.0 \\
\cmidrule(l){1-8}
\multirow{3}{*}{Factor} & stable & 80.9 & 76.9 & 74.1 & 76.2 & 76.9 & 74.1 \\
 & ts & 79.4 & 74.4 & 65.3 & 68.4 & 67.8 & 65.6 \\
 & svr & 81.9 & 84.4 & 57.5 & 87.8 & 62.8 & 61.6 \\
\cmidrule(l){1-8}
\multirow{3}{*}{Temp} & stable & 85.9 & 83.4 & 83.4 & 84.4 & 84.1 & 81.6 \\
 & ts & 84.4 & 85.3 & 85.0 & 85.0 & 85.3 & 82.8 \\
 & svr & 76.2 & 64.1 & 62.8 & 62.8 & 33.8 & 47.2 \\
\cmidrule(l){1-8}
\multirow{3}{*}{P-R} & stable & 79.1 & 79.4 & 73.8 & 78.4 & 78.4 & 68.1 \\
 & ts & 87.8 & 86.9 & 84.7 & 89.4 & 87.8 & 80.9 \\
 & svr & 68.1 & 70.6 & 59.4 & 69.1 & 69.1 & 56.9 \\
\cmidrule(l){1-8}
\multirow{3}{*}{Miss} & stable & 80.0 & 80.4 & 78.3 & 77.9 & 68.1 & 74.0 \\
 & ts & 80.0 & 79.4 & 78.3 & 80.6 & 70.8 & 73.5 \\
 & svr & 71.7 & 72.3 & 66.7 & 69.6 & 44.4 & 49.0 \\
\cmidrule(l){1-8}
\multirow{3}{*}{Annot} & stable & 69.4 & 70.0 & 64.7 & 65.0 & 65.3 & 64.4 \\
 & ts & 66.6 & 65.9 & 61.9 & 61.3 & 61.3 & 62.8 \\
 & svr & 55.9 & 56.2 & 55.0 & 56.2 & 56.2 & 48.4 \\
\midrule
\multicolumn{2}{l}{Overall drop (stable $\to$ svr)} & $-$5.1 & $-$5.1 & $-$10.6 & $-$1.8 & $-$16.9 & $-$16.4 \\
\bottomrule
\end{tabular}
\end{table}

\begin{table}[t]
\centering
\caption{T3 LLM methods: per-type $\times$ per-difficulty-class accuracy (\%, 4-seed mean). Each (type, difficulty) cell averages across questions in that type and personas in that difficulty class. GPT = GPT-5.4; Gem = Gemini 3.1; DS = DeepSeek V3.2; QW = Qwen3 235B. D = Direct; S = Schema. ts = temporal\_shift; svr = stated\_vs\_revealed.}
\label{tab:E2b}
\small
\setlength{\tabcolsep}{3pt}
\begin{tabular}{@{}llcccccccc@{}}
\toprule
Type & Diff. & GPT-D & GPT-S & Gem-D & Gem-S & DS-D & DS-S & QW-D & QW-S \\
\midrule
\multirow{3}{*}{A-Arbit} & stable & 75.8 & 77.7 & 74.6 & 79.2 & 66.5 & 66.2 & 56.0 & 60.4 \\
 & ts & 81.7 & 81.9 & 80.8 & 84.4 & 64.2 & 64.4 & 59.0 & 65.8 \\
 & svr & 62.1 & 60.4 & 60.4 & 61.7 & 52.1 & 52.5 & 33.8 & 34.0 \\
\cmidrule(l){1-10}
\multirow{3}{*}{B-Ident} & stable & 85.3 & 97.5 & 66.2 & 82.2 & 85.6 & 74.1 & 50.3 & 45.9 \\
 & ts & 77.2 & 94.7 & 64.4 & 76.6 & 75.0 & 63.1 & 39.4 & 38.1 \\
 & svr & 83.1 & 91.2 & 84.1 & 85.3 & 71.6 & 84.7 & 84.7 & 82.2 \\
\cmidrule(l){1-10}
\multirow{3}{*}{C-P-R} & stable & 70.6 & 70.6 & 70.9 & 70.3 & 46.6 & 47.8 & 61.6 & 62.8 \\
 & ts & 80.0 & 79.4 & 79.4 & 78.8 & 46.9 & 50.6 & 51.9 & 49.7 \\
 & svr & 58.8 & 58.1 & 59.1 & 58.8 & 45.9 & 46.2 & 60.3 & 60.9 \\
\cmidrule(l){1-10}
\multirow{3}{*}{D-Temp} & stable & 77.8 & 84.1 & 75.3 & 82.2 & 69.4 & 70.9 & 65.3 & 60.3 \\
 & ts & 80.0 & 82.5 & 77.2 & 82.2 & 54.4 & 77.5 & 62.8 & 49.1 \\
 & svr & 59.7 & 39.4 & 59.7 & 40.0 & 60.6 & 50.9 & 28.8 & 55.0 \\
\cmidrule(l){1-10}
\multirow{3}{*}{E-Factor} & stable & 77.8 & 78.4 & 78.8 & 76.2 & 44.7 & 46.6 & 6.6 & 20.9 \\
 & ts & 83.1 & 82.2 & 78.4 & 82.8 & 48.8 & 53.4 & 21.2 & 26.9 \\
 & svr & 51.6 & 52.2 & 51.6 & 51.6 & 73.8 & 79.1 & 3.4 & 13.8 \\
\cmidrule(l){1-10}
\multirow{3}{*}{F-Miss} & stable & 63.5 & 68.8 & 62.5 & 67.5 & 40.8 & 44.4 & 25.8 & 48.3 \\
 & ts & 66.0 & 70.6 & 65.4 & 67.9 & 40.8 & 49.6 & 26.9 & 50.0 \\
 & svr & 62.9 & 66.0 & 65.0 & 66.7 & 49.0 & 49.6 & 24.6 & 39.6 \\
\cmidrule(l){1-10}
\multirow{3}{*}{G-Annot} & stable & 35.3 & 34.1 & 58.4 & 56.2 & 58.1 & 47.8 & 37.8 & 49.1 \\
 & ts & 40.0 & 35.3 & 52.2 & 52.5 & 46.9 & 47.2 & 37.8 & 49.4 \\
 & svr & 37.2 & 33.8 & 33.1 & 40.9 & 44.7 & 50.9 & 40.0 & 42.5 \\
\cmidrule(l){1-10}
\multirow{3}{*}{Ctrl} & stable & 76.2 & 74.1 & 72.8 & 75.9 & 49.1 & 43.4 & 24.4 & 20.6 \\
 & ts & 81.9 & 78.4 & 79.1 & 79.7 & 49.1 & 43.8 & 45.9 & 45.3 \\
 & svr & 79.1 & 76.9 & 76.6 & 77.5 & 72.8 & 67.8 & 75.3 & 75.6 \\
\midrule
\multicolumn{2}{l}{Overall drop (stable $\to$ svr)} & $-$8.5 & $-$13.4 & $-$8.8 & $-$13.4 & +1.2 & +5.1 & +2.9 & +4.4 \\
\bottomrule
\end{tabular}
\end{table}

Table~\ref{tab:E2a} reveals a bias-modeling gradient on the stated\_vs\_revealed (svr) class: overall svr accuracy drops from stable by only $-$5.1 pp for DSNBF/NBF and $-$1.8 pp for SSB, by $-$10.6 pp for ABF, and by $-$16.9 pp (BCF) and $-$16.4 pp (MV). The gradient tracks how finely each method models source-specific distortion: MV ignores it entirely, BCF captures only a scalar shift, ABF blends bias-aware and bias-free kernels, while NBF/DSNBF learn full per-source confusion matrices (and, for DSNBF, additionally condition on inferred persona class). The type-level view exposes where this collapse concentrates: on D-Temp, BCF and MV fall to 33.8\% and 47.2\% under svr (vs. DSNBF 76.2\%); on A-Arbit, ABF/BCF/MV drop to 43-47\%; on F-Miss, BCF and MV drop to 44-49\%. In contrast, SSB's robustness is uneven: on E-Factor svr it reaches 87.8\% (the highest among non-reference methods), because it picks the objective\_log source and discards the misleading self-report; on A-Arbit svr it ties the confusion-matrix methods (81.5\%) for the same reason.

Table~\ref{tab:E2b} shows that frontier LLM Schema methods degrade sharply under svr: GPT Schema $-$13.4 pp, Gemini Schema $-$13.4 pp. The near-identical degradation across GPT and Gemini indicates a structural limitation of schema prompting, not a family-specific artifact. Schema accuracy falls below Direct under svr (GPT: 59.8\% vs. 61.8\%; Gemini: 60.3\% vs. 61.2\%), indicating overcorrection when stated preferences and revealed behavior diverge systematically. The type-level breakdown localizes where this collapse concentrates: D-Temp svr is where Schema fails most dramatically (GPT-S 39.4\%, Gem-S 40.0\%, vs. their Direct counterparts at 59.7\%); E-Factor svr is the second collapse zone for GPT/Gem Schema (51.6-52.2\%, far below SSB's 87.8\%). B-Ident svr is where Schema actually outperforms Direct (GPT-S 91.2\%, Gem-S 85.3\%), confirming schema descriptions help on identity-style questions but harm trend/factor reasoning under stated-vs-revealed bias.

Open-weight models show a different pattern: their svr drops are near zero or positive (DS Direct +1.2, DS Schema +5.1, QW Direct +2.9, QW Schema +4.4), but only because their stable-class baselines are already low (DS 55-58\%, QW 41-46\%). DeepSeek Schema actually improves on svr (+5.1 pp vs. stable), suggesting that schema descriptions interact differently with weaker models on directionally biased data. Qwen3 collapses on E-Factor across all difficulty classes (3.4-26.9\%), indicating it cannot perform multi-source factor attribution regardless of bias direction.

Across the no-skip tables (E2a, E2b), the relative method ordering on svr is type-stable: DSNBF/NBF lead on 7 of 8 types, BCF/MV trail on 7 of 8 (Annot is the only inversion, where bias modeling no longer helps). The cross-method spread on a given type also tracks how much the per-source signals diverge: D-Temp svr exhibits the largest spread (DSNBF 76.2 vs. BCF 33.8, gap about 42 pp), while B-Ident svr is the tightest among the stronger fusion methods (DSNBF, NBF, ABF, and SSB all fall within 2.5 pp). This monotone coupling between signal-divergence and method-ordering is a sanity check on the diagnostic design: harder difficulty classes do not flatten the comparison; they sharpen it.

Tables~\ref{tab:E2a-skip} and~\ref{tab:E2b-skip} report the SKIP variants on the same per-type $\times$ per-difficulty grid using selective accuracy (correct over answered, excluding skipped cells from the denominator); the second number in each cell is coverage (share of cells answered). The footer convention is stable $-$ SvR selective accuracy: positive values mean SvR is lower than stable, and negative values mean SvR is higher than stable.

The selective tables show a sharper split between methods that expose useful uncertainty and methods that do not. DSNBF+S and NBF+S keep their SvR deficits small (1.5 and 3.3 pp), while the no-skip comparators BCF and MV drop by more than 16 pp. ABF+S sits between them: it abstains aggressively where GT is available (D-Temp SvR coverage 39.4\%, Annot SvR 43.8\%) but still answers enough hard cells to leave an 8.5 pp deficit. T3 LLMs behave differently. GPT and Gemini have large positive footer drops ($+$8.6 to $+$13.5 pp), driven mainly by D-Temp overcorrection under Schema prompting; DeepSeek and Qwen3 show negative footer drops ($-$0.7 to $-$4.0 pp) because their stable-class baselines are weak and their skip rates are near zero. Overall, abstention helps most when the underlying scorer can identify low-confidence cases, which favors per-source confusion-matrix methods over threshold-only heuristics and end-to-end LLM judges.

Comparing answer-only tables (E2a, E2b) against selective tables (E2a-skip, E2b-skip) quantifies the marginal value of abstention per family. DSNBF and NBF show the clearest robustness recovery: answer-only accuracy is 5.1 pp lower on SvR than stable, whereas selective accuracy is lower by only 1.5 and 3.3 pp after thresholding. ABF abstains aggressively (SvR coverage 39.4\% on D-Temp, 43.8\% on Annot); its selective gap improves modestly, from a 10.6 pp answer-only drop to an 8.5 pp selective-accuracy drop. SSB+S shows a 1.1 pp selective drop. BCF and MV are no-skip comparators with 100\% coverage, so their answer-only and selective-table entries are numerically identical. On T3, GPT and Gemini +S keep coverage near 95\% and pay the schema-overcorrection penalty in selective accuracy (Schema drops of 13.5 and 13.1 pp); DS and Qwen3 +S report coverage of 100\% in every cell, indicating that the +SKIP prompt template does not elicit abstention from these models regardless of difficulty class. Abstention is therefore a model-capability axis orthogonal to raw QA accuracy: a method's SvR robustness depends both on whether it knows when to abstain and on whether its cost function makes abstention worthwhile.

\begin{table}[t]
\centering
\caption{T2 selective methods plus no-skip comparators: per-type $\times$ per-difficulty-class selective accuracy (\%, 4-seed mean, GPT-5.4-extracted $\hat\mu$). Each cell is \texttt{sel\_acc (coverage)}: selective accuracy counts only answered cells in the denominator; coverage is the share of cells actually answered. Footer row reports stable $-$ SvR selective accuracy; positive values mean lower SvR than stable, and negative values mean higher SvR than stable. svr = stated\_vs\_revealed; ts = temporal\_shift. DSNBF/NBF/ABF/SSB columns use SKIP variants; BCF and MV are no-skip comparators with 100\% coverage.}
\label{tab:E2a-skip}
\footnotesize
\setlength{\tabcolsep}{3pt}
\begin{tabular}{@{}llcccccc@{}}
\toprule
Type & Diff. & DSNBF+S & NBF+S & ABF+S & SSB+S & BCF & MV \\
\midrule
\multirow{3}{*}{Ident} & stable & 90.3 (96.9) & 90.4 (97.2) & 91.9 (84.4) & 89.0 (99.7) & 85.3 (100.0) & 90.0 (100.0) \\
 & ts & 87.8 (97.5) & 89.3 (93.8) & 90.5 (78.8) & 87.1 (99.1) & 84.7 (100.0) & 84.1 (100.0) \\
 & svr & 97.8 (100.0) & 97.5 (100.0) & 95.9 (84.1) & 96.1 (96.9) & 85.3 (100.0) & 87.2 (100.0) \\
\cmidrule(l){1-8}
\multirow{3}{*}{Arbit} & stable & 87.7 (86.0) & 87.4 (82.9) & 75.9 (78.5) & 73.4 (93.3) & 69.8 (100.0) & 72.9 (100.0) \\
 & ts & 90.2 (82.5) & 88.1 (84.2) & 74.0 (65.0) & 72.9 (87.5) & 60.8 (100.0) & 63.7 (100.0) \\
 & svr & 84.9 (90.8) & 85.8 (91.0) & 62.0 (55.4) & 85.1 (55.8) & 46.0 (100.0) & 43.5 (100.0) \\
\cmidrule(l){1-8}
\multirow{3}{*}{Ctrl} & stable & 91.7 (90.6) & 91.4 (91.2) & 91.8 (53.1) & 95.4 (68.1) & 82.2 (100.0) & 74.4 (100.0) \\
 & ts & 87.4 (89.1) & 88.1 (89.1) & 92.5 (45.6) & 91.7 (75.0) & 77.2 (100.0) & 76.9 (100.0) \\
 & svr & 92.6 (88.4) & 89.8 (92.2) & 93.8 (65.3) & 92.5 (87.5) & 84.1 (100.0) & 85.0 (100.0) \\
\cmidrule(l){1-8}
\multirow{3}{*}{Factor} & stable & 87.4 (77.2) & 86.4 (75.6) & 79.1 (77.8) & 83.1 (83.1) & 76.9 (100.0) & 74.1 (100.0) \\
 & ts & 85.4 (75.0) & 86.1 (74.1) & 77.2 (67.2) & 76.2 (77.5) & 67.8 (100.0) & 65.6 (100.0) \\
 & svr & 88.4 (85.9) & 88.2 (69.1) & 70.4 (69.7) & 90.9 (65.3) & 62.8 (100.0) & 61.6 (100.0) \\
\cmidrule(l){1-8}
\multirow{3}{*}{Temp} & stable & 92.9 (75.3) & 90.9 (78.8) & 93.0 (75.6) & 87.9 (88.1) & 84.1 (100.0) & 81.6 (100.0) \\
 & ts & 90.2 (79.7) & 91.5 (80.9) & 93.1 (76.9) & 87.5 (90.0) & 85.3 (100.0) & 82.8 (100.0) \\
 & svr & 84.9 (64.1) & 69.1 (73.8) & 46.0 (39.4) & 67.6 (81.9) & 33.8 (100.0) & 47.2 (100.0) \\
\cmidrule(l){1-8}
\multirow{3}{*}{P-R} & stable & 81.6 (88.4) & 81.9 (82.8) & 84.6 (71.2) & 80.9 (76.9) & 78.4 (100.0) & 68.1 (100.0) \\
 & ts & 91.3 (89.4) & 92.3 (85.3) & 95.0 (75.0) & 93.6 (78.4) & 87.8 (100.0) & 80.9 (100.0) \\
 & svr & 75.6 (70.3) & 75.7 (73.4) & 78.0 (55.3) & 75.6 (60.3) & 69.1 (100.0) & 56.9 (100.0) \\
\cmidrule(l){1-8}
\multirow{3}{*}{Miss} & stable & 84.7 (75.0) & 86.2 (78.8) & 91.7 (50.4) & 81.4 (93.1) & 68.1 (100.0) & 74.0 (100.0) \\
 & ts & 85.0 (74.8) & 86.6 (72.9) & 94.1 (49.8) & 83.5 (90.8) & 70.8 (100.0) & 73.5 (100.0) \\
 & svr & 87.2 (61.7) & 86.8 (64.8) & 90.6 (24.4) & 73.5 (74.6) & 44.4 (100.0) & 49.0 (100.0) \\
\cmidrule(l){1-8}
\multirow{3}{*}{Annot} & stable & 85.5 (55.9) & 90.7 (46.9) & 73.9 (64.7) & 66.2 (93.4) & 65.3 (100.0) & 64.4 (100.0) \\
 & ts & 77.2 (49.4) & 81.6 (45.9) & 68.2 (60.0) & 62.4 (94.7) & 61.3 (100.0) & 62.8 (100.0) \\
 & svr & 57.4 (33.8) & 66.3 (31.6) & 60.7 (43.8) & 59.0 (80.0) & 56.2 (100.0) & 48.4 (100.0) \\
\midrule
\multicolumn{2}{l}{Overall sel-acc drop (stable $\to$ svr)} & $+$1.5 & $+$3.3 & $+$8.5 & $+$1.1 & $+$16.9 & $+$16.4 \\
\bottomrule
\end{tabular}
\end{table}

\begin{table}[t]
\centering
\caption{T3 LLM methods with selective abstention (SKIP variants): per-type $\times$ per-difficulty-class selective accuracy (\%, 4-seed mean). Each cell is \texttt{sel\_acc (coverage)}; selective accuracy counts only cells the model elected to answer (\texttt{would\_skip = false}). Footer row reports stable $-$ SvR selective accuracy; positive values mean lower SvR than stable, and negative values mean higher SvR than stable. DeepSeek and Qwen3 essentially never abstain (coverage $\approx$ 100\%), so their selective accuracy equals answer-only accuracy. D = Direct prompt; S = Schema prompt; +S suffix marks the SKIP variant.}
\label{tab:E2b-skip}
\footnotesize
\setlength{\tabcolsep}{2pt}
\resizebox{\textwidth}{!}{%
\begin{tabular}{@{}llcccccccc@{}}
\toprule
Type & Diff. & GPT-D+S & GPT-S+S & Gem-D+S & Gem-S+S & DS-D+S & DS-S+S & QW-D+S & QW-S+S \\
\midrule
\multirow{3}{*}{Ident} & stable & 86.9 (92.8) & 97.8 (98.8) & 66.8 (98.8) & 82.2 (100.0) & 85.6 (100.0) & 74.1 (100.0) & 50.3 (100.0) & 45.9 (100.0) \\
 & ts & 78.3 (92.2) & 95.5 (97.2) & 65.0 (99.1) & 76.6 (100.0) & 75.0 (100.0) & 63.1 (100.0) & 39.4 (100.0) & 38.1 (100.0) \\
 & svr & 87.1 (87.5) & 92.9 (96.2) & 84.1 (100.0) & 85.3 (100.0) & 71.6 (100.0) & 84.7 (100.0) & 84.7 (100.0) & 82.2 (100.0) \\
\cmidrule(l){1-10}
\multirow{3}{*}{Arbit} & stable & 77.3 (95.6) & 78.3 (97.1) & 74.9 (99.6) & 79.3 (99.6) & 66.5 (100.0) & 66.2 (100.0) & 55.8 (99.4) & 60.5 (99.8) \\
 & ts & 83.1 (94.0) & 82.9 (96.2) & 81.2 (99.6) & 84.5 (99.6) & 64.2 (100.0) & 64.4 (100.0) & 58.9 (98.8) & 65.3 (97.9) \\
 & svr & 62.5 (93.3) & 61.2 (95.6) & 60.5 (98.8) & 61.6 (99.2) & 52.1 (100.0) & 52.5 (100.0) & 34.3 (95.4) & 34.4 (96.9) \\
\cmidrule(l){1-10}
\multirow{3}{*}{Ctrl} & stable & 78.1 (92.8) & 73.7 (98.8) & 72.7 (99.7) & 75.9 (100.0) & 49.1 (100.0) & 43.4 (100.0) & 24.5 (99.7) & 20.6 (100.0) \\
 & ts & 86.2 (88.4) & 79.8 (95.9) & 79.0 (99.7) & 79.7 (100.0) & 49.1 (100.0) & 43.8 (100.0) & 45.9 (100.0) & 45.3 (100.0) \\
 & svr & 82.9 (87.5) & 78.2 (96.2) & 76.6 (100.0) & 77.5 (100.0) & 72.8 (100.0) & 67.8 (100.0) & 75.3 (100.0) & 75.9 (99.7) \\
\cmidrule(l){1-10}
\multirow{3}{*}{Factor} & stable & 86.5 (83.1) & 84.6 (87.5) & 80.6 (96.6) & 77.0 (99.1) & 44.7 (100.0) & 46.6 (100.0) & 6.6 (99.4) & 21.0 (99.7) \\
 & ts & 86.9 (85.9) & 85.2 (90.6) & 78.8 (98.8) & 83.0 (99.4) & 48.8 (100.0) & 53.4 (100.0) & 21.2 (98.8) & 26.8 (99.1) \\
 & svr & 51.8 (97.8) & 53.3 (95.6) & 51.6 (100.0) & 51.4 (99.7) & 73.8 (100.0) & 79.1 (100.0) & 3.5 (98.4) & 12.9 (97.2) \\
\cmidrule(l){1-10}
\multirow{3}{*}{Temp} & stable & 79.3 (95.3) & 85.3 (97.5) & 75.5 (99.7) & 82.2 (100.0) & 69.4 (100.0) & 70.9 (100.0) & 65.3 (100.0) & 60.3 (100.0) \\
 & ts & 82.0 (93.8) & 83.3 (97.5) & 77.2 (100.0) & 82.2 (100.0) & 54.4 (100.0) & 77.5 (100.0) & 62.8 (100.0) & 49.1 (100.0) \\
 & svr & 60.9 (87.8) & 38.4 (95.9) & 60.1 (99.4) & 40.0 (100.0) & 60.6 (100.0) & 50.9 (100.0) & 28.8 (100.0) & 55.0 (100.0) \\
\cmidrule(l){1-10}
\multirow{3}{*}{P-R} & stable & 72.8 (88.4) & 73.5 (90.9) & 73.8 (90.6) & 72.5 (95.3) & 46.6 (100.0) & 47.8 (100.0) & 61.4 (99.7) & 62.8 (97.5) \\
 & ts & 80.1 (92.8) & 80.9 (94.7) & 80.0 (95.3) & 79.8 (95.9) & 46.9 (100.0) & 50.6 (100.0) & 51.9 (99.4) & 50.8 (93.4) \\
 & svr & 60.7 (85.9) & 62.2 (85.9) & 62.8 (86.6) & 61.4 (92.2) & 45.9 (100.0) & 46.2 (100.0) & 60.3 (95.9) & 61.1 (94.1) \\
\cmidrule(l){1-10}
\multirow{3}{*}{Miss} & stable & 63.8 (98.3) & 70.0 (97.1) & 62.4 (99.8) & 67.4 (99.8) & 40.8 (100.0) & 44.4 (100.0) & 25.8 (100.0) & 48.2 (99.8) \\
 & ts & 66.3 (99.0) & 71.0 (98.3) & 65.4 (100.0) & 67.8 (99.8) & 40.8 (100.0) & 49.6 (100.0) & 26.9 (99.8) & 50.4 (98.3) \\
 & svr & 62.8 (97.9) & 66.1 (99.6) & 65.0 (100.0) & 66.7 (100.0) & 49.0 (100.0) & 49.6 (100.0) & 24.6 (100.0) & 40.2 (98.5) \\
\cmidrule(l){1-10}
\multirow{3}{*}{Annot} & stable & 35.5 (93.4) & 34.3 (94.7) & 58.5 (98.8) & 56.4 (99.7) & 58.1 (100.0) & 47.8 (100.0) & 37.8 (100.0) & 49.1 (100.0) \\
 & ts & 39.5 (90.9) & 36.6 (93.1) & 52.2 (100.0) & 52.5 (100.0) & 46.9 (100.0) & 47.2 (100.0) & 37.8 (100.0) & 49.4 (100.0) \\
 & svr & 39.1 (87.8) & 33.7 (92.8) & 32.8 (99.1) & 40.9 (100.0) & 44.7 (100.0) & 50.9 (100.0) & 39.8 (99.7) & 42.5 (100.0) \\
\midrule
\multicolumn{2}{l}{Overall sel-acc drop (stable $\to$ svr)} & $+$8.9 & $+$13.5 & $+$8.6 & $+$13.1 & $-$0.7 & $-$4.0 & $-$1.4 & $-$2.2 \\
\bottomrule
\end{tabular}%
}
\end{table}

\subsection{Prediction Distributions on Failure Questions}\label{subsec:E-3}

Tables~\ref{tab:E3} and~\ref{tab:E3b} report Schema-Aware prediction distributions on four representative high-failure templates, split by difficulty class: E1 and E2 (factor attribution) in Table~\ref{tab:E3}, C2 (plan-reality) and F3 (missing-data) in Table~\ref{tab:E3b}. These diagnostic distributions use the locked artifacts available for this analysis: GPT seed 1 and 4-seed pooled DeepSeek/Qwen3 runs; Gemini is omitted to keep the table focused. The four diagnostic patterns identified after Table~\ref{tab:E3b} characterise distinct failure modes across model families.

\begin{table}[t]
\centering
\caption{Prediction distributions (\%) on E-type failure questions (E1, E2) by difficulty class. GT = seed 1 ground-truth distribution (40 personas per difficulty class). GPT = Schema-Aware seed 1 (40 per diff); DS/Qwen3 = Schema-Aware 4-seed pooled (160 per diff). Bold = modal prediction per (model, difficulty). Diff codes: St=stable, TS=temporal\_shift, SvR=stated\_vs\_revealed. `---' marks 0\% or missing answer key.}
\label{tab:E3}
\footnotesize
\setlength{\tabcolsep}{3pt}
\begin{tabular}{@{}lllcccc@{}}
\toprule
Q & Answer & Diff & GT & GPT & DS & Qwen3 \\
\midrule
\multirow{15}{*}{\textbf{E1}} & \multirow{3}{*}{no\_single\_factor} & St & 20.0 & 27.5 & \textbf{90.0} & \textbf{54.4} \\
 &  & TS & 25.0 & 17.5 & \textbf{64.4} & 36.9 \\
 &  & SvR & 92.5 & 12.5 & \textbf{78.8} & 20.0 \\
\cmidrule(l){2-7}
 & \multirow{3}{*}{no\_late\_nights} & St & 60.0 & \textbf{37.5} & --- & --- \\
 &  & TS & 27.5 & 25.0 & 0.6 & 0.6 \\
 &  & SvR & --- & --- & --- & --- \\
\cmidrule(l){2-7}
 & \multirow{3}{*}{social\_activity} & St & 20.0 & 35.0 & 1.9 & --- \\
 &  & TS & 27.5 & \textbf{40.0} & 0.6 & --- \\
 &  & SvR & 5.0 & \textbf{87.5} & --- & --- \\
\cmidrule(l){2-7}
 & \multirow{3}{*}{work\_activity} & St & --- & --- & 8.1 & 45.6 \\
 &  & TS & 20.0 & 17.5 & 34.4 & \textbf{62.5} \\
 &  & SvR & 2.5 & --- & 21.2 & \textbf{80.0} \\
\cmidrule(l){2-7}
 & \multirow{3}{*}{Accuracy} & St & --- & 62.5 & 24.4 & 15.6 \\
 &  & TS & --- & 85.0 & 46.9 & 33.1 \\
 &  & SvR & --- & 17.5 & 75.0 & 22.5 \\
\midrule
\multirow{12}{*}{\textbf{E2}} & \multirow{3}{*}{no\_fewer\_than\_30} & St & 72.5 & \textbf{72.5} & \textbf{93.8} & --- \\
 &  & TS & 65.0 & \textbf{70.0} & \textbf{72.5} & --- \\
 &  & SvR & 90.0 & \textbf{82.5} & \textbf{91.2} & --- \\
\cmidrule(l){2-7}
 & \multirow{3}{*}{between\_30\_60} & St & 25.0 & 20.0 & 3.8 & \textbf{97.5} \\
 &  & TS & 30.0 & 30.0 & 3.1 & \textbf{74.4} \\
 &  & SvR & 10.0 & 15.0 & 5.0 & \textbf{81.2} \\
\cmidrule(l){2-7}
 & \multirow{3}{*}{yes\_more\_than\_60} & St & 2.5 & 7.5 & 2.5 & 2.5 \\
 &  & TS & 5.0 & --- & 24.4 & 25.6 \\
 &  & SvR & --- & 2.5 & 3.8 & 18.8 \\
\cmidrule(l){2-7}
 & \multirow{3}{*}{Accuracy} & St & --- & 90.0 & 68.8 & 26.2 \\
 &  & TS & --- & 87.5 & 60.0 & 20.6 \\
 &  & SvR & --- & 82.5 & 83.1 & 5.0 \\
\bottomrule
\end{tabular}
\end{table}

\begin{table}[t]
\centering
\caption{Prediction distributions (\%) on C-type and F-type failure questions (C2, F3) by difficulty class. Same conventions as Table~\ref{tab:E3}; for width, F3 label \texttt{yes\_worked} abbreviates \texttt{yes\_worked\_despite\_no\_entry}.}
\label{tab:E3b}
\footnotesize
\setlength{\tabcolsep}{3pt}
\begin{tabular}{@{}lllcccc@{}}
\toprule
Q & Answer & Diff & GT & GPT & DS & Qwen3 \\
\midrule
\multirow{12}{*}{\textbf{C2}} & \multirow{3}{*}{25\_to\_50\_pct} & St & 57.5 & \textbf{55.0} & 1.2 & \textbf{98.1} \\
 &  & TS & 35.0 & \textbf{52.5} & --- & \textbf{98.8} \\
 &  & SvR & 47.5 & 22.5 & 0.6 & \textbf{99.4} \\
\cmidrule(l){2-7}
 & \multirow{3}{*}{below\_25\_pct} & St & 30.0 & 10.0 & 26.9 & 1.9 \\
 &  & TS & 47.5 & 17.5 & 41.9 & 1.2 \\
 &  & SvR & 22.5 & --- & 37.5 & 0.6 \\
\cmidrule(l){2-7}
 & \multirow{3}{*}{above\_50\_pct} & St & 12.5 & 35.0 & \textbf{71.9} & --- \\
 &  & TS & 17.5 & 30.0 & \textbf{58.1} & --- \\
 &  & SvR & 30.0 & \textbf{77.5} & \textbf{61.9} & --- \\
\cmidrule(l){2-7}
 & \multirow{3}{*}{Accuracy} & St & --- & 62.5 & 20.0 & 50.0 \\
 &  & TS & --- & 62.5 & 33.1 & 37.5 \\
 &  & SvR & --- & 35.0 & 26.2 & 55.6 \\
\midrule
\multirow{12}{*}{\textbf{F3}} & \multirow{3}{*}{both\_occurred} & St & 57.5 & \textbf{70.0} & 23.1 & --- \\
 &  & TS & 70.0 & \textbf{57.5} & 13.1 & --- \\
 &  & SvR & 87.5 & \textbf{80.0} & \textbf{47.5} & --- \\
\cmidrule(l){2-7}
 & \multirow{3}{*}{yes\_worked} & St & 30.0 & 30.0 & 11.9 & \textbf{100.0} \\
 &  & TS & 27.5 & 42.5 & \textbf{44.4} & \textbf{100.0} \\
 &  & SvR & 12.5 & 20.0 & 44.4 & \textbf{100.0} \\
\cmidrule(l){2-7}
 & \multirow{3}{*}{truly\_off} & St & 12.5 & --- & \textbf{65.0} & --- \\
 &  & TS & 2.5 & --- & 42.5 & --- \\
 &  & SvR & --- & --- & 8.1 & --- \\
\cmidrule(l){2-7}
 & \multirow{3}{*}{Accuracy} & St & --- & 62.5 & 28.7 & 34.4 \\
 &  & TS & --- & 72.5 & 25.6 & 26.2 \\
 &  & SvR & --- & 87.5 & 46.9 & 13.8 \\
\bottomrule
\end{tabular}
\end{table}

Four diagnostic patterns emerge from these distributions.

\noindent\textbf{Qwen3 saliency bias.} On E1, Qwen3's modal prediction flips from \texttt{no\_single\_factor} on stable (54.4\%) to \texttt{work\_activity} on TS (62.5\%) and SvR (80.0\%), despite SvR's GT being 92.5\% \texttt{no\_single\_factor}. The TS/SvR fixation on \texttt{work\_activity} drives E1 accuracy down to 33.1\% (TS) and 22.5\% (SvR). On E2, Qwen3 never predicts the GT modal \texttt{no\_fewer\_than\_30} in any difficulty and reaches only 5.0\% on SvR. The errors point to a narratively salient factor dominating elimination and negation answers under bias.

\noindent\textbf{DeepSeek plan-completion overestimation.} On C2, DS predicts \texttt{above\_50\_pct} at 71.9\% (St), 58.1\% (TS), and 61.9\% (SvR), although the GT share is only $12$--$30$\%; the GT modal label \texttt{25\_to\_50\_pct} barely appears ($<1.3$\% in every difficulty class). On F3, DS spreads mass across all three options and changes its modal answer with difficulty: stable favors \texttt{truly\_off} (65.0\%, GT only 12.5\%), TS favors \texttt{yes\_worked} (44.4\%), and SvR favors \texttt{both\_occurred} (47.5\%). The answer distribution therefore shifts with persona difficulty rather than preserving one stable error mode.

\noindent\textbf{Localized catastrophic profiles.} Using 30\% per-(question, difficulty) accuracy as the catastrophic threshold, Qwen3 fails on E1-St (15.6\%), E1-SvR (22.5\%), E2-St (26.2\%), E2-TS (20.6\%), E2-SvR (5.0\%), and F3-SvR (13.8\%). DS fails on E1-St (24.4\%), C2-St (20.0\%), C2-SvR (26.2\%), F3-St (28.7\%), and F3-TS (25.6\%). GPT has one catastrophic cell, E1-SvR (17.5\%). Cells in the 30--50\% range remain weak but fall outside this threshold. SvR concentrates more catastrophic cells across all three models, reinforcing stated-vs-revealed bias as the hardest difficulty class for schema-prompted LLMs.

\noindent\textbf{GPT's localized E1-SvR collapse.} GPT's sharpest failure is concentrated in one (question, difficulty) cell: on E1 SvR, it predicts \texttt{social\_activity} at 87.5\% while GT is 92.5\% \texttt{no\_single\_factor}. This mirrors the Qwen3 failure in form but with a different salient factor; when stated and revealed signals disagree, GPT commits to one plausible explanation instead of the elimination answer. The error stays localized, with F3-SvR accuracy at 87.5\%.

\section{Additional Robustness Analyses}\label{app:robustness}
\subsection{Cross-Seed Stability}\label{subsec:F-1}

We evaluate all methods across four independent random seeds, each instantiating 480 fresh personas. Table~\ref{tab:F1} reports per-seed accuracy for key methods ($\mu^\ast$ input in this diagnostic table).

\begin{table}[t]
\centering
\caption{Cross-seed stability: macro accuracy (\%) across 4 seeds. In this diagnostic table, T1/T2 use $\mu^\ast$ input; T3 uses NL input.}
\label{tab:F1}
\small
\begin{tabular}{@{}lccccc@{}}
\toprule
Method & seed1 & seed2 & seed3 & seed4 & $\sigma$ \\
\midrule
DSNBF & 83.0 & 81.4 & 82.8 & 82.2 & 0.6 \\
NBF & 82.9 & 81.0 & 81.7 & 82.6 & 0.8 \\
GPT-5.4 Direct & 68.9 & 69.3 & 67.6 & 68.6 & 0.6 \\
GPT-5.4 Schema & 69.4 & 70.2 & 68.9 & 70.2 & 0.6 \\
Gemini Direct & 68.1 & 67.3 & 66.9 & 69.0 & 0.9 \\
Gemini Schema & 70.0 & 70.6 & 69.0 & 70.5 & 0.6 \\
DeepSeek Direct & 55.7 & 57.1 & 55.6 & 55.9 & 0.6 \\
DeepSeek Schema & 58.4 & 57.1 & 55.6 & 56.5 & 1.1 \\
Qwen3 Direct & 41.6 & 41.4 & 42.4 & 42.9 & 0.6 \\
Qwen3 Schema & 47.3 & 48.7 & 47.8 & 48.1 & 0.5 \\
\bottomrule
\end{tabular}
\end{table}

Cross-seed standard deviations are $\leq$1.1 pp for all methods. Fusion methods are the most stable ($\sigma \leq 0.8$). All four LLM families show $\sigma \leq 1.1$ pp, confirming that the main findings are stable and not driven by idiosyncratic seed-specific persona configurations. In particular, the 10+ pp trained-resolver vs. zero-shot/prompted LLM gap for frontier models and the 30+ pp gap for open-weight models are reproduced across all four seeds.

\subsection{Training Size Sensitivity}\label{subsec:F-2}

We vary the training set size from 50 to 216 personas (the full training split) and re-train all parametric fusion methods (4-seed mean $\pm$ $\sigma$, $\mu^\ast$ input). Table~\ref{tab:F2} reports the results.

\begin{table}[t]
\centering
\caption{Training size ablation: macro accuracy (\%, 4-seed mean $\pm$ $\sigma$) vs. number of training personas ($\mu^\ast$ input).}
\label{tab:F2}
\small
\setlength{\tabcolsep}{3pt}
\begin{tabular}{@{}lcccccc@{}}
\toprule
Training Size & DSNBF & NBF & ABF & BCF & MV & SSB \\
\midrule
50 & 80.9$\pm$0.6 & 80.5$\pm$0.6 & 71.8$\pm$1.0 & 69.3$\pm$0.7 & 69.5$\pm$0.6 & 78.2$\pm$0.5 \\
100 & 82.0$\pm$0.7 & 81.7$\pm$0.6 & 71.8$\pm$0.9 & 69.3$\pm$0.7 & 69.5$\pm$0.6 & 78.8$\pm$0.6 \\
150 & 82.3$\pm$0.7 & 81.9$\pm$0.6 & 71.8$\pm$0.9 & 69.4$\pm$0.7 & 69.5$\pm$0.6 & 78.8$\pm$0.6 \\
216 (full) & \textbf{82.3$\pm$0.7} & \textbf{82.0$\pm$0.9} & 71.8$\pm$1.0 & 69.4$\pm$0.7 & 69.5$\pm$0.6 & 79.0$\pm$0.5 \\
\bottomrule
\end{tabular}
\end{table}

Performance largely saturates by $N = 100$: the gain from doubling the training set (100 $\to$ 216) is $\leq$0.36 pp (4-seed mean). This suggests the source-to-GT mappings are regular enough to estimate from the train split. MV is invariant to training size by construction (no learnable parameters). ABF and BCF have learnable parameters but change by $\leq$0.1 pp over this range, suggesting their low-dimensional parameterizations also saturate quickly.

\subsection{GPT-5.4 vs. Gemini Cross-Condition Comparison}\label{subsec:F-3}

\begin{table}[t]
\centering
\caption{GPT-5.4 vs. Gemini 3.1 Pro (seed 1, all four information conditions).}
\label{tab:F3}
\small
\begin{tabular}{@{}lccc@{}}
\toprule
Information Condition & GPT-5.4 (\%) & Gemini (\%) & $\Delta$ (pp) \\
\midrule
LLM Direct (NL) & 68.9 & 68.1 & $-$0.8 \\
Schema Aware (NL+bias hints) & 69.4 & 70.0 & +0.6 \\
{$\hat\mu$ input} & 73.3 & 75.5 & +2.2 \\
{$\mu^\ast$ input} & 74.2 & 74.6 & +0.4 \\
\bottomrule
\end{tabular}
\end{table}

GPT-5.4 and Gemini remain within $\pm$2.2 pp under all four information conditions (seed 1). The Direct and Schema rows are consistent with the 4-seed pooled T3 rows in Table~\ref{tab:4}. The $\hat\mu$ input and $\mu^\ast$ input rows overlap with the LLM column of Table~\ref{tab:6} (\S~\ref{s4:sub3}). This aggregate similarity coexists with large per-question extraction differences in Table~\ref{tab:F6b}.

\subsection{DGP Perturbation}\label{subsec:F-4}

We vary the DGP's two core distributional parameters (bias scale and device dropout scale) across a 3$\times$3 grid ($\{0.5\times, 1.0\times, 2.0\times\}^2$), yielding 9 distribution variants per seed. The default configuration is $1.0\times / 1.0\times$. Table~\ref{tab:F4} reports macro accuracy (\%) for key methods across all 9 variants (4-seed mean $\pm$ $\sigma$, $\mu^\ast$ input).

\begin{table}[t]
\centering
\caption{DGP perturbation: macro accuracy (\%, 4-seed mean $\pm$ $\sigma$) across 9 bias$\times$dropout variants ($\mu^\ast$ input). b = bias scale; d = dropout scale. Default variant in bold. SSB-G denotes a global single-source baseline used only for this perturbation table; it is not the per-question train-selected SSB row in Table~\ref{tab:4}.}
\label{tab:F4}
\small
\setlength{\tabcolsep}{3pt}
\begin{tabular}{@{}lccccccc@{}}
\toprule
Variant & DSNBF & NBF & ABF & BCF & MV & SSB-G & Ref. \\
\midrule
b0.5, d0.5 & 84.4$\pm$0.9 & 83.7$\pm$0.8 & 74.5$\pm$0.3 & 71.6$\pm$0.7 & 73.0$\pm$0.7 & 76.2$\pm$0.7 & 93.7$\pm$0.3 \\
b0.5, d1.0 & 83.1$\pm$1.2 & 82.7$\pm$0.8 & 74.5$\pm$1.0 & 70.2$\pm$0.7 & 72.4$\pm$0.7 & 77.4$\pm$0.8 & 94.0$\pm$0.3 \\
b0.5, d2.0 & 83.6$\pm$1.2 & 83.0$\pm$0.9 & 73.9$\pm$0.9 & 68.3$\pm$1.0 & 71.5$\pm$0.9 & 78.6$\pm$0.9 & 94.1$\pm$0.5 \\
b1.0, d0.5 & 84.2$\pm$1.0 & 83.7$\pm$0.8 & 74.3$\pm$0.8 & 71.3$\pm$0.7 & 69.8$\pm$0.3 & 68.9$\pm$0.8 & 92.9$\pm$0.4 \\
\textbf{b1.0, d1.0} & \textbf{82.3$\pm$0.6} & \textbf{82.0$\pm$0.7} & \textbf{72.0$\pm$0.8} & \textbf{69.8$\pm$0.8} & \textbf{69.5$\pm$0.6} & \textbf{70.1$\pm$0.9} & \textbf{93.2$\pm$0.7} \\
b1.0, d2.0 & 81.4$\pm$0.5 & 80.6$\pm$0.6 & 70.5$\pm$0.8 & 68.0$\pm$0.4 & 68.6$\pm$0.6 & 71.4$\pm$0.7 & 92.8$\pm$0.4 \\
b2.0, d0.5 & 84.0$\pm$0.5 & 83.0$\pm$0.5 & 71.3$\pm$0.7 & 70.6$\pm$0.9 & 63.4$\pm$0.2 & 57.3$\pm$1.3 & 91.5$\pm$0.4 \\
b2.0, d1.0 & 82.4$\pm$1.1 & 80.7$\pm$1.0 & 70.3$\pm$0.5 & 69.4$\pm$0.8 & 63.0$\pm$0.4 & 58.5$\pm$1.2 & 91.2$\pm$0.5 \\
b2.0, d2.0 & 80.8$\pm$0.7 & 79.1$\pm$0.9 & 68.6$\pm$0.7 & 67.9$\pm$0.6 & 62.6$\pm$0.5 & 59.7$\pm$1.1 & 90.5$\pm$0.3 \\
\bottomrule
\end{tabular}
\end{table}

The mean pairwise Kendall $\tau$ across all $\binom{9}{2} = 36$ variant-pair ranking comparisons (canonical 7-column set: DSNBF, NBF, ABF, BCF, MV, SSB-G, Source Reachability reference) is 0.77, with a bootstrap 95\% CI of $[0.74, 0.80]$ over 144 pairs (4 seeds $\times$ 36 pairs).\footnote{We report aggregate effect-size statistics rather than per-pair significance because each pairwise Kendall test compares only $n = 7$ ranks, which yields a theoretical $p$-value floor of $\approx 0.0014$ regardless of the underlying data. The aggregate signal across 144 pairs (mean $\tau$ lies $\approx 52$ standard errors above the random-ranking null of 0) is the appropriate evidence for ranking stability.} Every one of the 144 pairs has $\tau > 0$, and 143 of 144 have $\tau > 0.5$. No aggregate cell shows a full rank reversal, and only one falls below the strong-correlation threshold.

Each variant is retrained on its own training split, so this tests whether the model class is appropriate across bias scales, not whether specific parameters transfer. Under simultaneously doubled bias and dropout (b2.0, d2.0), bias-unaware methods (MV) degrade sharply ($-$6.9 pp from the default setting, 4-seed mean), while DSNBF degrades only $-$1.6 pp. DSNBF leads or ties (within cross-seed variability) every other fusion method in 32/36 $\times$ (seed, variant) cells; in the remaining 4 cells, NBF leads DSNBF by $< 0.7$ pp, well within $\sigma_{\text{DSNBF}}$.

SSB-G's accuracy range across the 9 DGP variants is 21.3 pp (4-seed mean), compared to 3.7 pp for DSNBF and 4.6 pp for NBF. Under low bias (b0.5), SSB-G benefits from device-log dominance and performs well (76–79\%); under high bias (b2.0), device-log dropout compounds with amplified distortion in other sources, and SSB-G collapses to 57–60\%. The global source-selection heuristic cannot adapt to distributional shifts where no single source remains uniformly reliable, whereas confusion-matrix methods redistribute weight across sources as bias conditions change.

\subsection{Extraction Noise Tolerance (Full Analysis)}\label{subsec:F-5}

We simulate extraction errors by randomly flipping a fraction of $\mu^\ast$ values to uniformly random labels. Methods are fit and calibrated on the unperturbed train and calibration splits; only test atoms are flipped. Table~\ref{tab:F5} reports macro accuracy under increasing noise rates (4-seed mean, $\mu^\ast$ as baseline).

\begin{table}[t]
\centering
\caption{Extraction noise tolerance: macro accuracy (\%, 4-seed mean) under random label flips applied to $\mu^\ast$. MC = Majority Class.}
\label{tab:F5}
\small
\setlength{\tabcolsep}{3pt}
\begin{tabular}{@{}lcccccc@{}}
\toprule
Flip Rate & DSNBF & NBF & ABF & BCF & MV & MC \\
\midrule
0\% & 82.3 & 82.0 & 72.0 & 69.8 & 69.5 & 57.1 \\
10\% & 74.8 & 75.5 & 68.0 & 64.8 & 65.6 & 57.1 \\
20\% & 67.0 & 68.5 & 63.0 & 60.5 & 61.0 & 57.1 \\
30\% & 59.7 & 62.3 & 58.0 & 55.9 & 55.9 & 57.1 \\
50\% & 42.3 & 47.0 & 46.4 & 44.7 & 45.2 & 57.1 \\
\bottomrule
\end{tabular}
\end{table}

At 30\% noise, all learned methods converge toward the Majority Class baseline (57.1\%), as expected: random flips destroy the signal that confusion matrices rely on. Majority Class is noise-invariant because it ignores input entirely and always predicts the per-question modal label.

A reversal occurs at $\epsilon \geq 0.1$: NBF overtakes DSNBF, and the gap widens from $-$0.6 pp at $\epsilon = 0.1$ to $-$4.7 pp at $\epsilon = 0.5$ (4-seed mean). DSNBF's difficulty-stratified weights concentrate influence on hard personas, which are also the most noise-sensitive. DSNBF's cross-seed variability increases under perturbation, while NBF remains more stable. This pattern indicates that difficulty conditioning helps under clean atoms but can amplify random extraction noise when the atom table is heavily corrupted.

\subsection{Cross-Extractor Robustness}\label{subsec:F-6}

Main paper results use GPT-5.4 as the extraction backend (NL $\to \hat\mu$). To answer whether fusion accuracy is an artifact of a single extractor, we repeat the full extraction pipeline with Gemini 3.1 Pro (the other closed frontier family evaluated as an LLM-Direct answerer) on all four seeds and feed the resulting atoms $\hat\mu^{\text{Gem}}$ to every fusion method. The split and fitted fusion parameters are held fixed; only the test-set extraction backend changes. Table~\ref{tab:F6} compares the two extraction backends; the direct readout column $\mu^\ast$ is the no NL extraction reference.

\begin{table}[t]
\centering
\caption{Fusion macro accuracy (\%, 4-seed mean) under two extraction backends. $\mu^\ast$: direct structured readout reference. $\hat\mu^{\text{GPT}}$, $\hat\mu^{\text{Gem}}$: GPT-5.4 / Gemini 3.1 Pro extraction. $\Delta$: Gemini $-$ GPT. Non-selective variants.}
\label{tab:F6}
\small
\begin{tabular}{@{}lcccc@{}}
\toprule
Method & $\mu^\ast$ & $\hat\mu^{\text{GPT}}$ & $\hat\mu^{\text{Gem}}$ & $\Delta$ (pp) \\
\midrule
DSNBF-NS & 82.3 & 80.3 & 79.6 & $-$0.7 \\
NBF-NS   & 82.1 & 79.8 & 78.9 & $-$0.9 \\
SSB      & 79.0 & 77.2 & 74.9 & $-$2.3 \\
ABF-NS   & 72.0 & 72.0 & 72.4 & \phantom{$-$}0.4 \\
BCF(4p)  & 69.8 & 69.2 & 68.3 & $-$0.9 \\
MajVote  & 69.5 & 68.8 & 67.4 & $-$1.3 \\
\bottomrule
\end{tabular}
\end{table}

The extractor swap preserves the main method ordering: under both extractors, DSNBF $\approx$ NBF $>$ SSB $>$ ABF $\approx$ BCF $\approx$ MajVote, with no rank flip among the top four. DSNBF-NS drops only $0.7$ pp under Gemini extraction, comparable to its own cross-seed variability under each extractor ($\sigma_{\text{GPT}} = 0.2$, $\sigma_{\text{Gem}} = 0.6$ pp; 4-seed pstdev). This supports the \S~\ref{s4:sub2} attribution of the trained-resolver vs. prompted-LLM gap across two comparable closed frontier extractors. Sensitivity is method dependent: SSB loses $2.3$ pp because it commits to a single source and inherits that source's extraction quality directly, while ABF is essentially unchanged ($\Delta = +0.4$ pp; sign mixed across seeds). NBF and DSNBF are less sensitive because their learned source-to-GT maps tolerate small extractor changes. The table tests transfer under a fixed fitted resolver; per-extractor refitting would be a separate condition.

Selective variants show the same qualitative pattern: DSNBF selective accuracy changes from $85.3\%$ at $78.3\%$ coverage with GPT extraction to $83.1\%$ at $79.3\%$ coverage with Gemini extraction (all 4-seed means; full table in the data release). In this GPT-to-Gemini check, the same thresholding protocol remains effective.

\paragraph{Scope.} We test GPT-5.4 and Gemini 3.1 Pro. Their LLM-Direct accuracies are within $\pm 1$ pp of each other (Table~\ref{tab:F3}), so this check compares two broadly similar extractor tiers. The statement concerns answer-level Direct accuracy, while the per-question atom extraction differences are reported below. An extractor with substantially lower baseline quality would be expected to widen the downstream gap in proportion to its extraction faithfulness rate (Appendix~\ref{app:audit}).

\paragraph{Per-question extraction accuracy.}
To diagnose where the two extractors diverge, Table~\ref{tab:F6b} reports per-question atom-level extraction accuracy $P(\hat\mu = \mu^\ast)$ against the direct readout, aggregated over all four seeds (480 test personas per question, non-null-cell denominator).

\begin{table}[t]
\centering
\caption{Per-question $\times$ per-difficulty extraction accuracy (\%, 4-seed mean $\pm$ $\sigma$, 160 test personas per (question, difficulty) cell). Columns: GPT-5.4 and Gemini 3.1 Pro extraction vs.\ direct readout atoms $\mu^\ast$. Difficulty codes: St=stable, TS=temporal\_shift, SvR=stated\_vs\_revealed. $\Delta$ = Gemini mean $-$ GPT mean. Bold marks $|\Delta| \geq 5$ pp.}
\label{tab:F6b}
\scriptsize
\setlength{\tabcolsep}{2.5pt}
\begin{tabular}{@{}llrrr|llrrr@{}}
\toprule
Q & Diff & GPT & Gem & $\Delta$ & Q & Diff & GPT & Gem & $\Delta$ \\
\midrule
\multirow{3}{*}{A1} & St & $92.0{\pm}0.9$ & $91.6{\pm}1.1$ & $-$0.5 & \multirow{3}{*}{E1} & St & $88.1{\pm}4.8$ & $73.3{\pm}0.3$ & $\mathbf{-14.8}$ \\
 & TS & $93.9{\pm}1.9$ & $93.9{\pm}1.5$ & $+$0.0 &  & TS & $86.1{\pm}2.7$ & $72.8{\pm}1.3$ & $\mathbf{-13.3}$ \\
 & SvR & $99.2{\pm}1.0$ & $99.5{\pm}0.5$ & $+$0.3 &  & SvR & $82.5{\pm}1.7$ & $70.8{\pm}4.9$ & $\mathbf{-11.7}$ \\
\cmidrule(l){1-5}\cmidrule(l){6-10}
\multirow{3}{*}{A2} & St & $87.4{\pm}0.6$ & $92.8{\pm}1.8$ & $\mathbf{+5.4}$ & \multirow{3}{*}{E2} & St & $88.5{\pm}2.0$ & $89.5{\pm}2.4$ & $+$1.0 \\
 & TS & $86.2{\pm}1.5$ & $92.6{\pm}1.6$ & $\mathbf{+6.4}$ &  & TS & $89.8{\pm}2.5$ & $89.9{\pm}2.2$ & $+$0.1 \\
 & SvR & $77.2{\pm}0.8$ & $76.4{\pm}0.7$ & $-$0.9 &  & SvR & $94.7{\pm}2.6$ & $96.4{\pm}0.8$ & $+$1.7 \\
\cmidrule(l){1-5}\cmidrule(l){6-10}
\multirow{3}{*}{A3} & St & $99.2{\pm}0.7$ & $98.3{\pm}1.2$ & $-$0.9 & \multirow{3}{*}{F1} & St & $91.6{\pm}1.3$ & $92.2{\pm}1.3$ & $+$0.6 \\
 & TS & $98.0{\pm}0.5$ & $98.6{\pm}0.8$ & $+$0.6 &  & TS & $92.3{\pm}2.0$ & $92.7{\pm}2.1$ & $+$0.3 \\
 & SvR & $88.8{\pm}1.3$ & $95.5{\pm}1.7$ & $\mathbf{+6.7}$ &  & SvR & $96.0{\pm}2.4$ & $98.1{\pm}1.5$ & $+$2.1 \\
\cmidrule(l){1-5}\cmidrule(l){6-10}
\multirow{3}{*}{B2} & St & $95.8{\pm}1.5$ & $97.1{\pm}0.6$ & $+$1.4 & \multirow{3}{*}{F2} & St & $77.1{\pm}3.4$ & $68.9{\pm}2.2$ & $\mathbf{-8.2}$ \\
 & TS & $95.9{\pm}1.4$ & $97.0{\pm}1.5$ & $+$1.1 &  & TS & $82.9{\pm}2.1$ & $72.0{\pm}4.2$ & $\mathbf{-10.9}$ \\
 & SvR & $96.8{\pm}1.3$ & $99.0{\pm}0.4$ & $+$2.2 &  & SvR & $84.0{\pm}1.9$ & $75.4{\pm}1.2$ & $\mathbf{-8.6}$ \\
\cmidrule(l){1-5}\cmidrule(l){6-10}
\multirow{3}{*}{B3} & St & $83.6{\pm}5.2$ & $78.8{\pm}1.3$ & $-$4.8 & \multirow{3}{*}{F3} & St & $93.1{\pm}1.5$ & $83.8{\pm}3.1$ & $\mathbf{-9.4}$ \\
 & TS & $83.5{\pm}3.3$ & $80.8{\pm}2.0$ & $-$2.7 &  & TS & $96.6{\pm}2.4$ & $90.2{\pm}3.2$ & $\mathbf{-6.4}$ \\
 & SvR & $95.0{\pm}1.6$ & $85.5{\pm}1.4$ & $\mathbf{-9.5}$ &  & SvR & $96.9{\pm}1.3$ & $95.2{\pm}0.9$ & $-$1.7 \\
\cmidrule(l){1-5}\cmidrule(l){6-10}
\multirow{3}{*}{C2} & St & $98.3{\pm}1.3$ & $98.8{\pm}0.4$ & $+$0.4 & \multirow{3}{*}{G1} & St & $97.9{\pm}1.2$ & $98.5{\pm}1.1$ & $+$0.6 \\
 & TS & $98.4{\pm}1.5$ & $99.0{\pm}0.3$ & $+$0.6 &  & TS & $99.4{\pm}0.4$ & $99.2{\pm}0.6$ & $-$0.2 \\
 & SvR & $96.5{\pm}1.8$ & $95.4{\pm}1.2$ & $-$1.0 &  & SvR & $97.9{\pm}1.3$ & $99.8{\pm}0.4$ & $+$1.9 \\
\cmidrule(l){1-5}\cmidrule(l){6-10}
\multirow{3}{*}{C3} & St & $76.9{\pm}2.3$ & $77.8{\pm}2.4$ & $+$0.9 & \multirow{3}{*}{G2} & St & $86.1{\pm}1.0$ & $90.9{\pm}1.7$ & $+$4.8 \\
 & TS & $92.2{\pm}1.0$ & $94.4{\pm}1.9$ & $+$2.2 &  & TS & $86.2{\pm}0.4$ & $88.0{\pm}2.8$ & $+$1.7 \\
 & SvR & $71.2{\pm}3.6$ & $74.4{\pm}4.7$ & $+$3.1 &  & SvR & $84.4{\pm}2.2$ & $83.5{\pm}3.4$ & $-$0.8 \\
\cmidrule(l){1-5}\cmidrule(l){6-10}
\multirow{3}{*}{D1} & St & $92.5{\pm}1.6$ & $94.4{\pm}1.6$ & $+$1.9 & \multirow{3}{*}{Ctrl1} & St & $98.5{\pm}1.2$ & $78.5{\pm}0.9$ & $\mathbf{-20.0}$ \\
 & TS & $91.9{\pm}1.6$ & $92.7{\pm}1.4$ & $+$0.8 &  & TS & $97.5{\pm}1.9$ & $80.2{\pm}3.3$ & $\mathbf{-17.3}$ \\
 & SvR & $89.6{\pm}1.8$ & $91.7{\pm}1.9$ & $+$2.1 &  & SvR & $97.7{\pm}0.9$ & $76.9{\pm}7.4$ & $\mathbf{-20.8}$ \\
\cmidrule(l){1-5}\cmidrule(l){6-10}
\multirow{3}{*}{D2} & St & $99.4{\pm}1.1$ & $100.0{\pm}0.0$ & $+$0.6 & \multirow{3}{*}{Ctrl2} & St & $99.4{\pm}1.1$ & $98.8{\pm}0.4$ & $-$0.6 \\
 & TS & $100.0{\pm}0.0$ & $99.4{\pm}0.6$ & $-$0.6 &  & TS & $99.4{\pm}0.7$ & $97.9{\pm}1.7$ & $-$1.5 \\
 & SvR & $96.2{\pm}1.8$ & $98.8{\pm}0.0$ & $+$2.5 &  & SvR & $93.1{\pm}3.0$ & $94.2{\pm}2.5$ & $+$1.0 \\
\midrule
\multicolumn{10}{l}{Overall (St): GPT $90.9{\pm}0.6\%$, Gemini $88.6{\pm}0.6\%$; $\Delta = -2.3$ pp.} \\
\multicolumn{10}{l}{Overall (TS): GPT $92.0{\pm}0.7\%$, Gemini $89.8{\pm}0.4\%$; $\Delta = -2.2$ pp.} \\
\multicolumn{10}{l}{Overall (SvR): GPT $90.9{\pm}1.1\%$, Gemini $88.9{\pm}0.4\%$; $\Delta = -2.0$ pp.} \\
\bottomrule
\end{tabular}
\end{table}

The per-question extraction differences are clustered. Most of the overall Gemini deficit ($-2.0$ to $-2.3$ pp across difficulties) comes from five question families that degrade on all three difficulty classes: Ctrl1 ($-17$ to $-21$), E1 ($-12$ to $-15$), F2 ($-8$ to $-11$), F3 ($-6$ to $-9$ on St/TS), and B3 (especially on SvR at $-9.5$). Ctrl1 and E1 show Gemini underperformance that does not track persona difficulty, consistent with a structural extractor weakness. Some gaps are difficulty-sensitive: Gemini outperforms GPT on A3 SvR ($+6.7$) and A2 St/TS ($+5.4$/$+6.4$), but A2 reverses on SvR ($-0.9$), while B3 widens on SvR. The error balance is also asymmetric. Gemini produces more false-negative cells ($\mu^\ast$ non-null, Gemini emits null), consistent with a more conservative extractor that drops atoms under uncertainty; we document the per-seed counts in the data release.

Despite per-(question, difficulty) extraction gaps reaching $20.8$ pp, the fusion-level macro accuracy gap (Table~\ref{tab:F6}) stays within $|\Delta| \leq 1.2$ pp for confusion-matrix methods (ABF $+0.4$, DSNBF $-0.7$, NBF $-0.9$, BCF $-1.2$). The same fitted confusion-matrix resolver absorbs these extractor differences at the aggregate level. SSB's steeper $-2.3$ pp drop reflects its single-source design: it reads one chosen source atom and has no learned cross-source absorption mechanism. The selective variants in Appendix~\ref{subsec:D-2} show the same pattern.

\subsection{Cross-Parameter Transfer Without Refit}\label{subsec:F-7}

Table~\ref{tab:F7} tests a stricter transfer setting than Table~\ref{tab:F4}. DSNBF is fit and calibrated only on the default projection setting ($b{=}1.0,d{=}1.0$), then evaluated without refit on shifted bias/dropout variants. The target-retrained column is recomputed on the same rebuilt variant, so both columns use the same test data. All rows use direct-readout atoms $\mu^\ast$.

\begin{table}
\centering
\definecolor{csevenLow}{HTML}{F4F6F8}
\definecolor{csevenHigh}{HTML}{6BAED6}
\caption{Cross-parameter transfer without refit: DSNBF macro accuracy (\%, 4-seed mean $\pm$ $\sigma$). Transfer fits once on the default $b{=}1.0,d{=}1.0$ setting and evaluates on each target variant without refit. Rows vary bias scale $b$; columns vary dropout scale $d$. Each cell reports no-refit transfer accuracy; the second line reports transfer $-$ target-retrained gap. Darker cells indicate higher no-refit transfer accuracy.}
\label{tab:F7}
\small
\begin{tikzpicture}[
  every node/.style={font=\scriptsize},
  csevencell/.style={rectangle, minimum width=28mm, minimum height=10.5mm,
                     draw=white, line width=0.45pt, align=center,
                     inner sep=1pt, text=black},
  csevendefault/.style={rectangle, minimum width=28mm, minimum height=10.5mm,
                        draw=black!70, line width=0.55pt, align=center,
                        inner sep=1pt, text=black},
  csevenlabel/.style={font=\scriptsize\bfseries, align=center},
]
\newcommand{\csevencell}[5]{%
  \pgfmathtruncatemacro{\mix}{round(max(0,min(100,(#3-78)*12.5)))}%
  \node[csevencell, fill=csevenHigh!\mix!csevenLow] at (#1,#2) {#4\\{\scriptsize #5}};%
}
\newcommand{\csevendefaultcell}[5]{%
  \pgfmathtruncatemacro{\mix}{round(max(0,min(100,(#3-78)*12.5)))}%
  \node[csevendefault, fill=csevenHigh!\mix!csevenLow] at (#1,#2) {\textbf{#4}\\{\scriptsize \textbf{#5}}};%
}

\node[csevenlabel] at (1.9,0.65) {$d{=}0.5$};
\node[csevenlabel] at (4.9,0.65) {$d{=}1.0$};
\node[csevenlabel] at (7.9,0.65) {$d{=}2.0$};
\node[csevenlabel, anchor=east] at (0.25,-0.35) {$b{=}0.5$};
\node[csevenlabel, anchor=east] at (0.25,-1.55) {$b{=}1.0$};
\node[csevenlabel, anchor=east] at (0.25,-2.75) {$b{=}2.0$};

\csevencell{1.9}{-0.35}{83.7}{83.7$\pm$1.0}{gap $-$0.7$\pm$0.2}
\csevencell{4.9}{-0.35}{83.1}{83.1$\pm$0.8}{gap 0.0$\pm$0.6}
\csevencell{7.9}{-0.35}{81.6}{81.6$\pm$1.0}{gap $-$2.0$\pm$0.3}

\csevencell{1.9}{-1.55}{82.9}{82.9$\pm$1.1}{gap $-$1.4$\pm$1.3}
\csevendefaultcell{4.9}{-1.55}{82.3}{82.3$\pm$0.6}{gap 0.0$\pm$0.0}
\csevencell{7.9}{-1.55}{80.9}{80.9$\pm$0.6}{gap $-$0.5$\pm$0.3}

\csevencell{1.9}{-2.75}{80.2}{80.2$\pm$1.6}{gap $-$3.8$\pm$1.2}
\csevencell{4.9}{-2.75}{80.0}{80.0$\pm$1.5}{gap $-$2.4$\pm$0.9}
\csevencell{7.9}{-2.75}{79.2}{79.2$\pm$1.2}{gap $-$1.6$\pm$1.1}
\end{tikzpicture}
\end{table}

Across the eight shifted variants, DSNBF trained only on the default projection setting is 1.6 pp below DSNBF refit on each target setting on average, with a worst gap of 3.8 pp at $b{=}2.0,d{=}0.5$. This supports moderate cross-parameter stability across projection settings. The negative gaps also show that target-distribution calibration can still help under larger shifts.

\end{document}